\documentclass[5p,10pt, twocolumns]{elsarticle}

\newcommand{\scalefactor}{1.0}
\usepackage{graphicx}
\usepackage{cuted}
\usepackage{multicol}
\usepackage{multirow}

\usepackage{url}
\usepackage{amsmath,amssymb}

\let\yesnumberold=\yesnumber\relax
\let\yesnumber\relax
\usepackage[math]{easyeqn}
\let\yesnumber\yesnumberold
\usepackage{bm}
\usepackage{subfig}
\usepackage[hidelinks]{hyperref}
\usepackage{nomencl}
\makenomenclature

\hypersetup
{
	pdftitle = {title},
	pdfauthor = {authors},
	pdfsubject = {climbing robot},
	pdfkeywords = {keywords},
	pdftoolbar = true,
	colorlinks = false,
	linkcolor = black,
	citecolor = black,
	urlcolor = black,
}

\usepackage[usenames,dvipsnames]{xcolor}
\definecolor{blue_iit}{RGB}{51,51,255}
\usepackage[acronym, nonumberlist]{glossaries}

\usepackage[tight]{units}
\usepackage[normalem]{ulem} 
\definecolor{Gray}{gray}{0.9}
\usepackage{multirow}
\usepackage{booktabs}

\newacronym{lf}{LF}{Left-Front}
\newacronym{rf}{RF}{Right-Front}
\newacronym{lh}{LH}{Left-Hind}
\newacronym{rh}{RH}{Right-Hind}

\newacronym{haa}{HAA}{Hip Adduction-Abduction}
\newacronym{hfe}{HFE}{Hip Flexion-Extension}
\newacronym{kfe}{KFE}{Knee Flexion-Extension}

\newacronym{imu}{IMU}{Inertial Measurement Unit}
\newacronym{dofs}{DoFs}{Degrees of Freedom}
\newacronym{rt}{RT}{Real Time}

\newacronym{urdf}{URDF}{Unified Robot Description Format}

\newacronym{com}{CoM}{Center of Mass}
\newacronym{cop}{CoP}{Center of Pressure}
\newacronym{zmp}{ZMP}{Zero Moment Point}
\newacronym{icp}{ICP}{Instantaneous Capture Point}
\newacronym{cmp}{CMP}{Centroidal Moment Pivot}
\newacronym{grfs}{GRFs}{Ground Reaction Forces}

\newacronym{ls}{LS}{Least Square}
\newacronym{lp}{LP}{Linear Program}

\newacronym{slip}{SLIP}{Spring Loaded Inverted Pendulum}
\newacronym{eom}{EoM}{Equation of Motions}
\newacronym{qp}{QP}{Quadratic Program}
\newacronym{sqp}{SQP}{Sequential Quadratic Programming}
\newacronym{mic}{MIC}{Mixed-Integer Convex}
\newacronym{cmaes}{CMA-ES}{Covariance Matrix Adaptation Evolution Strategy}
\newacronym{ara}{ARA*}{Anytime Repairing A*}
\newacronym{pca}{PCA}{Principal Component Analysis}
\newacronym{cpg}{CPG}{Central Pattern Generator}
\newacronym{wbc}{WBC}{Whole-Body Control}
\newacronym{cwc}{CWC}{Contact Wrench Cone}
\newacronym{fwp}{FWP}{Feasible Wrench Polytope}
\newacronym{nlp}{NLP}{Non Linear Program}
\newacronym{ocp}{OCP}{Optimal Control Problem}
\newacronym{pd}{PD}{Proportional-Derivative}

\newacronym{mpc}{MPC}{Model Predictive Control}
\newacronym{nmpc}{NMPC}{Nonlinear Model Predictive Control}

\newacronym{stance}{STANCE}{\textbf{S}oft \textbf{T}errain \textbf{A}daptation a\textbf{n}d \textbf{C}ompliance \textbf{E}stimation}

\newacronym{wbopt}{WBOpt}{Whole Body Optimization}

\newacronym{hc}{HC}{Hunt and Crossley's}
\newacronym{kv}{KV}{Kelvin-Voigt's}

\newacronym{wllsr}{WLLSR}{Weighted Linear Least Squared Regression}

\newacronym{mae}{MAE}{Mean Absolute Tracking Error}

\newacronym{ode}{ODE}{Open Dynamics Engine}
\newacronym{lip}{LIP}{Linear Inverted Pendulum}
\newacronym{srbd}{SRBD}{Single Rigid Bidy Dynamics}

\newcommand{\Rnum}{\mathbb{R}} 

\newcommand{\vect}[1]{\mathbf{#1}} 

\DeclareMathOperator*{\argmin}{\arg\!\min}				
\newcommand{\mat}[1]{\ensuremath{\begin{bmatrix}#1\end{bmatrix}}}	
\newcommand{\x}{\ensuremath{\times}}

\newcommand{\atandue}{\textrm{atan2}}
\newcommand{\leg}{\mathrm{leg}}

\newcommand{\eref}[1]{(\ref{#1})}

\newcommand\BibTeX{{\rmfamily B\kern-.05em \textsc{i\kern-.025em b}\kern-.08em
T\kern-.1667em\lower.7ex\hbox{E}\kern-.125emX}}

\makeatletter
\newcounter{definition*}
\newenvironment{definition*}[1][htb]
{\renewcommand{\ALG@name}{Definition}
	\let\c@algocf\c@megaalgorithm
	\begin{algorithm*}[#1]%
	}{\end{algorithm*}}
\makeatother

\makeatletter
\newcounter{definition}

\makeatother

\definecolor{sfahmi_blue}{RGB}{0.19,0.51,0.74}
\definecolor{LightBlue}{RGB}{0.4,0.4,1}

\begin{document}

\begin{frontmatter}

\title{ALPINE: a climbing robot for operations in mountain environments}
\author[1,7]{Michele Focchi\corref{cor1}}
\ead{michele.focchi@unitn.it}

\author[2]{Andrea Del Prete}
\ead{andrea.delprete@unitn.it}

\author[3]{Daniele Fontanelli}
\ead{daniele.fontanelli@unitn.it}

\author[4]{Marco Frego}
\ead{marco.frego@unibz.it}

\author[5]{Angelika Peer}
\ead{angelika.peer@unibz.it}

\author[6]{Luigi Palopoli}
\ead{luigi.palopoli@unitn.it}

\cortext[cor1]{Corresponding author}

\affiliation[1]{organization={Dipartimento di Ingegneria and Scienza dell'Informazione (DISI), University of Trento}, 
	addressline={via Sommarive 9},
	postcode={38123}, 
	city={Trento}, 
	country={Italy}}
\affiliation[2]{organization={Dipartimento di Ingegneria Industriale (DII), University of Trento}, 
	addressline={via Sommarive 9},
	postcode={38123}, 
	city={Trento}, 
	country={Italy}}
\affiliation[3]{organization={Dipartimento di Ingegneria Industriale (DII), University of Trento}, 
	addressline={via Sommarive 9},
	postcode={38123}, 
	city={Trento}, 
	country={Italy}}
\affiliation[4]{organization={Faculty of Engineering, Free University of Bozen-Bolzano}, 
	addressline={via Volta 13/A},
	postcode={39100}, 
	city={Bolzano-Bozen}, 
	country={Italy}}
\affiliation[5]{organization={Faculty of Engineering, Free University of Bozen-Bolzano}, 
	addressline={via Volta 13/A},
	postcode={39100}, 
	city={Bolzano-Bozen}, 
	country={Italy}}
\affiliation[6]{organization={Dipartimento di Ingegneria and Scienza dell'Informazione (DISI), University of Trento}, 
	addressline={via Sommarive 9},
	postcode={38123}, 
	city={Trento}, 
	country={Italy}}
 \affiliation[7]{organization={Dynamic Legged Systems, Istituto Italiano di Tecnologia (IIT), Genova}, 
	addressline={via San Quirico 19d},
	postcode={16163}, 
	city={Genova}, a
	country={Italy}}

\begin{abstract}
	Mountain slopes are perfect examples of harsh environments in which
	humans are required to perform difficult and dangerous operations
	such as removing unstable boulders, dangerous vegetation or
	deploying safety nets. A good replacement for human intervention can
	be offered by climbing robots. The different solutions existing in
	the literature are not up to the task for the difficulty of the
	requirements (navigation, heavy payloads, flexibility in the
	execution of the tasks). In this paper, we propose a robotic
	platform that can fill this gap. Our solution is based on a robot
	that hangs on ropes, and uses a retractable leg to jump away from
	the mountain walls. Our package of mechanical solutions, along with
	the algorithms developed for motion planning and control, delivers
	swift navigation on irregular and steep slopes, the possibility to
	overcome or travel around significant natural barriers, and the
	ability to carry heavy payloads and execute complex tasks.  In the
	paper, we give a full account of our main design and algorithmic
	choices and show the feasibility of the solution through a large
	number of physically simulated scenarios.
\end{abstract}

\journal{Robotics and Autonomous Systems}
\end{frontmatter}

\section*{Supplementary Material}
{
	\begin{itemize}
		\item Video of experimental results is available at: 
		\href{https://youtu.be/FqsREaoe-28}{\texttt{https://youtu.be/FqsREaoe-28}}
		\item Code available  at: \\
		\href{https://github.com/mfocchi/climbing_robots2}{https://github.com/mfocchi/climbing\_robots2} \\
		(with source code used to generate all the figures in Section \ref{sec:results}).
	\end{itemize}
}

\section{Introduction}
\label{sec:introduction}
Mountain environments are extremely vulnerable to climate change and
often subject to landslides, floods and avalanches. Such ruinous
events endanger a large number of economic activities (first and
foremost tourism and agriculture) and put at risk the very survival of
small towns and villages.
%
%
Any realistic strategy to counter these risks is necessarily based on
constant monitoring and on regular maintenance operations on the
mountains slopes. Such activities include detaching from the mountain
walls dangerous boulders (a.k.a. scaling), loosing potentially
unstable shrubs and bushes, and deploying landslide protection
networks. The interested areas are often difficult to reach and
maintenance activities are typically performed manually by highly
trained human operators. These tasks are challenging and inherently
unsafe due to the presence of unstable rocks in the operation area.
As it frequently happens, a dangerous human activity in harsh
scenarios provides strong motivations for the development
of \emph{ad-hoc} robotics solutions, which in our case take the shape
of \emph{climbing robots}. A climbing robot is endowed with a package
of technical solutions that enable its navigation along difficult and
unstable mountain slopes and the execution of complex tasks, such as
rock  stability assessment, setting anchors in the rock, and scaling loose or dangerous boulders.  

\noindent
{\bf Related work.}
The ability of climbing robots to move  on vertical
surfaces~\cite{Nishi1986} naturally discloses a wide range of opportunities in
application areas such as
%
glass cleaning of tall buildings, pipeline maintenance or
infrastructure inspection, such as bridges.
In this regard, \cite{potenza2020robotics} evaluates various solutions to the problem 
of automating bridge inspection and maintenance tasks using robotic systems.
Compared to the application of flying robots for inspection
tasks~\cite{seo2018drone}, the use of climbing robots holds the
promise of longer mission durations and more accurate operations.

%
%
%
%
%
A first family of robots in this context are walking climbing robots
that are often bio-inspired~\cite{ji2018bio} and have been popular in
the last three decades.  An example is the Stickybot~\cite{stickybotIII}, a
gecko-inspired robot with adhesive structure under its toes to hold
itself on any kind of surface.  Others again took inspiration from the
flexible claws of wasps or flies~\cite{ji2018bio}.
Such wall-crawling solutions, although fascinating in their design principles, have not made
inroads into the market. Their main limitation is the risk of
accidental falls, possibly caused 
by strong wind and/or by the surface condition (e.g., the feet could slip away
from the slope in conditions of wet and irregular terrain).
The same problems determine strong limits on the payload that these robots can carry.

A different family of solutions is based on hybrid flying/climbing
robots, i.e., robots that can fly and that, at the same time can land
on, adhere to and ascend vertical surfaces. An example of this kind is
Scamp~\cite{Pope2017}, which uses a propeller to stick to the
wall. 
Caros (Climbing Aerial RObot System)~\cite{Myeong2015} is a fairly
conventional quadrotor equipped with four wheels and capable of
transitioning from the ground to the wall. Its rotors are
tilt-controlled, hence their thrust is used to generate aerodynamic
adhesion to stick to the wall.  The limitation is that the robot can
only ``slide'' on the surface and cannot overcome obstacles.
Hybrid propeller-wheeled wall-climbing robots achieve vertical locomotion with wheels  by pushing
the robot body against the wall using the thrust force of the propellers \cite{wall1, wall2}.
Because they can maintain a constant distance to the wall (differently from UAVs), 
they can capture high quality images of structures for failure/ defect detection. This simplifies 
the estimation of crack length and width \cite{crack_detection}.
They can also perform periodic  infrastructure health monitoring by performing a hammering test \cite{hammering}.
However, the reduced payload is a strong limitation for the execution of
different maintenance tasks.

%
A third family is given by climbing robots attached to ropes. They are
developed for different reasons, ranging from automated cameras
hanging over a stadium for dynamic recordings~\cite{qian:2018}, to
windows cleaning robots, cranes and other support
systems~\cite{korayem:2014,zhang:2018}, rovers for cave explorations
on other planets~\cite{Newdick:2023}. The main differences are the
positions of the anchor points where the ropes are fixed, which
depends on the application. The reachable space of the robot is
defined by the number and the position of the attachment points of the
cables.  Most results involve fixed extrema for all ropes (one or
more)~\cite{lahouar:2009}, and the control is done with non standard
optimisation techniques, that combine optimal control or MPC and
simplification steps like linearisation, because of the presence of
kinematic loops and high nonlinearities in the model.  On the other
hand, if one end of the ropes is free, the robot will swing and other positioning
methods should be envisaged: in fact, the tension is not bilateral.
Examples of this family of solution are given by robots hanging on
ropes~\cite{glacier_climbing_robot, bladebug,aerone} (hanging robots)
and are credited with the potential to address the problem of payload
limits.  In addition, a robot hanging on ropes can potentially explore
large areas, with limited power consumption and with a remarkable
speed.  An example of a hanging robots with high technology readiness
level are BladeBUG \cite{bladebug} and Aerone \cite{aerone} robots,
which can inspect, diagnose, repair, and clean wind turbine blades.
Although well engineered, these robots can only operate on surfaces with a regular
geometry (e.g., wind turbine blades). Furthermore, they moves with a low speed,
due to the crawling motion along the blades. 
Another example is the Axel/DuAxel robot~\cite{axelRobot}a rover for space exploration that moves by reeling/unreeling  its built-in tether,  lowering itself down almost any type of terrain, which has been optimised for resisting to the very cold temperatures of the Moon or Mars and their sand traps. Finally, the aerial robot \cite{polishCrevassExploration} is a tethered bicopter with horizontal propellers, that by swinging manoeuvres is able to navigate glacial-inspired scenarios, where there are  obstacles. 

%
%
Other hanging robot solutions such as the novel bio-inspired dragline
locomotion~\cite{Wang2014} or  CLIO~\cite{focchi23icra}, can reach high navigation speed
using an actuated winding/releasing mechanism, which is a decisive advantage for applications requiring
a prompt intervention over
different solutions such as climbing robots that use
sticky pads and gaits to climb up/down~\cite{Kim2008,Riskin2009, lemur3, ROCR}.

An efficient locomotion mechanism is certainly key for a climbing
robot. However, other aspects are gaining an equal level of
importance. The peculiar nature of the activities required to climbing
robots, makes them a difficult fit for full autonomy. Existing
solutions require a tight supervision by humans in visual contact from
helicopters or special structures, such as telescopic platforms or
scaffolds. The difficulty and the costs of this type of human
intervention calls for an increased level of autonomy for climbing
robots.
Two key aspects for increasing autonomy are control and motion planning. 
%
%
%
Some approaches involved numerical optimisation applied to
lower-dimensional template models~\cite{Hoffman2021} or hierarchical
whole-body controllers to plan the motion  of the flying
base~\cite{Coelho2021}.  In particular, \cite{Hoffman2021} optimises a
multi-jump trajectory where the contact locations and the jump time
are free variables, thus overcoming a gap obstacle.  However, no
physical simulation of the approach was provided.

In our previous paper~\cite{focchi23icra}, we showed an optimisation
of a jump trajectory with a \textit{single} rope.
%
A rope-based climbing robot moving with jumps is somewhat close to the
Salto-1P jumping robot~\cite{Haldane2017}, but the key feature of
jumping with a rope lies in the ability to address terrains of high
inclination (up to vertical).  In legged robots, jumping motions on
the ground are usually synthesised by means of sophisticated numerical
optimisation techniques~\cite{Nguyen2019,Chignoli2021}.  For example,
Ding et al.~\cite{Ding2021a} consider the dynamics up to the actuation
level and employ direct optimal control approaches to derive a
trajectory for the centre of mass that satisfies constraints and
avoids obstacles (e.g. a gap on the wall) and mixed-integer
programming approaches in the case that the contact location is not
known a priori.  Since the main drawback of these solutions is the
exponential computation time, the use of approximations becomes
mandatory. 
As an example, Jiang et al. \cite{Jiang2022} and Grandia et
al.~\cite{ Grandia2022} propose using convex polyhedrons to
approximate the terrain and constrain the reachability of the robot
feet.  

\begin{figure}
	\centering
	\includegraphics[width=0.5\dimexpr\scalefactor\textwidth]{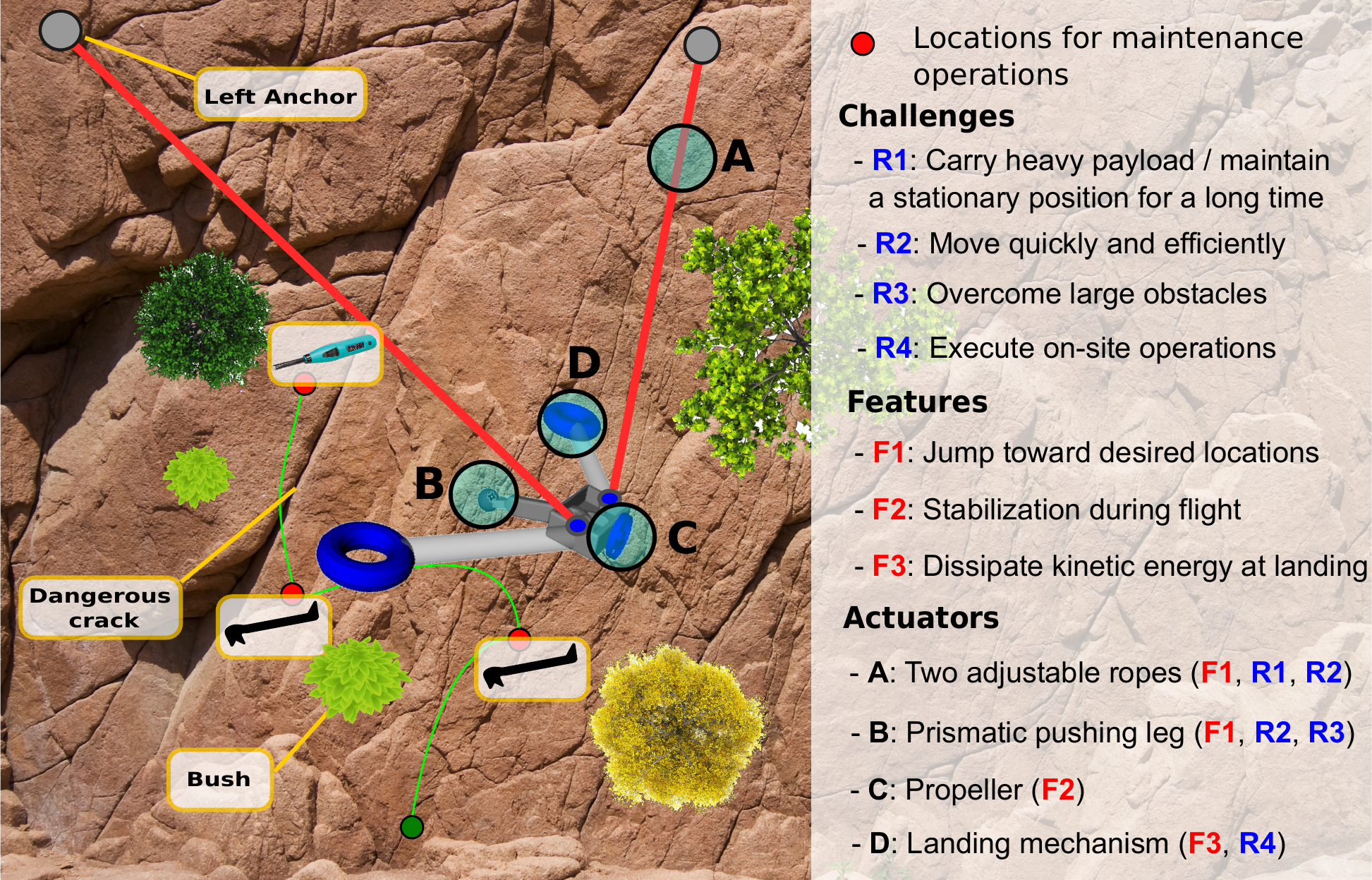}
	\caption{\small Use cases and challenges for tethered climbing robots and main capabilities of the ALPINE platform.}
	\label{fig:gazebo_simulation}
\end{figure}

\noindent
\noindent
{\bf Example Use Case and Requirements.}  
The system requirements are best illustrated through a realistic use case related to a typical maintenance operation in a mountainous environment (see Figure~\ref{fig:gazebo_simulation}).  
In this example, the system's objective is to monitor the condition of a steep mountain wall (the slope could be $0.1$ radians from the vertical line). Reaching the site is a highly challenging task for humans, requiring advanced mountaineering techniques. Not only are the rock walls steep and difficult to climb, but they are also obstructed by obstacles up to $2$ m in size, such as bushes, rock protrusions, and cracks. Replacing humans with a robot necessitates replicating the same navigation capabilities in these harsh conditions.  

\begin{figure}[t]
\centering
\includegraphics[width=0.8\dimexpr\scalefactor\columnwidth]{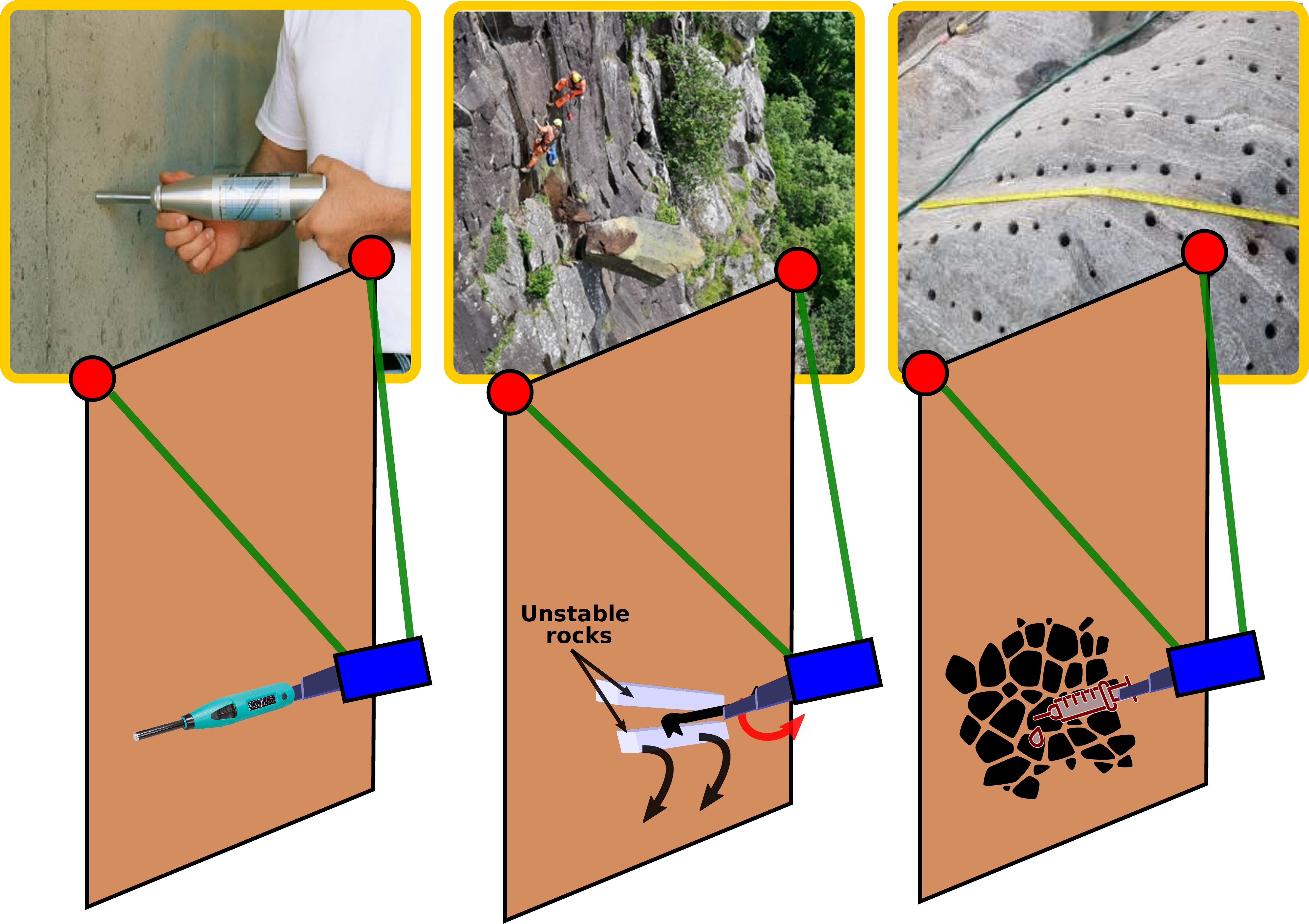}
\caption{\small Examples of maintenance operations: rock inspection with rebound hammering (left),  debris removal (middle), injection of demolition resin  (right).}
\label{fig:applications}
\end{figure}

In the outlined scenario, we assume that the robot has access to a 3D map (e.g., acquired by a drone), but it still requires onboard sensing capabilities to localise itself within the map and to measure its distance from the wall. Moreover, even after reaching the intervention site, the robot must be capable of executing potentially complex operations. For instance, assessing the stability of a boulder might involve hammering and conducting non-destructive testing. Typical equipment for hammering operations \cite{smidthhammer} is designed to be portable and lightweight, but the duration of such tests can be substantial (in the order of $30$ minutes).  
To function effectively in challenging scenarios such as the one described, the robot must meet the following requirements:  

\textbf{R1} - Carry a payload (e.g., in the order of kilograms) and maintain a stationary position for extended periods without consuming energy (e.g., to perform inspection or maintenance operations such as hammering; see Fig.~\ref{fig:applications}).  

\textbf{R2} - Move quickly and efficiently (e.g., in the order of seconds) with sufficient accuracy (e.g., in the order of centimetres) along very steep, vertical, or slanted mountain slopes.  

\textbf{R3} - Traverse irregular surfaces, overcoming obstacles (e.g., in the order of metres) such as bushes and rock formations.  

\textbf{R4} - Autonomously or semi-autonomously execute a wide range of on-site operations.

\noindent
{\bf Mapping requirements into design choices.}
%
The system requirements had a significant influence on the design choices, as detailed below:
\begin{itemize}
    \item {\bf R1:} ALPINE operates using two ropes, enabling it to carry heavy loads and remain stationary with minimal energy consumption using brakes; the anchor point is designed in such a way that, for the intended motion of the ropes, there is not interference nor friction with other parts of the anchor (e.g. the anchor ring is protruding from the rock).
    \item {\bf R2:} ALPINE coordinates the two ropes and a prismatic leg to quickly push itself away from its resting position for navigation. During jumps and flight, the ropes are independently wound or unwound to control its trajectory, while an auxiliary rotor stabilises the flight. This approach, similarly to the Salto-1p robot~\cite{Haldane2017}, demonstrates energy efficiency and high performance in terms of travel time.
    \item {\bf R3:} The paper presents motion planning and control strategies for navigating while overcoming obstacles (see Figure~\ref{fig:gazebo_simulation}). A simplified model of the system allows for efficient numerical solutions to the multi-jump planning and control problem.
    \item {\bf R4:} A landing mechanism dissipates the excess of kinetic energy and stabilises the robot on the wall during task execution. Combined with appropriately designed planning and control components, this provides a high degree of operational flexibility.
\end{itemize}
%
%
%

%
\noindent
{\bf Scientific challenges.}
The first scientific challenge of the ALPINE robot lies in its
under-actuation:  it is impossible
to \emph{fully} control the robot's \gls{com} when not in contact with
the wall. Second, one of the actuators (the leg) operates in an
impulsive way, while the ropes can only operate in the pulling
direction (unilateral actuation). The problem is partially alleviated
by the auxiliary propeller during the flight phase, but the resulting
system dynamics is hybrid (the continuous evolution is interspersed
with discrete changes).  As regards motion planning, since the motion
of the robot depends on both the impulse exerted on the wall  and 
the winding/unwinding of the ropes, a successful strategy should consider the combined effects of
under-actuation, rope constraints, the actuator limits and the contact
interaction (i.e., friction). Whilst numeric optimisation is certainly
a powerful tool~\cite{Nguyen2019,Ding2020}, the combination of these
factors and  hybrid dynamics, makes the problem tractability far
from obvious.\\
\noindent
{\bf Paper Contribution and Summary.}

The contributions of the paper can be summarised as follows.
\begin{itemize}

\item The conceptual design of the jumping robot platform ALPINE
	(Section~\ref{sec:model});

\item A reduced-order model to simplify the solution of the optimal
control strategies (Section~\ref{sec:model});

\item A static analysis to evaluate the maximum value of operating
 forces that the system is able to withstand in the execution of its
 tasks (Section~\ref{sec:static});

\item A computationally efficient planning algorithm to generate a
 jump to reach desired targets while overcoming obstacles
 (Section~\ref{sec:motion_planning});

\item A motion control strategy to track the reference trajectories
with high landing accuracy on approximately locally flat surfaces
(Section~\ref{sec:fly_motion_control}).

\end{itemize}
The efficacy of the system design, as well as of motion planning and
control, is tested in a large set of physical simulations
(Section~\ref{sec:results}), while the limits and the future work
directions are summarised in Section~\ref{sec:conclusion}.

\noindent
{\bf Improvements over the preliminary version.}
The paper builds on a preliminary idea introduced in a conference paper~\cite{focchi23icra} but presents a significant  package of innovations that greatly enhance the platform’s effectiveness and operational capabilities. Key advancements include:
\begin{itemize}
    \item {\bf Enhanced actuation mechanism}: Introducing two ropes and a propeller (instead of one rope) expands the workspace and improves stability during flight.
    \item {\bf Advanced Control Strategy}: Transitioning to a Model Predictive Controller enables obstacle avoidance and represents a substantial improvement over the original Proportional-Derivative controller paired with a basic motion planning mechanism.
    \item {\bf Improved landing mechanism}: The new design addresses the limitations of the original retractable leg by incorporating a more realistic and robust landing system.
    \item {\bf Static equilibrium analysis}:  This addition provides a clear evaluation of the device's operational capabilities.
    \end{itemize}
Extensive simulations validate the realism and practicality of these innovations.

\nomenclature[01]{$	n 	 				$}{Number of \gls{dofs} of the system}        

\nomenclature[03]{$	\vect{p}   			$}{Position of the robot \gls{com} (reduced-order model)}		 
\nomenclature[04]{$	\vect{p}_{a,i}    	$}{Position of $i$-th anchor}						 
\nomenclature[05]{$	\vect{p}_{h,i}	$}{Position of $i$-th  hoist}                       
\nomenclature[06]{$	\vect{p}_{l,i}	$}{Position of the $i$-th landing wheel } 
\nomenclature[07]{$	\vect{J}_{h,i}	$}{Jacobian of the $i$-th  hoist location}          
\nomenclature[08]{$	\vect{J}_r			$}{Jacobian mapping rope forces/velocities on the \gls{com}}  
\nomenclature[09]{$	\vect{J_c}			$}{Jacobian of the prismatic leg's foot contact point}  
\nomenclature[10]{$	  \vect{M} 			$}{Inertia matrix}							     
\nomenclature[11]{$	  \vect{h} 			$}{Bias terms (Centrifugal,   Coriolis and Gravity)} 
\nomenclature[12]{$   \boldsymbol{\tau}_a$}{Actuated generalized forces} 				 
\nomenclature[13]{$	\mu 				$}{Friction coefficient} 
\nomenclature[14]{$    d_h  				$}{Distance of hoist positions on top the base link}	  
\nomenclature[15]{$	d_a 				$}{Distance between anchor points}  
\nomenclature[16]{$	d_b 				$}{Distance between landing feet}   
\nomenclature[17]{$	d_w 				$}{Distance of landing feet w.r.t. base frame (along  $X$)} 
\nomenclature[18]{$    \psi				$}{Reduced-order model state: angle of the ropes w.r.t. to the vertical} 
\nomenclature[19]{$    l_1 				$}{Reduced-order model state: length of the left rope } 
\nomenclature[20]{$    l_2 				$}{Reduced-order model state: length of the right rope} 
\nomenclature[21]{$\vect{A}_d,\vect{b}_d $}{Reduced order model dynamic terms} 
\nomenclature[22]{$\vect{A}_p,\vect{b}_p $}{Polytope constraints matrix}  
\nomenclature[23]{$	f_{r,\max/\min}		    $}{Maximum/minimum rope force}  
\nomenclature[24]{$  	f_{\leg, \max}	$}{Maximum (normal) leg force}  
\nomenclature[25]{$	\vect{n}_{\perp}    $}{unit vector perpendicular to the rope plane} 
\nomenclature[26]{$\vect{n}_{\parallel}  $}{unit vector passing through the anchor points} 
\nomenclature[27]{$\vect{n}_c			$}{normal of the surface in contact with  the  prismatic leg's foot}  
\nomenclature[30]{$    N                 $}{\gls{nlp} Discretisation steps}   
\nomenclature[31]{$    N_\text{mpc}	    $}{\gls{mpc} Discretisation steps}   
\nomenclature[32]{$    dt_\text{sim}     $}{Simulation 	time interval}     
\nomenclature[33]{$    dt				$}{Discretisation time interval for the \gls{nlp} optimisation} 
\nomenclature[34]{$	dt_\text{mpc}		$}{Discretisation time interval for the \gls{mpc} optimisation} 
\nomenclature[35]{$ 	t_{\text{th}}  		$}{Thrust impulse duration}	 
\nomenclature[36]{$     w_{s}            $}{\gls{nlp} Smoothing weight }     
\nomenclature[37]{$     w_{hw}           $}{\gls{nlp} Hoist work weight }     
\nomenclature[38]{$     w_{i}            $}{\gls{nlp} Impulse work weight }  
\nomenclature[39]{$w_{p/pf/u,\text{mpc}} $}{\gls{mpc}  weights}     
\nomenclature[40]{$\boldsymbol{\delta}_i $}{Impulsive disturbance}     
\nomenclature[41]{$\boldsymbol{\delta}_c $}{Constant disturbance}          
\nomenclature[42]{$K_L, D_L 				$}{Landing strategy impedance parameters}    
\nomenclature[43]{$v_{l,l},v_{l,r} $}{Scalar velocity of the center wheels (parallel to the wall)}
\nomenclature[44]{$v_{r,l}$,$v_{r,r}$}{Scalar rope speed along the rope axes}
\nomenclature[45]{$R_w$}{Landing wheels radius}
\nomenclature[46]{$  	f_{\leg}	$}{Leg impulse force} 
\nomenclature[47]{$	\vect{a}_{r,i}    	$}{$i$-th rope axis}	
\nomenclature[48]{$	f_{r,i}    	$}{$i$-th rope force}
\nomenclature[49]{$   f_p$}{Propeller force} 	
\begin{small} 
\printnomenclature
 Unless specified, all vectors are expressed in an inertial frame $\mathcal{W}$ frame (attached to the left anchor). Vectors and matrices are highlighted in bold. 
\end{small}
\section{Robot Modeling}
\label{sec:model}
As mentioned above, a robot hanging on a rope has the ability
to preserve energy, when static, by simply engaging the brakes in
the hoist.
%
However, a robot hanging on a single rope~\cite{focchi23icra,
  Hoffman2021} has severe limitations in terms of possible lateral
motions, and hence reachable locations. Additionally, the lateral pull
of the rope makes the system unstable when static, limiting its
ability to execute tasks.  A possible way to tackle these issues is to
have an additional prismatic joint (a slider) that enables the motion
of the rope anchor point.
While easy to control, this solution requires that the slider be mounted
on a path clear from obstacles, which is difficult in environments where
rock protrusions and bushes are  common.

%
With these limitations in mind, we opted for a different design based
on a second rope attached to an additional \textit{fixed} anchor on
the wall. Both ropes can be \textit{independently} wound/unwound by
means of hoist motors (see Fig.~\ref{fig:3dmodel2anchors_propellers}).
Deploying the two anchors independently removes the limitation of
having a clear path for the slider, while the mechanical design is
significantly simplified. Another advantage is an increased robustness
to disturbances coming from the operations, as detailed in
Section~\ref{sec:static}. The price to pay is a higher control
complexity, since the release/winding of the ropes needs to be
accurately coordinated.
\subsection{Full robot model}
\label{sec:full_model}

\begin{figure}[t]
  \centering
  \includegraphics[width=0.8\dimexpr\scalefactor\columnwidth]{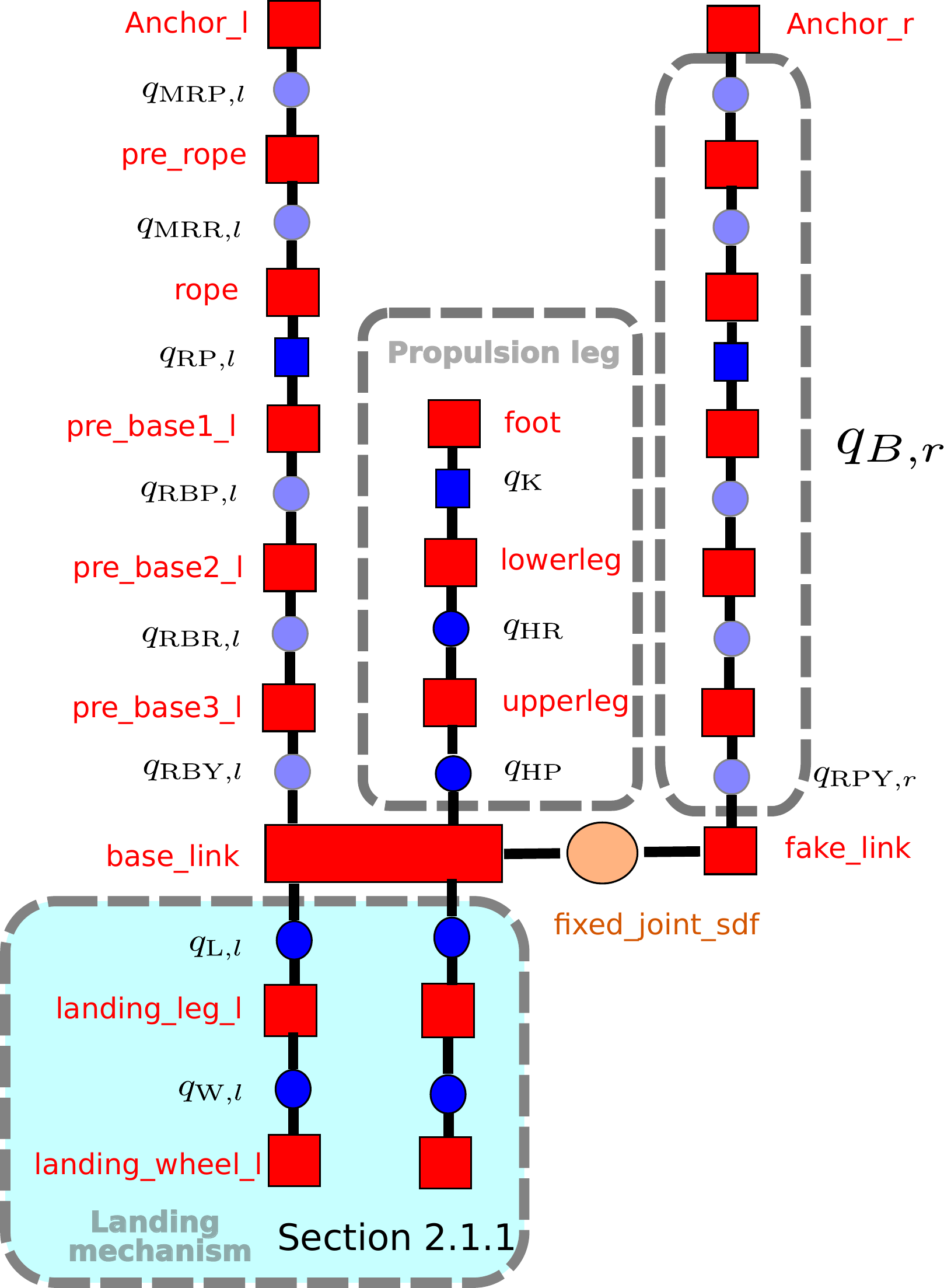}
  \caption{Topology of the joints for the full detail model. Links are
    red boxes and joints blue circles (revolute) or blue boxes
    (prismatic). Shaded joints are the passive ones, not shaded ones
    are the active ones. }
  \label{fig:topology}
\end{figure}

We model the robot as 3 kinematic chains branching from the base link
(see Fig.~\ref{fig:topology}).
One chain represents the propulsion mechanism (a 3-DoF prismatic
leg)~\cite{focchi23icra}: at the extreme of the leg there is a
point-like foot. The leg is also endowed with two \textit{adjacent}
rotational joints, called hip pitch ($q_{HP}$, rotating about the base $Y$
axis) and hip roll ($q_{HR}$, rotating about the base $X$ axis).  These
joints are needed to align the leg to the \textit{thrusting} impulse,
so as to avoid the generation of centroidal moments that would pivot
the robot around the rope axis.  A prismatic knee joint ($q_{K}$) is
used to generate the \textit{thrusting} impulsive force.  The landing
mechanism is represented by two additional rotational joints that move
the landing links (see Fig.~\ref{fig:landing_propellers}) with two
passive wheels at the extreme as described in more detail in
Section~\ref{sec:landing}.

\begin{figure}[t]
  \centering
  \includegraphics[width=0.6\dimexpr\scalefactor\columnwidth]{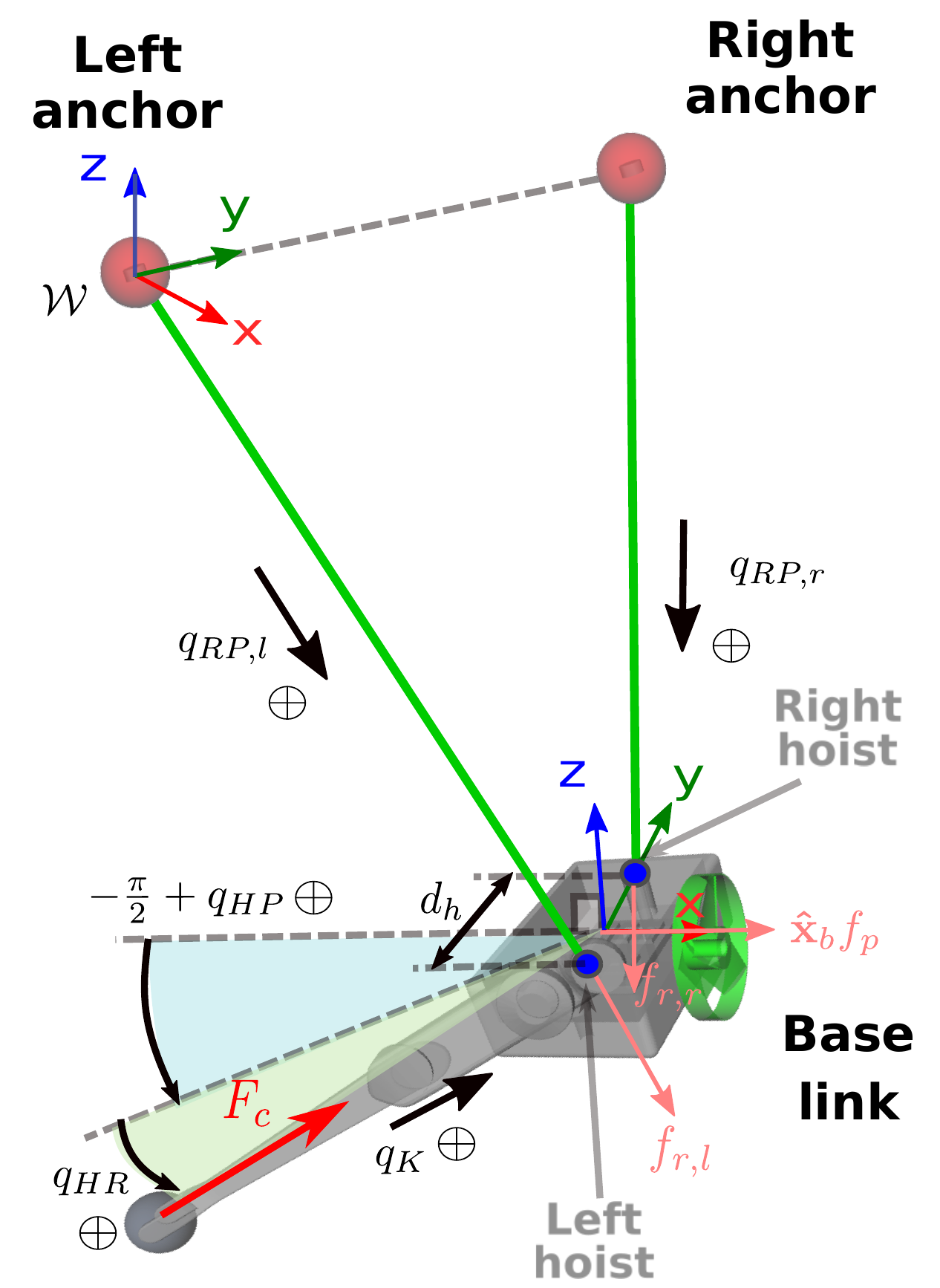}
  \caption{Kinematic model of the ALPINE robot with two ropes
    (standard definitions).  A propeller is mounted on the rear of the robot. In pink are depicted all the actuation forces. The inertial ($\mathcal{W}$) frame is
    attached to the left anchor frame.}
  \label{fig:3dmodel2anchors_propellers}
\end{figure}

The other two kinematic chains model the two ropes.  To host the
hoist motors, the attachment points of the ropes are mounted with an offset $d_h/2$ w.r.t. the base frame (see Fig.~\ref{fig:3dmodel2anchors_propellers}).  
We model the attachment between each anchor
and the corresponding rope by 2 \textit{passive} (rotational) joints.
Each rope can be seen as an \textit{actuated} prismatic joint
($q_{RP,l}$ or $q_{RP,r}$),  followed by 3 \textit{passive} rotational joints
to model the connection between the rope and the \textit{base link}.
The described mounting choice is indeed equivalent to allocating $3$
joints at the anchor point and $2$ at the base.  However, to avoid a
redundant representation, there must be only one passive rotational
joint aligned with the rope.  To increase the robot controllability
(e.g. capability to apply forces on the base) along the direction
perpendicular to the ropes plane, we added a propeller mounted on the
back of the base link, as depicted in
Fig.~\ref{fig:3dmodel2anchors_propellers}.  This brings several
advantages. 1) It allows the robot to reject disturbances and reduce
tracking errors (during the flight) in the direction perpendicular to
the ropes plane, enhancing controllability. 2) It increases the
maximum force the robot can withstand during operations without losing
contact. This is particularly relevant when dealing with harder rocks:
by activating the propellers to provide additional push against the
walls, we can increase the force margin available for more demanding operations, such
as drilling.  3) It enables the control of the robot orientation both
in contact and 4) during the flight, by actuating the propellers in a
differential way. In this paper, we will focus on exploiting only
the disturbance rejection feature 1), and the reorientation 3)
leaving the other two features for future works.  Theoretically, the
propeller itself could be used in place of the prismatic leg to generate the
push from the wall. However, the explosive motion required to achieve
high acceleration in a short time interval is difficult to achieve
with a propeller. Therefore, we chose to rely on a  prismatic actuator specifically
designed for this purpose. In this design, we
preferred to keep the propeller size contained and use it only for
corrections of the deviations from the desired trajectory due to
environmental disturbances (see
Section \ref{sec:mpc}).  The differential commanding of a couple of
propellers could enable the horizontal alignment of the leg to the
\textit{thrusting} impulse, removing the need for the $q_{HR}$
joint. We showcase in the accompanying video the solutions with both $q_{HR}$ and $q_{HP}$ joints and with propellers plus $q_{HP}$ joint.

Neglecting the $4$ joints of the landing mechanism (see Section \ref{sec:landing}) and the propeller,
the total number of \gls{dofs} is $n=15$, represented by the
configuration vector $\vect{q} \in \Rnum^{15}$. Ten of these joints
are relative to the attachment of the ropes to the hoist and the anchor and are
passive.  Fig.~\ref{fig:topology} illustrates the joint definitions:
$MRX$ (Mountain Rope X = Pitch/Roll), $RP$ (Rope Prismatic) and $RBX$
(Rope Base X = Roll/Pitch/Yaw).  By stacking the different joint
variables related to the left rope, we obtain the vector
$\vect{q}_{B,l} = \mat{q_{MRP} & q_{MRR}& q_{RP} & q_{RBP} & q_{RBR}&
  q_{RBY}}^T$.  We can repeat the same for the right rope stacking the
joint variable into the vector $\vect{q}_{B,r}$ and come up with the
definition of the joint state
$\vect{q} = \mat{\vect{q}_{B,l}^T & \vect{q}_{B,r}^T & q_{HP} & q_{HR}
  & q_{K}}^T$, where additionally we have $HX$ (Hip X = roll/pitch)
and $K$ (Knee) joints.  The dynamic equation of motion is subject to a
holonomic constraint because the two attachment points
$\vect{p}_{h,l}(\vect{q})$, $\vect{p}_{h,r}(\vect{q})$ should maintain a fixed distance $d_h$ among them, i.e.
\begin{align}
  &\Vert \vect{p}_{h,l}(\vect{q})  - \vect{p}_{h,r}(\vect{q})
    \Vert^2 = d_h^2 .
    \label{eq:holonomic_constraint}
\end{align}
This description creates a kinematic loop represented by the trapezoid
having the two ropes as edges. 
%
%
The full dynamic equation is reported here just for reference
{\small
\begin{align}
  \begin{cases}
    &\vect{M} (\vect{q}) \vect{\ddot{q}} +
    \vect{h}(\vect{q},\vect{\dot{q}}) = \mat{\vect{0}_{10 \times 1} \\
      \boldsymbol{\tau}_a} + \vect{J}_c(\vect{q})^T \vect{f}_c + \vect{J}_p(\vect{q})^T f_p , \\
    &\vect{A}(\vect{q})\dot{\vect{q}} = \vect{0}, \\
    &\tau_{RP,l} \geq 0,\\
    &\tau_{RP,r} \geq 0,
  \end{cases}
      \label{eq:full_dyn}
\end{align}}
where $\vect{M} \in \Rnum^{n \times n}$ is the inertia matrix,
$\vect{h} \in \Rnum^n$ represents the bias terms (Centrifugal,
Coriolis and Gravity), and $\vect{J_c} \in \Rnum^{3 \times n} $ is the
Jacobian relative to the prismatic leg contact point (foot) that maps the contact force
$\vect{f}_c \in \Rnum^3$ into the generalised coordinate space. 
$f_p \in \Rnum$ is a scalar and represents the magnitude of the propeller force that is mapped into the robot dynamics
through the transpose of   $\vect{J}_p(\vect{q}) = \vect{\hat{x}}_b^T \vect{J}_b(\vect{q})\in \Rnum^{1 \times n}$,
where $\vect{\hat{x}}_b$ is the $X$ axis of the base link
and  $\vect{J}_b(\vect{q})= 
\mat{\frac{\partial \vect{p}(\vect{q})}{\partial \vect{q}}} \in \Rnum^{3 \times n}$
is the Jacobian of the robot base. 
The $\vect{A}(\vect{q})
\dot{\vect{q}} = \vect{0} $ constraint is \eqref{eq:holonomic_constraint} rewritten at the velocity level.
The force/torque variable of the the actuated joints are grouped into
the vector
$\boldsymbol{\tau}_a = \mat{ \boldsymbol{\tau}_{RP,l} &
  \boldsymbol{\tau}_{RP,r} & \boldsymbol{\tau}_{\text{leg}}}^T \in
\Rnum^5$. We will employ the right hand side of~\eqref{eq:full_dyn} to
express the mapping of the contact force and of the propeller force
into joint torques.

\subsubsection{Landing Mechanism}
\label{sec:landing}

\begin{figure}[t]
  \centering
  \includegraphics[width=0.7\dimexpr\scalefactor\columnwidth]{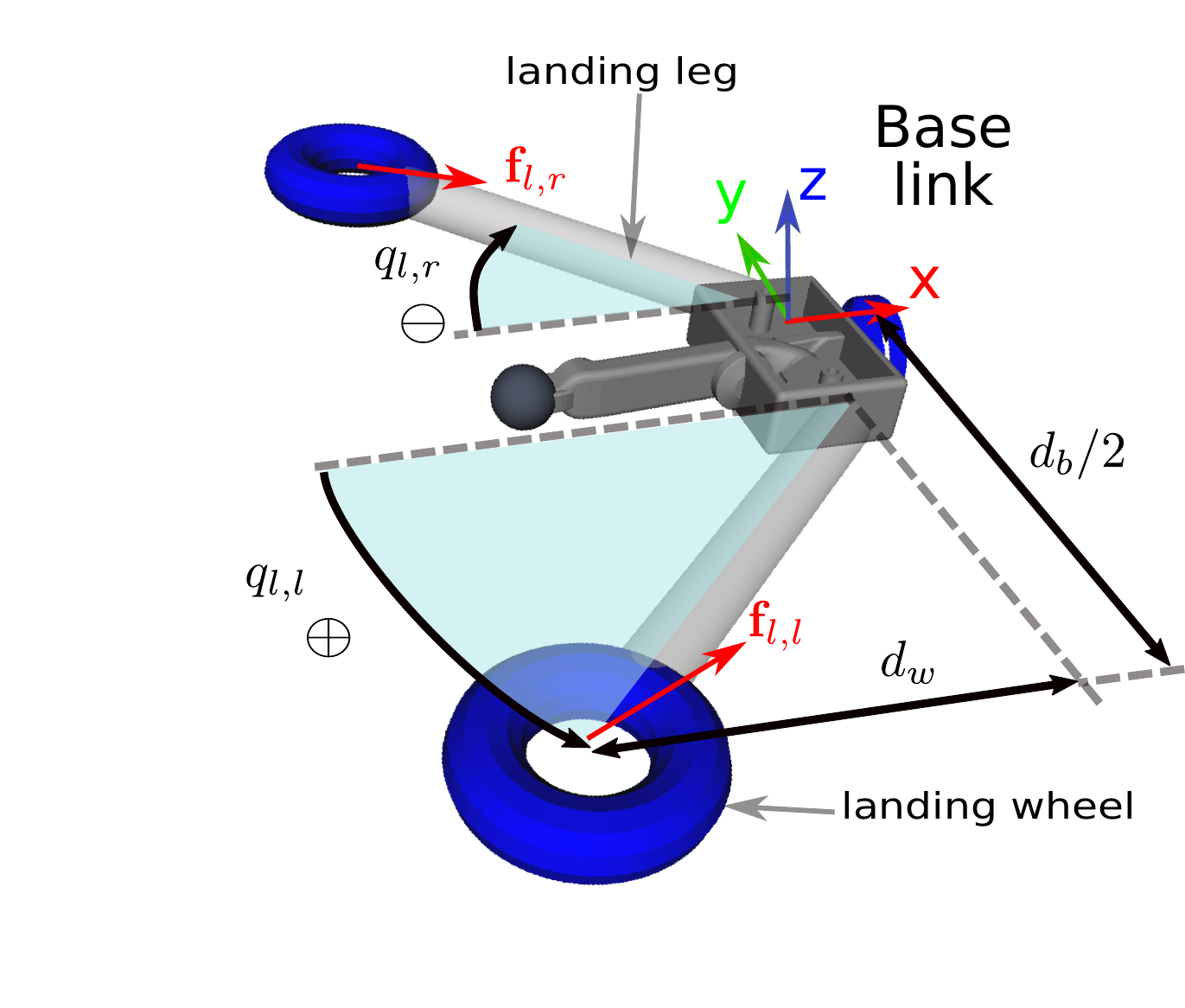} 
  \caption{Overview of the landing mechanism, with joint and variable
    definitions. $d_w$ and $d_b$ are the wall clearance and distance between landing wheels. Passive wheels are attached to the extremes of the
    landing legs.  }
  \label{fig:landing_propellers}
\end{figure}

The robot in Fig.~\ref{fig:3dmodel2anchors_propellers} would not be able to
self-stabilise on the wall with a single leg, therefore we designed a
\textit{landing} mechanism, depicted in
Fig.~\ref{fig:landing_propellers}, formed  by two additional rotational
joints $q_{L,l}$ and $q_{L,r}$ located on the sides of the base link allowing to activate two landing legs.
Moreover, we added two wheels at the tips
of the landing legs, which can move \textit{laterally} during the
contact and, hence, avoid the generation of \textit{internal} forces (i.e., parallel to the wall).

\subsection{Reduced-order model with minimal representation}
\label{sec:reduced_order_model}
The high number of states and the constraint in~\eqref{eq:full_dyn}
makes the full dynamics hardly tractable for control design.  For this
reason, we derive a lower dimensional (reduced-order) model with only
$3$ \gls{dofs} that captures the dominant dynamics of the system along
with the holonomic kinematic
constraint~\eqref{eq:holonomic_constraint}.  To this end, we make the
following simplifying assumptions: 1) the mass is entirely
concentrated in the body attached to the rope and we neglect the
angular dynamics and 
2) during the winding/rewinding of the ropes they remain completely
tight (i.e., they are not bending).

%
%

\begin{figure}
	\centering
  \includegraphics[width=0.68\dimexpr\scalefactor\columnwidth]{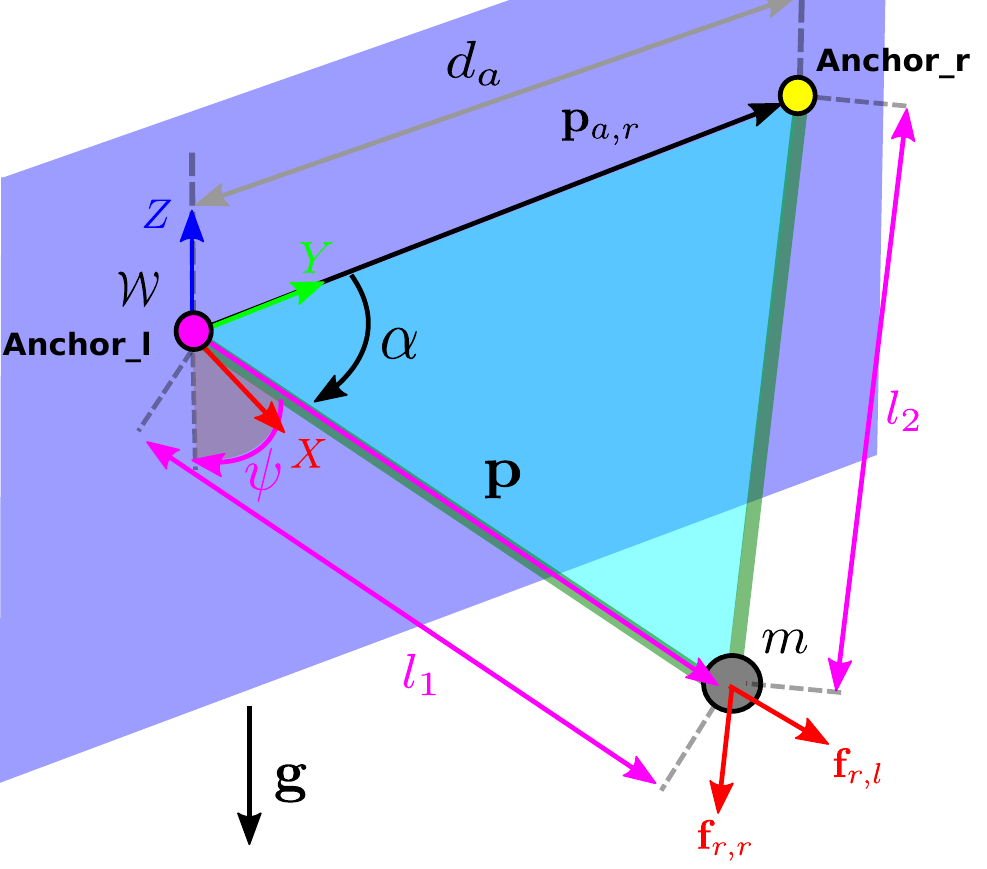}
  \includegraphics[width=0.28\dimexpr\scalefactor\columnwidth]{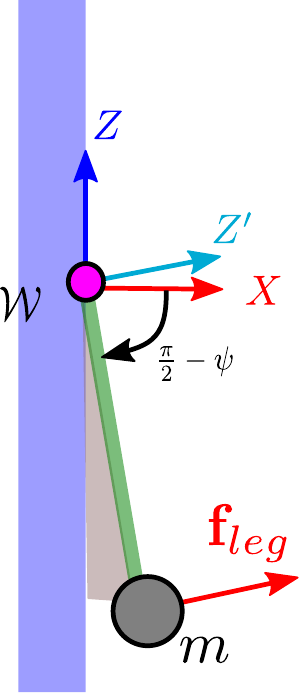}
  \caption{Reduced order model with two anchor points: standard
    definitions (left) and side view (right). The impulsive force $\vect{f}_{\text{leg}}$ is
    applied only when the leg is in contact.}
  \label{fig:2anchorsModel}
\end{figure}

In order 
to simplify the  control  design,
we approximate the rope attachment points as coincident and then choose a 
\textit{minimal} representation for the state with $3$ \gls{dofs}. Specifically,
the reduced state $\vect{q}_r$ is defined as $\vect{q}_r =\mat{ \psi & l_1 & l_2} \in \Rnum^3$, where $\psi$
is the angle formed from the ropes plane and the wall, and $l_1$,
$l_2$ the length of the left and right ropes,
respectively.
A geometric scheme of the reduced model with vector definitions is shown
in Fig.~\ref{fig:2anchorsModel}.
Assuming the inertial frame $\mathcal{W}$ attached to the
\textit{left} anchor point, the dynamics of the point $\vect{p}$
(where the robot mass is concentrated, by assumption) is defined by
the Newton Equation:
\begin{equation}
  m (\vect{\ddot{p}} - \vect{g}) = \underbrace{\vect{\hat{a}}_{r,l}
    f_{r,l}}_{\vect{f}_{r,l}} + \underbrace{\vect{\hat{a}}_{r,r}
    f_{r,r}}_{\vect{f}_{r,l}} + \vect{f}_{leg} + \vect{\hat{x}}_b f_p,
  \label{eq:newton}
\end{equation}
where $\vect{\hat{a}}_{r,i} = \frac{\vect{p} - \vect{p}_{a,i}}{\Vert
  \vect{p} - \vect{p}_{a,i} \Vert} \in \Rnum^3$ and
$f_{r,i} \in \Rnum$, with $i = \{r,l\}$, are the rope axes and the
magnitude of the exerted forces, respectively. 
The term $\vect{\hat{x}}_bf_p$ represents the propeller's force where
$\vect{\hat{x}}_b \in \Rnum^3$ is the base link $X$ unit axis which is perpendicular to
the ropes' plane.  \footnote{Differently from \eqref{eq:full_dyn},
  because we adopt a point mass assumption, the base link is always
  aligned with the ropes' plane
  (i.e. $\vect{n}_{\perp}=\vect{\hat{x}}_b$).}

The two anchor point positions are  given by $\vect{p}_{a,l}$ and
$\vect{p}_{a,r}$. The input
variables  are: 1) the rope forces 
$\vect{f}_{r,i}$ oriented along the rope axes; 2) an
impulsive pushing force $\mathbf{f}_{\text{leg}}$ (applied at $\vect{p}$) 
that the robot  generates
when in contact with the mountain  (see
Fig.~\ref{fig:2anchorsModel}).
%
Therefore, the expression of the forward kinematics for the position
$\vect{p}$ of the robot (origin of the base-link) as a function of
$\vect{q}_r$ is given by: \renewcommand{\arraystretch}{1.5}
%
%
\begin{equation}
  \vect{p}(\vect{q}_r) = 
  \mat{l_1 \sin(\psi) \sqrt{1 - \frac{(d_a^2 + l_1^2 - l_2^2)^2}{4 d_a^2 l_1^2}} \\	
    (d_a^2 + l_1^2 - l_2^2)/(2 d_a) \\
    -l_1 \cos(\psi) \sqrt{1 - \frac{(d_a^2 + l_1^2 - l_2^2)^2}{4 d_a^2l_1^2}}} ,
  \label{eq:fwd_kin_minimal}
\end{equation}
\renewcommand{\arraystretch}{1.0}\noindent
where the holonomic constraint~\eqref{eq:holonomic_constraint} has
been embedded (by construction) thanks to specific choice of  coordinates $\vect{q}_r$ and $d_a$ is the distance between the anchors. 
The dynamics of the point mass as defined in \eqref{eq:newton} is
given by the second derivative of~\eqref{eq:fwd_kin_minimal}, which
gives an implicit expression for the coupled derivatives
$\ddot{\psi}$, $\ddot{l_1}$ and $\ddot{l_2}$, thus leading 
to the following matrix form of~\eqref{eq:newton}
\begin{equation}
  m \underbrace{\left(\vect{A}_{d} \mat{\ddot{\psi}\\ \ddot{l}_1\\
        \ddot{l}_2} + \vect{b}_{d}\right) }_{\ddot{\vect{p}}}=
  \underbrace{ m \vect{g}+ \vect{f}_{leg}+ \vect{J}_r \mat{f_{r,l}
      \\f_{r,r}}  +  \vect{\hat{x}}_b f_p}_{\vect{f}_{tot}} ,
  \label{eq:simplified_2ropes_minimal_1}
\end{equation}
where $\vect{A}_{d}$ and $\vect{b}_{d}$ are
reported 
in the Appendix, and
$\vect{J}_r = \mat{\vect{\hat{a}}_{r,l} & \vect{\hat{a}}_{r,r} } \in
\Rnum^{3 \times 2}$.
Finally, we obtain:
\begin{equation}
  \mat{\ddot{\psi}\\ \ddot{l}_1\\ \ddot{l}_2} = \vect{A}_{d}^{-1}\left[ \frac{1}{m} \vect{f}_{tot} - \vect{b}_{d} \right] .
\label{eq:simplified_2ropes_minimal}
\end{equation}
Note that this model has a singularity when the term $\sin(\psi)$ in
$\vect{A}_{d}$ becomes zero. The configuration $\psi = 0$ corresponds
to the robot base lying on the vertical wall, which never happens if
the ropes are connected directly (i.e., without obstacles in between)
to the robot base, due to the presence of the leg. For over-hanging
walls, no rope-based system can be employed because the contact cannot
be ensured nor maintained.

\section{Static Analysis}
\label{sec:static}
\subsection{Feasible Polytope}
\label{sec:polytopes}
In this section we present a numerical procedure to evaluate the
\textit{static} stability of the robot for a set of locations once
landed on the wall. Since the robot with only one leg would be
inherently unstable on the wall, we consider the ensemble robot {\em
  and} landing mechanism. The analysis aims to answer the following
queries: (\textbf{Q1}) Is it possible to find a static balance between
gravity, rope tensions and forces at the landing wheels for a given
position that allows the robot to remain in contact with the wall?
The presence of this equilibrium is key to the execution of any robot
task.  (\textbf{Q2}) How do the limits on the actuators and the
presence of external forces affect static stability?  More generally,
for each robot location on the wall it is important to evaluate the maximum
operation forces that can be generated.  Both aspects are fundamental
to execute any task, since the contact forces generated for
maintenance operation (called operation forces) must be balanced by
the landing wheels and by the ropes.  For instance, if operation
forces required a force on the feet violating the unilateral
condition, the robot will tip
over. 
%
%
To answer \textbf{Q2} we consider unilaterality, friction and
\textit{actuation} constraints, while to answer \textbf{Q1} only
unilaterality and friction are sufficient.

\begin{figure}[t]
  \centering
  \includegraphics[width=0.9\dimexpr\scalefactor\columnwidth]{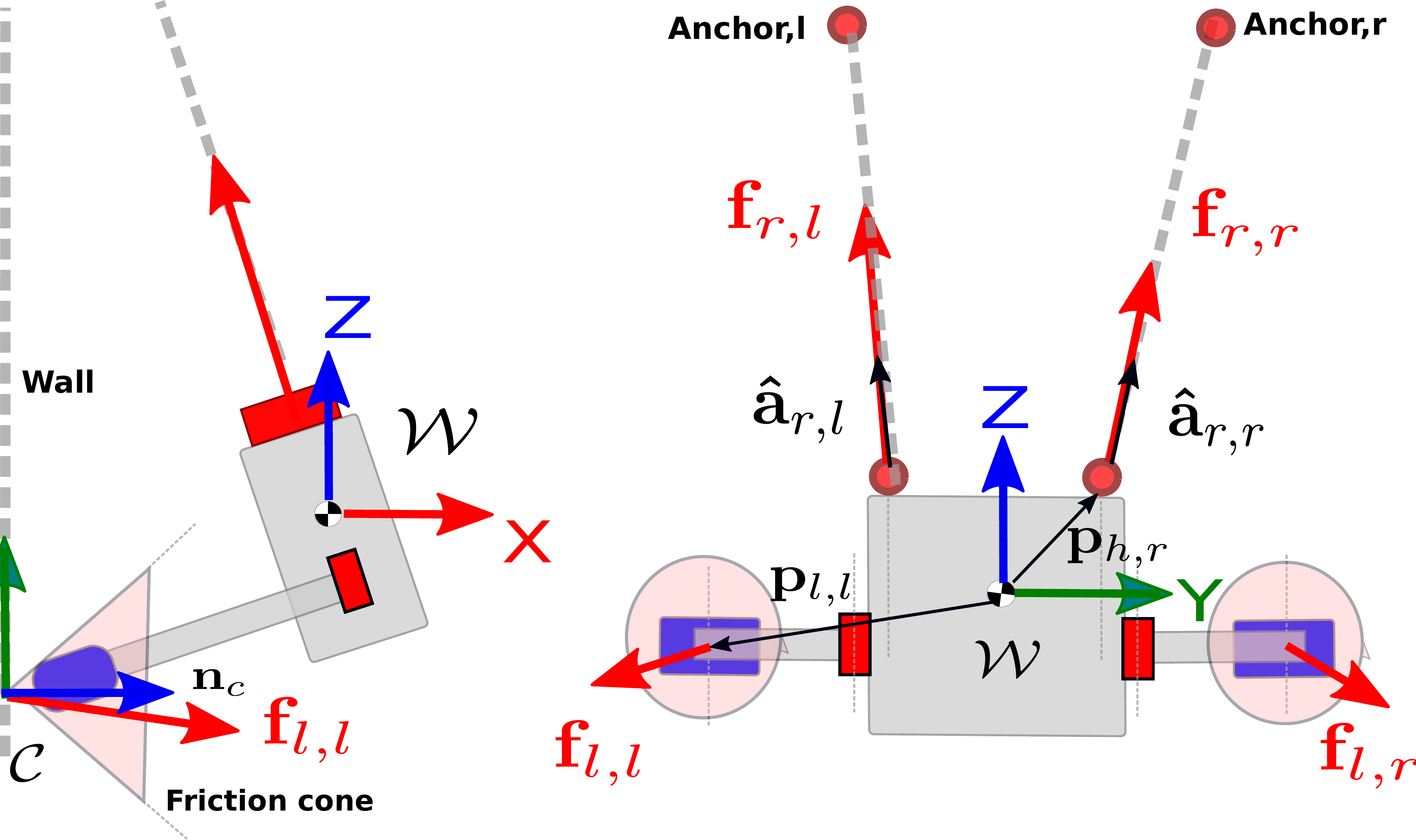}
  \caption{Definition of vectors and frames for the rigid body model
    used in the static analysis: (left) side view (right) front
    view.  Differently from Section \ref{sec:model}, 
    the inertial frame W is attached to the \gls{com}.}
  \label{fig:static_definitions}
\end{figure}

The following assumptions underlie the present analysis: 1) we
approximate the assembly robot+landing mechanism as a \textit{rigid
  body} which is in contact with the wall with the \textit{landing
  wheels}; 2) we neglect the masses of the legs of the landing
mechanism; 3) we assume that the contact forces for the landing wheels
are linear forces (i.e.  point contacts) limited by
the \textit{unilateral} and \textit{friction cone} constraints. For
this analysis we introduce the concept of
\gls{fwp}~\cite{orsolino18ral} and we refer to
Fig.~\ref{fig:static_definitions} for standard definitions.

The \gls{fwp} represents the set of \textit{feasible} wrenches at the
centroid (i.e., forces and moments) than can be tolerated while
satisfying 1) the balance equation of forces under the action of
gravity (static condition), 2) unilateral conditions of the
contacts, 
3) friction constraints
and 4) actuation limits.  Therefore, the feasible wrenches can be seen
as a compact representation of the mentioned constraints 1-4 projected
at the \gls{com} by means of a Minkowski sum operation.

Proceeding with the computation of the \gls{fwp}~\cite{orsolino18ral},
the first step is to compute the Force polytopes $\mathcal{F}_i$
associated with the feet and rope forces.  We define 4
unilateral contacts, two at the \textit{landing wheels} and two at the
rope attachment points with the base (refer to
Fig.~\ref{fig:static_definitions}).
In the case of the landing wheels,
friction constraints are present while  the ropes have 
an additional constraint on the direction: namely, the force has
to be acting along the rope axis.
%
Adopting an approximation that is commonplace in robotics, 
we (conservatively) approximate the cone constraint  with an inner
pyramid and bound them along the normal direction with the 
maximum $f_{\leg, \max}$.
Note that the friction cone implicitly encodes also the unilateral
constraint ($\vect{n}_c^T \vect{f}_{l,i} >0$).
This way, we encode in a
single matrix of constraints:  unilaterality, 
friction and actuation bound for the $i$-th wheel.
The vertices of the obtained force polytope are stored as columns:
{\small
\begin{align}
\vect{f}^{\lim}_{l,i}  = {}_{\mathcal{W}} \vect{R}_c (\vect{n}_c) \mat{ 0  & \mu  &  -\mu   &-\mu  & \mu 	\\
												     0  & \mu  &   \mu   &-\mu	&-\mu  	\\
												     0  &   1  &  	1	 & 1	& 1 	} 	f_{\leg, \max}, i \in \{l, r\},
\label{eq:feet_force_polytope}
\end{align}}
where ${}_{\mathcal{W}}\vect{R}_c$ is the rotation 
matrix related to the contact frame $\mathcal{C}$ and 
$f_{\leg, \max}$ is the maximum force in the normal direction $\vect{n}_c$.
Regarding the rope forces, they  are constrained to lie on a mono-dimensional manifold 
(i.e. along the axis $\vect{\hat{a}}_{r,i}$) and  subject to unilateral ($f_{r,i}<0$) 
and actuation constraints ($-f_{r,i}^{\max} \leq f_{r,i}$). 
Then the Force Polytope associated to rope $i$ boils down to  a segment with only two vertices: 
\begin{equation}
\label{eq:rope_force_polytope}
    \vect{f}^{\lim}_{r,i} =  \mat{ - \vect{\hat{a}}_{r,i} &
      \vect{0}_{3 \times 1} }f_{r, \max} , i \in \{l, r\},
\end{equation} 
Next, for each vertex, we add the moments  that are generated in correspondence to 
the maximum 
forces:
\begin{equation}
\vect{w}_{l, i} =
\mat{\dots
	\vect{f}^{\lim}_{l,i,k}  \dots \\
	 \dots\vect{p}_{l,i} \times \vect{f}^{\lim}_{l,i,k}  \dots} \quad  \text{with }  k = 1, \dots, 5,
\label{eq:feet_wrench}
\end{equation}
where $\vect{p}_{l,i} \in \Rnum^3$ represents the position of the
$i$-th landing wheel and $\vect{w}_{l,i,k}\in \Rnum^6$ represents a
wrench that can be realised at that wheel. Likewise, for ropes:
\begin{equation}
\vect{w}_{r, i} = \mat{  - \vect{\hat{a}}_{r,i}             & \vect{0}_{3 \times 1} \\
						 - \vect{p}_{h,i} \times \vect{\hat{a}}_{r,i}    & \vect{0}_{3 \times 1}}f_{r, \max},
\label{eq:rope_wrench}
\end{equation}
where $\vect{p}_{h,i}   \in \Rnum^3$ is the $i$-th rope attachment point.
Therefore, the set of admissible wrenches that can be applied at the
\gls{com} by the $i-{th}$ wheel is:
\begin{equation}
    \mathcal{W}_{l, i} = ConvexHull(\vect{w}_{l,i, 1}, \dots, \vect{w}_{l,i,5} ), \quad  i \in \{l, r\}.
\end{equation}

The convex hull operation should be performed also for ropes to
compute $\mathcal{W}_{r, i}$ and has the purpose to eliminate internal
vertices.  We now have 4 \textit{wrench polytopes} $\mathcal{W}_i$
that contain all the admissible wrenches that can be applied to the robot's \gls{com}.

Finally, the \gls{fwp} is computed through the Minkowski sum of the
$\mathcal{W}_i$ for  all the contacts:
\begin{equation}\label{eq:mink_sum}
  FWP = \oplus_{i = 1}^{4} \mathcal{W}_i . 
\end{equation}
Since we used a vertex description ($\mathcal{V}$-description), the
Minkowski sum can be efficiently obtained as in
\cite{Delos2015}. \textbf{Remark:} To avoid polytopes becoming
flat\footnote{A flat polytope is a polytope that has some vertex lying
  on its facets.} it is preferable to define all quantities w.r.t. an
inertial frame placed in a location different from the anchors.
More precisely, 
it is convenient to refer all quantities about the \gls{com} (see
Fig. \ref{fig:static_definitions}).  Henceforth, \textbf{only for this
  section}, we assume that the inertial frame is attached to the
\gls{com} and not to the left
anchor. 
In a preliminary analysis, to answer \textbf{Q1}, we assume there are
no actuation limits.  The \gls{fwp}, in this case becomes unbounded,
and becomes a \gls{cwc} \cite{cwc}.  We define a robot position
\textit{feasible} if there exists a set of rope and wheel forces that
1) ensures static equilibrium, 
2) ensures 
unilateral and friction constraints.
\subsection{Gravitational wrench}
\begin{figure}
\centering
  \includegraphics[width=0.8\dimexpr\scalefactor\columnwidth]{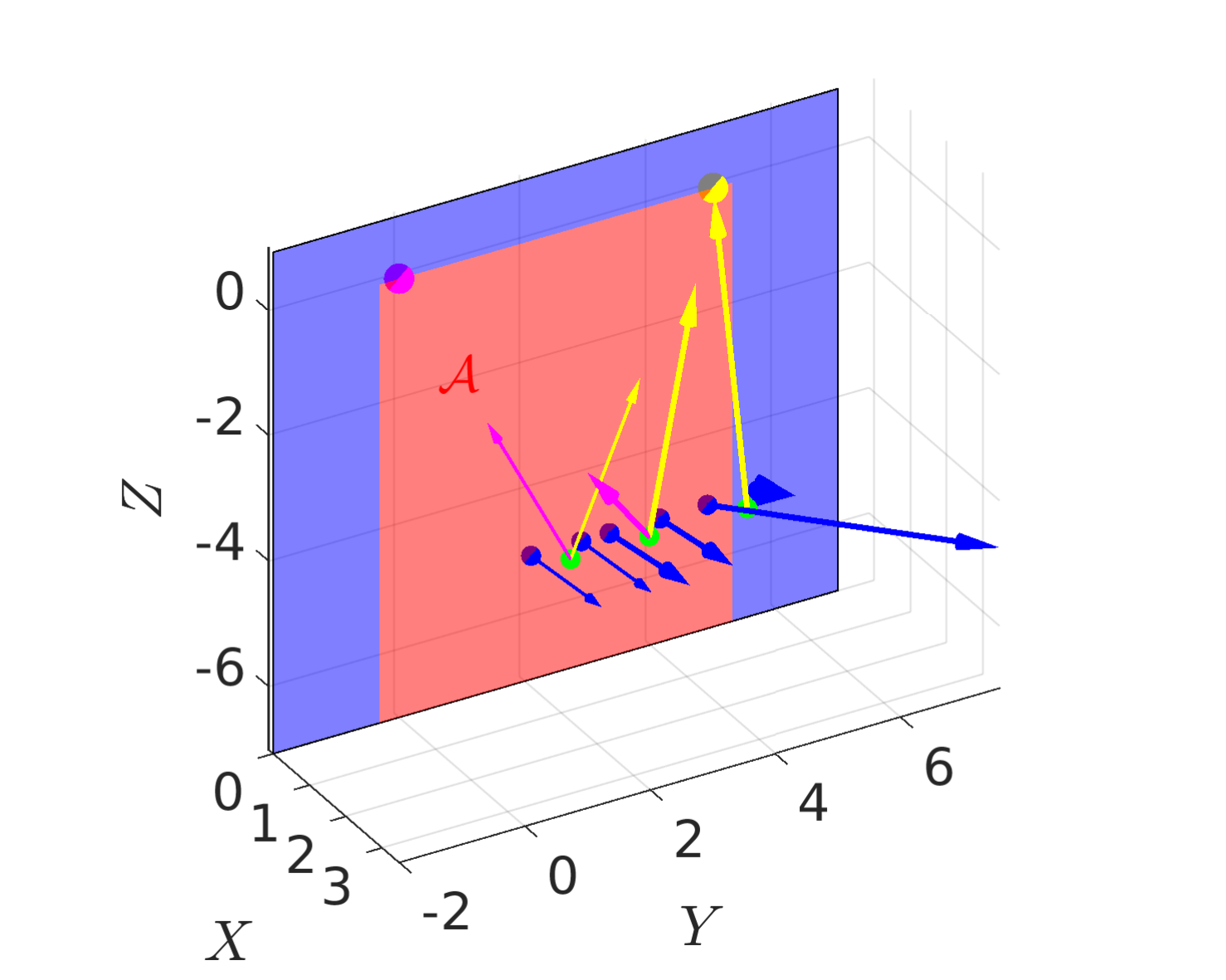}
  \caption{Results of the static analysis in three representative
    points for $\vect{p}_z = 5$~m.  The red shaded area $\mathcal{A}$
    indicates all the robot positions for which static equilibrium, unilateral
    and friction constraints are fulfilled. We report the forces for 3 representative robot positions. 
    The figure shows that, the rope connected to the closer
    anchor point gets gradually loaded when moving towards it, while the
    rope connected to the farther anchor gets gradually
    unloaded.}
  \label{fig:static_analysis_3_points}
\end{figure}

To evaluate the feasibility of a robot location, we first 
compute the gravito-inertial wrench $\vect{w}_G$ in the
specific robot state.
%
Because we are dealing with a static analysis  we neglect the inertial effects:
\begin{equation}
	\vect{w}_G =   \begin{bmatrix}
	m \vect{g} \\
	\vect{p} \times m \vect{g} 
\end{bmatrix},
\end{equation}
where $\vect{p}$ is the robot \gls{com} position (equal to $\vect{0} $
because the inertial frame is attached to the \gls{com}).  A
\textit{feasibility criterion} 
can be written as:
\begin{equation}
    \vect{w}_{G} \in FWP.
\label{eq:stab-crit}
\end{equation}
The criterion \eqref{eq:stab-crit} tells us that it exists a
feasible 
set of forces
$\mat{(\hat{\vect{a}}_{r,l}f_{r,l}), (\hat{\vect{a}}_{r,r}f_{r,r}),
  \vect{f}_{l,l}, \vect{f}_{l,r}}$ that generates the wrench
$\vect{w}_{G}$.
%
We evaluate \eqref{eq:stab-crit} at different  positions 
on the wall,   on a fine grid.  We set a friction coefficient of $\mu=0.8$,
a distance of the anchors  $d_a=5$~m, 
the relative distance of the landing wheels $d_b=0.8$~m  and the 
wall clearance $d_w=0.4$~m (see Fig.~\ref{fig:landing_propellers}).
The outcome of this analysis is that the set of equilibrium locations on
the wall is mostly limited to the rectangle 
delimited by two anchors (red shaded area  $\mathcal{A}$ in
Fig.~\ref{fig:static_analysis_3_points}).  

To answer \textbf{Q2}, we are interested in the \textit{additional}
operation forces that can be tolerated when the robot is in
\textit{equilibrium}, for which we want to evaluate the operating
margin.  This time, we also consider \textit{actuation} limits, and
find that the answer is closely correlated to the concept of
\textit{feasibility margin}, which can be computed from the~\gls{fwp}. We recall that the \gls{fwp} represents the set of admissible centroidal wrenches for a specific position of the robot on the wall, where: 
1) centroidal it means applied at the \gls{com},
2)  admissible it means that this is a  "mapping" of  contact forces satisfy both friction and unilateral constraints at the landing legs and ropes have tensions such that do not get unloaded or exceed their limits.
When the wrench goes out of the \gls{fwp} it means that one of the above constraints 
if violated (e.g. one leg slips or it becomes unloaded, and the robot tips over).

The feasibility margin of applicable operating forces is defined as
the smallest distance between the gravitational wrench $\vect{w}_{G}$
and the boundaries represented by the \gls{fwp} facets.  Authors in
\cite{orsolino18ral} show that an estimate of this margin could be
obtained directly from the vertex ($\mathcal{V}$) description.
However, a margin computed with the vertex description is a normalised
value. Additionally, we might be interested in evaluating the maximum
wrench and then the relative margin in a \textit{specific} direction
$\hat{\vect{v}} \in \Rnum^6$.  In this case the vertex description is
no longer sufficient, and we ought to derive the half-plane
(i.e. $\mathcal{H}$-description) of the \gls{fwp} from the
$\mathcal{V}$-description via the \textit{double description}
algorithm \cite{dd}.
%
%
In the $\mathcal{H}$-description, the \gls{fwp} set can be written in terms of
half-spaces as:
\begin{equation}
	FWP = \{ \vect{w} \in \Rnum^6| \hat{\vect{a}}_j^T \vect{w} \leq \vect{0}, j = 1, \dots n_h \},
\end{equation}
where $n_h$ is the number of half-spaces of the \gls{fwp} and $\hat{\vect{a}}_j \in
\Rnum^6$ is the normal vector to the $j$-th facet. 
The feasibility criterion expressed in \eref{eq:stab-crit} can thus be formulated as:
\begin{equation}
\label{eq:stability2}
	\vect{A}_p \vect{w}_{G}\leq \vect{b}_p,
\end{equation}
where $\vect{A}_p \in \Rnum^{n_h \times 6 }$ is the matrix whose rows are $\hat{\vect{a}}_j$.
%
\subsection{Evaluating the feasibility margin}
The feasibility margin $\gamma \in \Rnum$ in the direction $\hat{\vect{v}}$ is the distance 
(along 
$\hat{\vect{v}}$) between  $\vect{w}_{G}$ and the  
polytope boundary and can be obtained by 
solving the following \gls{lp}:
%
\begin{subequations}
  \label{eq:lp}
  \begin{align}
    \gamma^{\star} =  \argmin_\gamma & -\gamma   \label{eq:cost} \\ 
    \text { s.t. } & \vect{A}_p (\vect{w}_{G}+\gamma \vect{\hat{v}}) \leq \vect{b}_p,  \label{eq:poly_constraint}
  \end{align}
\end{subequations}
Hereby, \eqref{eq:poly_constraint} ensures that the total wrench
$\vect{w}_{G}+\gamma \hat{v}$ is inside the \gls{fwp} set.
\begin{figure}[t]
  \centering
  \includegraphics[width=1.0\dimexpr\scalefactor\columnwidth]{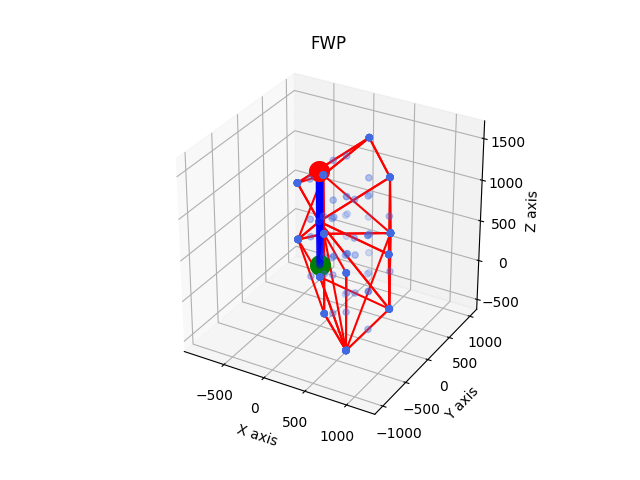}
  \caption{Representation of the linear part of the \gls{fwp} computed
    for a position $\vect{p} = [1.5, 2.5, -6.5]^T$. The green point is
    the gravitational force $\vect{w}_{{G}_{lin}}$.  The red point is
    the maximum force along the direction
    $\hat{\vect{v}}=[0,0,1,0,0,0]$ (blue line). Shaded blue points are
    internal points that are removed by the convex hull operation.}
  \label{fig:fwp}
\end{figure}
The solution $\gamma^{\star}$ of \eqref{eq:lp} gives the surplus of
additional external wrench $\gamma^{\star} \vect{\hat{v}}$ that can be
tolerated by the system
\cite{orsolino18ral}. 
Since it is not possible to visually represent a set of 6D elements using a 3D figure, in Fig.~\ref{fig:fwp},
we depict only the linear component of the \gls{fwp} (i.e., the set of admissible \emph{centroidal} linear forces), computed for the position 
$\vect{p} = \mat{1.5& 2.5& -6.5}^T$.
In this representation, the green dot indicates the gravitational force, 
while the red dot shows the maximum additional force that can be applied in the direction 
of the blue line (oriented vertically) before  becoming unfeasible.
\\
%
\subsection{Numeric evaluation}
\begin{figure*}
    \centering
     \includegraphics[width=0.24\dimexpr\scalefactor\textwidth]{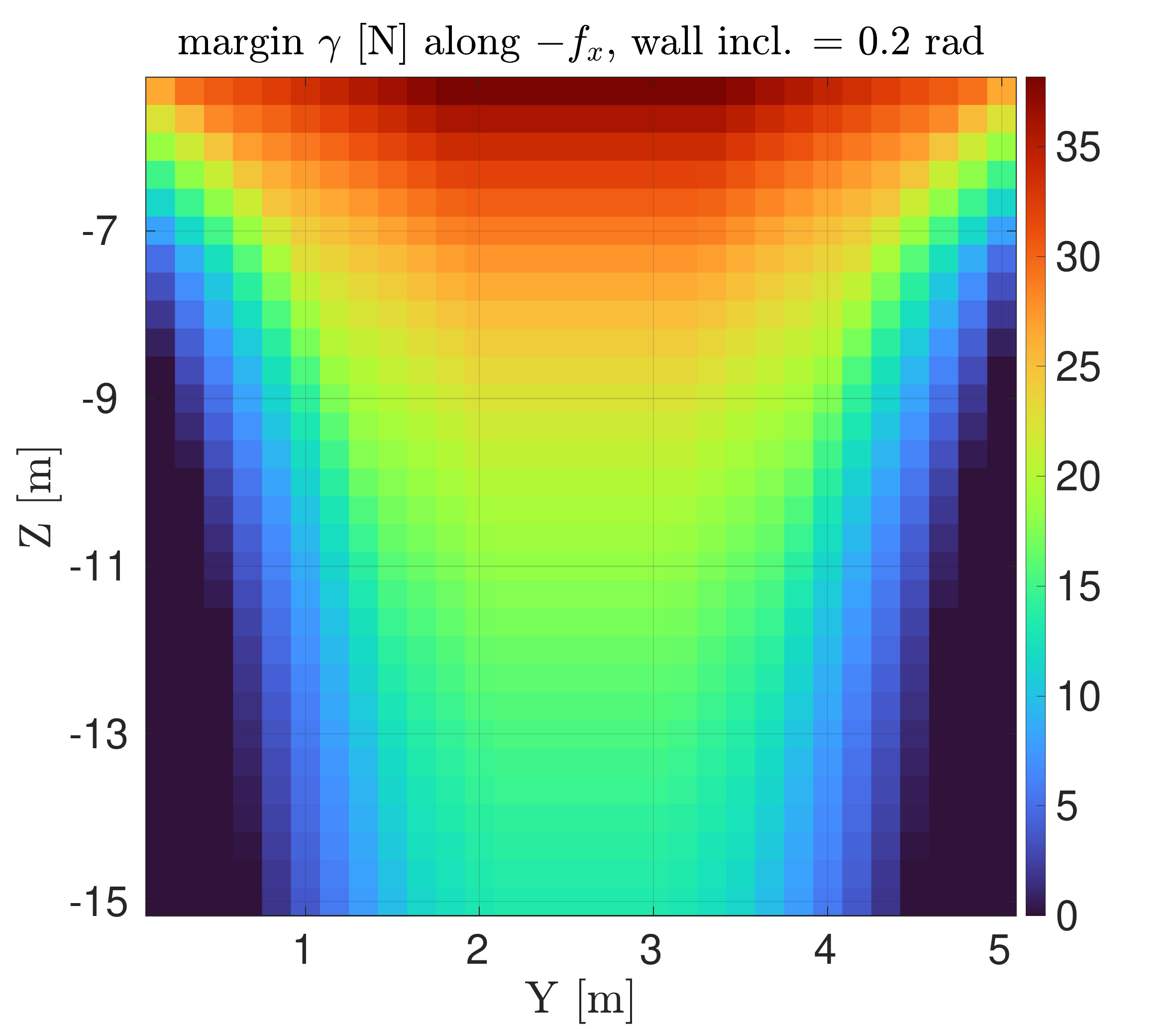}
      \includegraphics[width=0.24\dimexpr\scalefactor\textwidth]{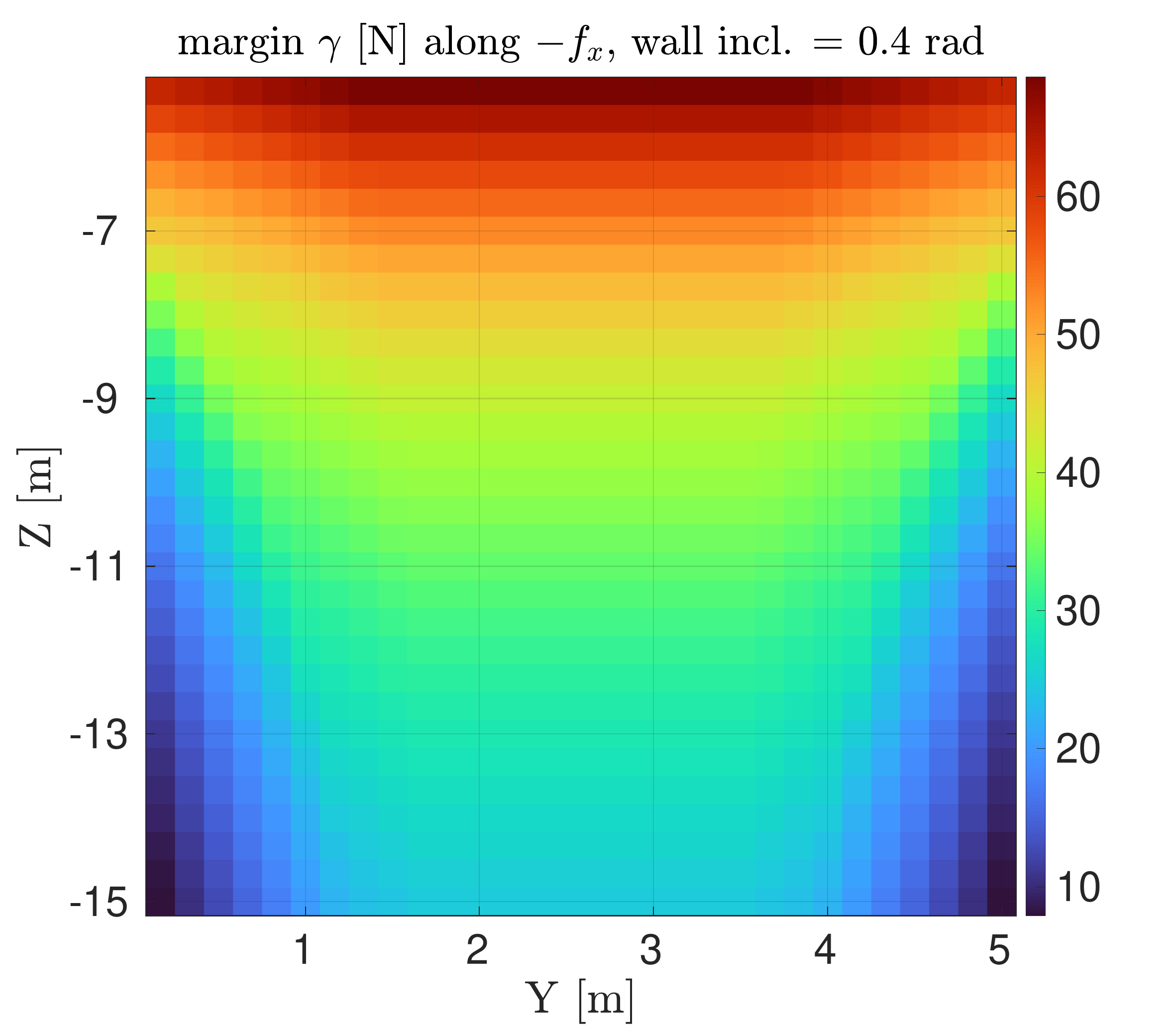} 
    \includegraphics[width=0.24\dimexpr\scalefactor\textwidth]{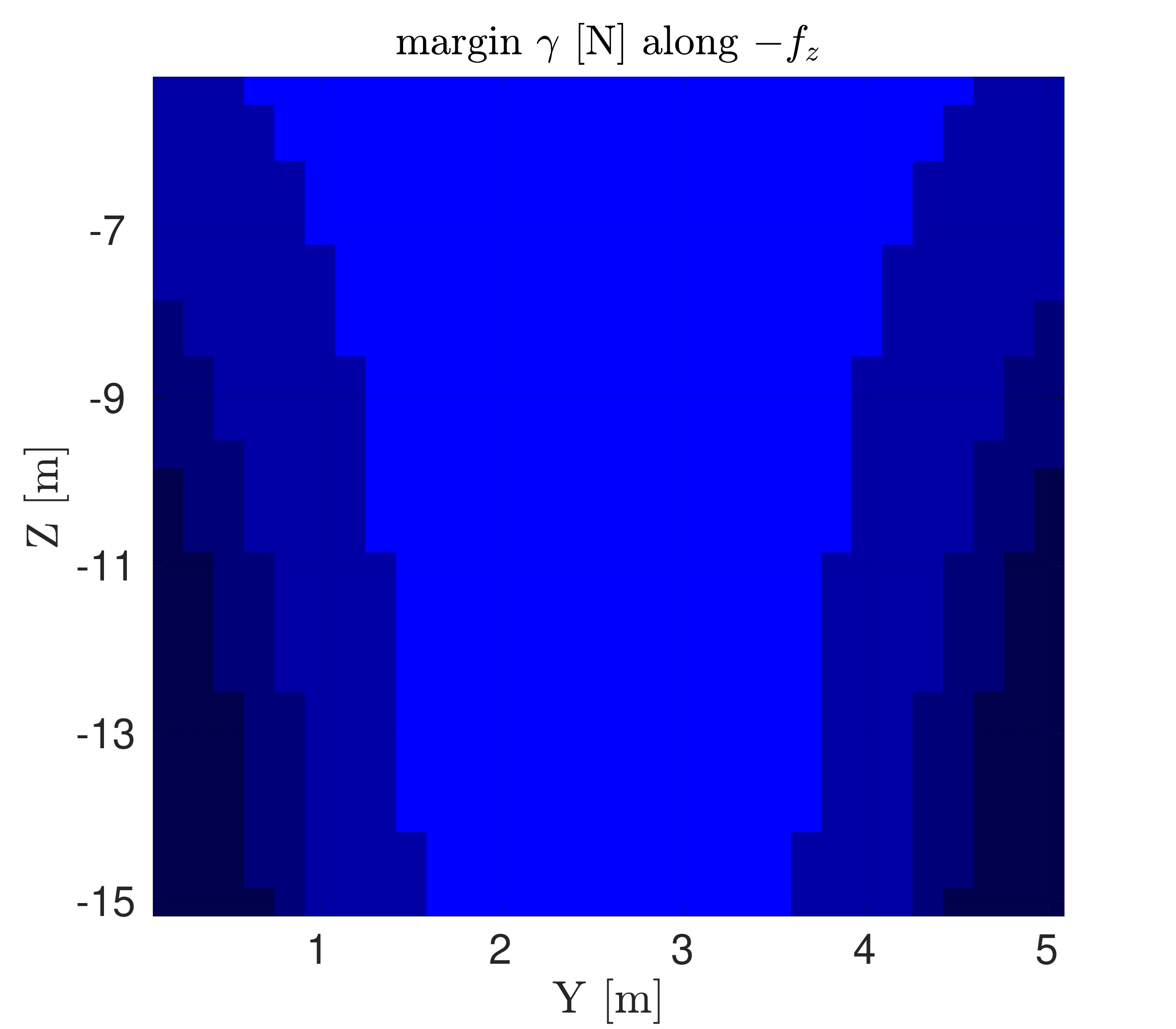}
     \includegraphics[width=0.24\dimexpr\scalefactor\textwidth]{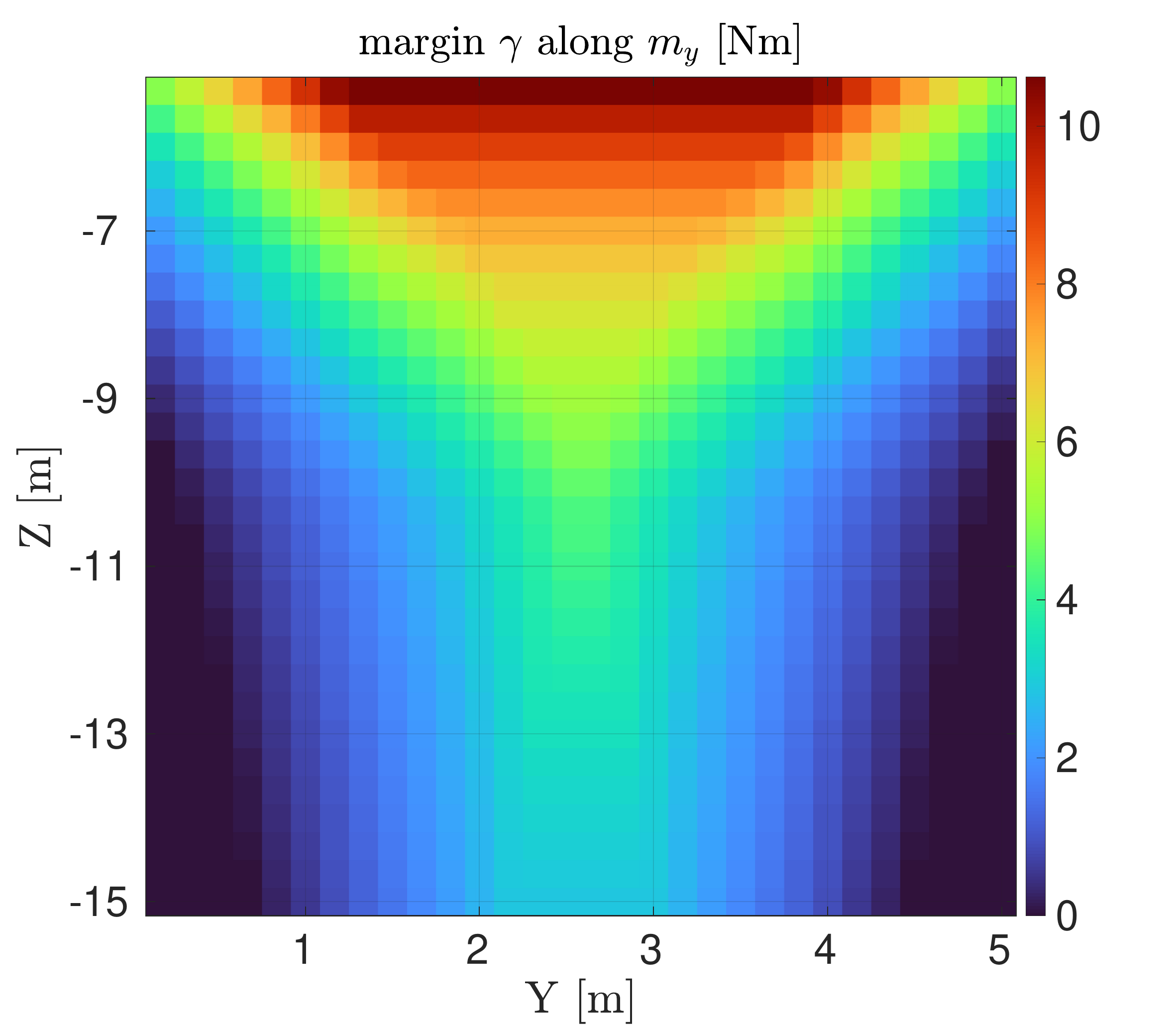}
	\caption{Heat-maps of the feasibility margin computed for
          different positions along the wall, (left plots) along 
          $-\vect{f}_x = [-1, 0, 0, 0, 0, 0]^T$ for a hammering
          application, for a wall inclination of 0.2 rad (first), and
          0.4 rad (second) and (right plots) along 
          $-\vect{f}_z =[0, 0,-1,0,0,0]^T$ (third) and
          $\vect{m}_y = [0, 0,0,0,-1,0]^T$ direction (fourth) relative
          to a debris removal operation which involves the application
          of a not purely centroidal force (also a moment).  In the
          third plot we reported the results for 3 values of the
          friction coefficient $\mu = \{0.4, 0.6, 0.8\}$.  The feasibility
          margin is a synthetic metric to estimate the amount of
          external forces and moments available for operations. We
          report the absolute value of the margin $\gamma$.}
	\label{fig:heatmap}
\end{figure*}

In a first representative operation, that ensembles a hammering
operation, we compute the feasibility margin to generate a pure force
(i.e. passing through the \gls{com}) along the $X$ direction
$\hat{\vect{v}} = \mat{-1 &0&0&0&0&0}^T$ for robot positions with $Y$
ranging between the two anchors (i.e. from $0$ to $5$~m) and the $Z$
component ranging from $-2$ to $-10$~m.  We repeat the analysis for
two different wall inclinations (0.2 and
0.4~rad) that will result in different $\vect{n}_c$.
To have the robot in contact, the $X$ component
is set to $1.7$ and $3$~m, respectively.  Figure~\ref{fig:heatmap}
(left plots) reports the 2D (heat-map) representation of the value of
the feasibility margin for the two different wall inclinations.  The
friction coefficient is set to $\mu = 0.8$.
As expected, the margin decreases with the distance to the anchors
because the angle $\psi$ decreases with rope elongation (for the same
wall clearance), and pushing on the wall becomes more difficult
because the legs get more and more unloaded. For the same reason, a
more slanted wall involves overall higher gravity components and,
hence, overall more feet loading, thus higher values for the margin
can be observed. 
The diagonal shape is related to the friction constraint which is violated in the dark blue area.
In these cases the whole static feasibility is invalidated (i.e. $\vect{w}_{G}$ goes
out of the polytope) and we set the margin to be null.
%

In a second analysis (see Fig.~\ref{fig:heatmap}, right plots) we
consider a \textit{composite}
loading 
as observed in a \textit{debris removal} application, pictorially
represented in Fig.~\ref{fig:applications}, middle. This time we
consider a purely vertical wall (i.e. $\vect{n}_c = \mat{1&0&0}^T$)
with a constant $\vect{p}_x = 1.5$~m. Interestingly, for the margin in
the vertical direction, only two values are possible: zero in the
region for which friction constraints are violated (we did a
parametric analysis for $\mu = \{0.4, 0.6, 0.8\}$ to highlight its
influence), and the value of the gravity ($147$~N) in the remaining
region. This means that the vertical push is obtained through the
action of gravity and cannot go beyond that. We also noticed that the
position of the feet is fundamental to achieve a stable configuration
(i.e. they cannot be shifted too much w.r.t the \gls{com} in the
vertical direction, cf. Fig \ref{fig:static_definitions}).  
The results for the moment $\vect{m_y}$ (fourth plot) are more inline with
the $\vect{f}_x$ case, where higher margins are related to positions
between the anchors.
This is a \textit{quantitative} evaluation that is important for real
operations:
%
being able to evaluate the operating margin is a feature useful to
optimise costs-related maintenance operations depending on the type of
the rock.  
Being able to estimate the maximum force
applicable gives benefits in terms of cost effectiveness, because it
allows to minimise the time budget for the operation.
\section{Motion Planning}
\label{sec:motion_planning}
Setting up a navigation approach for the ALPINE robot is a non-trivial
task.  First of all, even the simplified dynamics~\eqref{eq:simplified_2ropes_minimal} is highly nonlinear.  Second,
the configurations between which the robot moves are usually distant,
and therefore, linearising the dynamics around an
operating point would lead to inaccurate results.  Third, one
of the inputs, the thrusting force $\mathbf{f}_{\text{leg}}$, has an
impulsive nature, operating at discrete time instants, and its
tangential component is constrained by friction. Therefore, it is not
possible to use it in any feedback control scheme.

As customary in robotics, we split the navigation problem into two sub-problems:
1) motion planning, i.e., deciding a trajectory before the motion
starts, 2) motion control, i.e., a feedback controller that operates the rope/propeller
forces to compensate for small deviations to track the trajectory and land in proximity of the expected position.
%
This section focuses on step 1 and considers the robot 
\textit{without} the landing mechanism. The motion plan for the robot is
decided by solving a \textit{nonlinear} optimal control problem
(OCP), which in general is modelled as follows: 
\begin{subequations}	
	\begin{align}
		 \min_{\vect{x}(t), \vect{u}(t)} & \,\, m_f(\vect{x}(t_f)) + \int_{0}^{t_f}
	   l\left(\vect{x}(t),\vect{u}(t)\right) dt , \\  \nonumber
		\text{s.t.} &\\
		&\,\,\dot{\vect{x}}(t) = \vect{f}(\vect{x}(t),\vect{u}(t)),\\
		&\,\,\vect{u}(t) \in \mathcal{U},\\
		&\,\,\vect{h}\left(\vect{x}(t),\vect{u}(t)\right) \leq 0,\\
        &\,\,\vect{B}(\vect{x}(t_0), \vect{x}(t_f))=0,
	\end{align}
\label{eq:OCP_global}
\end{subequations}
where $l(\vect{x},\vect{u})$ and $m_f(\vect{x})$ are the running and
the terminal costs, $\vect{x}$ is state, $\vect{U}$ is the control input constrained in the convex
set $\mathcal{U}$, $\vect{f}$ is the dynamics,
$\vect{h}$ are the \textit{path} constraints, and $\vect{B}$ are the boundary conditions. 
To reduce the computational effort, we consider as $\vect{f}$ the reduced-order model~\eqref{eq:simplified_2ropes_minimal}, which implicitly
includes the closed-loop kinematic constraints.  The control inputs should include both the rope forces $\vect{u} = \mat{f_{r,l}, f_{r,r}} \in \Rnum^2$ and the leg impulse. However, the latter is applied only during the thrusting phase, which typically lasts a short amount of time $t_{\text{th}}\in \Rnum$. For this reason, we assume it remains constant during this phase, and treat it as an additional \textit{state} with null dynamics.  As decision variables, we
consider: 1) the vector of control inputs
$\vect{U} = \mat{ \vect{F}_{r,l}, \vect{F}_{r,r}}$ where
$\vect{F}_{r,l}, \vect{F}_{r,r}$ are the trajectories of the rope forces
along the horizon, 2) the horizon length $t_f\in \Rnum$, 
and 3) the initial leg impulse $\vect{f}_{\text{leg}} \in \Rnum^3$.

\noindent
{\bf System Dynamics.}
Because we define \textit{a priori} the duration of the thrusting
phase, the dynamics of our system is no longer hybrid but \textit{time-varying}. When the leg is in contact, the dynamics accounts for the force $\mathbf{f}_{\leg}(t)$ generated by the thrusting leg.
Rather than modelling this force as an impulse, we model it as:
\begin{align}
\vect{f}_{\leg}(t) =
	\begin{cases}
		\vect{f}_{\leg}  \quad 0 \leq t \leq t_{\text{th}} ,\\
		\vect{0}_{3\times1}  \quad t > t_{\text{th}} .
\end{cases}
\end{align}
Hence, during the contact $t \in \mat{0 & t_{\text{th}}}$, the leg applies a
constant force $\vect{f}_{leg}$ while, for $t \in \mat{t_{\text{th}} & t_{f}}$, this force is zero. 
Given the nature of our actuation  mechanism, a realistic choice
is to have a duration in the order of  tens of milliseconds, which consequently determines a remarkable intensity of the impulse.
We incorporate the leg force $\vect{f_{\text{leg}}}(t)$ in the state vector $\vect{x}$ with a null dynamics. 
The state of our problem, hence, is defined as:
\begin{equation}
	\vect{x} = \mat{\psi & l_1 & l_2 & \dot{\psi} & \dot{l_1} &
          \dot{l_2} & \vect{f}_{\text{leg}}^T }^T \in \Rnum^9 .
\end{equation}
However, when the contact is broken at $t = t_{th}$, the variable $\vect{f}_{\text{leg}}$ must disappear from the state.
We model this change in state dimension by defining a matrix $\vect{S}$ that selects the first 6 elements the 9D state vector:
\begin{align}
    \begin{array}{ll}
      \vect{x}^{+}= \vect{S} \, \vect{x} , & t = t_{th} , \label{eq:ss_dynamics_resetmap}
    \end{array}
\end{align}
Consequently, the \textit{state-space} representation of the
nonlinear dynamics, casted in input-affine form, can be derived
from~\eqref{eq:simplified_2ropes_minimal}:
\begin{align}
    \begin{array}{ll}
      \dot{\vect{x}} = \vect{f}_{leg}(\vect{x}) + \vect{g}_{leg}(\vect{x}) \vect{u},  & t < t_{\text{th}}, \\
      \dot{\vect{x}} = \vect{f}_{noleg}(\vect{x}) + \vect{g}_{noleg}(\vect{x}) \vect{u},  & t > t_{\text{th}},
    \end{array}
\end{align}
where:
\begin{align}
	\small
	&\vect{f}_{leg}(\vect{x}) =
	\mat{
		\vect{x}_{4\dots 6}\\
		\vect{A}_{d}^{-1}  \left[ \frac{\vect{x}_{7\dots9}}{m}+ \vect{g} - \vect{b}_{d}  \right]  \\
		\vect{0}_{3 \times 1}
    } 
  \\
  & \vect{g}_{leg}(\vect{x}) = \mat{ \vect{g}_{noleg}(\vect{x}) \\ \vect{0}_{3 \times 2} }\\
  &\vect{f}_{noleg}(\vect{x})=
	\mat{
		\vect{x}_{4\dots 6}\\
		\vect{A}_{d}^{-1}  \left (\vect{g} - \vect{b}_{d}  \right)  \\
    } 
  \\
	&\vect{g}_{noleg}(\vect{x}) = 
    \mat{  		\vect{0}_{3 \times 1}	        	& 				    \vect{0}_{3 \times 1} \\
				 \vect{A}_{d}^{-1}\hat{\vect{a}}_{r,l}  & \vect{A}_{d}^{-1}\hat{\vect{a}}_{r,r}
      }
\label{eq:DynamicsStateSpaceMinimal}
\end{align}

To solve OCP~\eqref{eq:OCP_global} we apply a \textit{direct method} to
transform  the infinite dimensional problem into an \gls{nlp}.  In
particular, we chose a \textit{single shooting} approach where we
discretised the rope forces along the horizon in $N$ knots equally
spaced by $dt = t_f/N$ time intervals and regarded the states
$\vect{x}(k) \in [0, N]$ as \textit{dependent} variables, obtaining the
following \gls{nlp}:
\begin{subequations}
\label{eq:NLP}
\begin{align}	
	\min_{\vect{U},\vect{f}_{leg},  t_f} & \sum_{k=0}^{N-1} \ell \left(\mathbf{x}_k, \mathbf{u}_k\right)+\ell_{\mathrm{f}}\left(\mathbf{x}_N\right)   \label{eq:cost_discrete}\\
	\text { s.t. } & \mathbf{x}_0=\hat{\mathbf{x}}_0,																					                   \label{eq:bound_discrete}   \\
	              & \mathbf{x}_{k+1}=  \vect{f}_k(\mathbf{x}_k) + \vect{g}_k(\vect{x}_k) \mathbf{u}_k, k \in \mathbb{I}_0^{N-1},      				   \label{eq:dyn_discrete}   \\
	              & \vect{h}_k \left(\mathbf{x}_k, \mathbf{u}_k\right) \leq 0, \quad k \in \mathbb{I}_0^{N-1},											   \label{eq:path_discrete}
\end{align}
\end{subequations}
%
We compute the state vector $\vect{x}_{k+1}$ from the input
$\vect{u}_k$, by integrating the dynamics
$\eqref{eq:dyn_discrete}$ via a fourth-order method (i.e.  Runge-Kutta
4) starting from the initial state $\vect{x}_0=\hat{\vect{x}}_0$.  
The number of knots $N$ should be
roughly adjusted according to an estimate of the jump duration (e.g. by
heuristics) in order to reduce the impact of integration errors.
Furthermore, to mitigate this issue, we
found beneficial to perform integration on a finer grid
(cf. paragraph ``Integration errors'' in Sec.~\ref{sec:gazebo_simulation} for a performance evaluation of
different integration schemes).  This has the advantage to improve
integration accuracy without increasing the problem size.

\noindent {\bf Boundary Conditions.}
To start the integration of the dynamics we need to set the initial value for the
state:
\begin{equation}	
	\hat{\vect{x}} = \mat{\psi_0 & l_{1,0} & l_{2,0} & \dot{\psi}_0 & \dot{l}_{1,0} & \dot{l}_{2,0} & \vect{f}_{\text{leg}} }^T.
\end{equation}
Apart from $\vect{f}_{leg}$ that is an optimisation variable that we
can arbitrarily set (see the \emph{Initial Guess} paragraph at the end
of this section), the other entries can be obtained from the initial
robot Cartesian position/velocity. 

The inverse kinematics mapping, to convert the robot position $\vect{p}$ into $\vect{q}_r$
can be obtained with simple geometric analysis:
\begin{align}
\begin{cases}
  &  \psi = \arctan2(\vect{p}_x, - \vect{p}_z) , \\
  & l_1 = \Vert \vect{p} - \vect{p}_{a,l} \Vert , \\
  & l_2 = \Vert \vect{p} - \vect{p}_{a,r} \Vert , \\
  & \dot{\phi} = \frac{(\vect{n}_{\parallel}\times \vect{n}_{\perp})^T
    \dot{\vect{p}}}
  {(\vect{n}_{\parallel}\times \vect{n}_{\perp})^T
    (\vect{p}_{a,l}-\vect{p}_{a,r})} , \\
  &\dot{l}_1 = (\vect{p} - p_{a,l})^T \dot{\vect{p}} , \\
  &\dot{l}_2 = (\vect{p} - p_{a,r})^T \dot{\vect{p}} , \\
\end{cases}
  \label{eq:ik_minimal}
\end{align}
where
$\vect{n}_{\parallel} = (\vect{p}_{a,l} - \vect{p}_{a,r})/\Vert
\vect{p}_{a,l} - \vect{p}_{a,r} \Vert$ is the unit vector passing through the
anchor points and $\vect{n}_{\perp} = I_{3 \times 3} - \vect{n}_{\parallel}
{\vect{n}_{\parallel}}^T$ perpendicular to the rope plane.

Likewise, for cost and constraint computation, which involve the
position/velocity of the robot, these are related to the state
variables by means of~\eqref{eq:fwd_kin_minimal} and its derivative.
To avoid overloading the notation, henceforth, we implicitly assume
that $\vect{p}$, $\dot{\vect{p}}$ are expressed as a function of the
state $\vect{x}$.  Finally, we wish to enforce that the robot is at
the target position $\vect{p}_{tg}$, at the end of the horizon
(i.e. end of the jump).  The use of hard constraints can prevent
convergence when the problem is close to unfeasibility or
ill-conditioned.
We found it useful to relax the constraint of the terminal state by adding a fixed slack ($s$):
\begin{align}
\Vert \vect{p}(t_f) - 	\vect{p}_{tg} \Vert  - s \leq 0 .
\end{align}
Finally the time $t_f$ is bound to be positive. 

\noindent
{\bf State Constraints.}
State constraints are related to regions of the operation space that
are not accessible.  A constraint is added to prevent collision with
the wall for the whole jump duration
\begin{align}
  \vect{n}^T\vect{p}    \geq \epsilon ,
 \label{eq:wall_constraint}
\end{align}
where $\vect{n}$ is the normal pointing outwards the wall.
We set a threshold $\epsilon$ to avoid the  singular configuration  $\vect{p}_x = 0$ ($\psi = 0$)  for the model \eqref{eq:simplified_2ropes_minimal}.
%
To encourage the optimisation to make the robot detach from the wall
with a desired clearance $c \in\Rnum$ we force a via point inequality (e.g. at
the half of the duration ($t_f/2$) of the trajectory).
\begin{align}
	\vect{n}^T\vect{p}(t_f/2)    \geq c .
    \label{eq:viapoint_constraint}
\end{align}

If the rocky wall has an irregular shape, or there are obstacles that
the robot should overcome in order to reach its final destination,
the via point constraint should be replaced by one that forces
the robot to avoid the obstacle. We can model the inaccessible area as
a region whose boundary is a surface. We assume that this surface can
be expressed by a smooth 2D manifold expressed by
$Q(\vect{x}) = 0$.  Therefore the admissible region is generally given
by $Q(\vect{x}) \geq 0$ and could be non-convex.

\noindent
{\bf Actuation Constraints.}
The rope forces 
are bounded by the limits of the
actuators (e.g. a hoist) and by unilateral constraints:
\begin{align}
  -f_{r,i}^{\max} \leq f_{r,i} \leq 0 , \quad   i = \{l,r \} .
  \label{eq:max_rope_force}
\end{align}
Likewise, the norm of the impulse force $\mathbf{f}_{\leg}$ is upper-bounded by the
actuation limit:
\begin{align}
  \Vert \vect{f}_{\leg} \Vert \leq f_{\leg, \max} ,
\end{align}
while the unilateral constraint is encoded from its normal component:
\begin{align}
 \vect{n}_c^T \vect{f}_{\leg} \geq 0 ,
\end{align}
where $\vect{n}_c$ is the surface normal at the contact location. This can be locally 
different from the wall inclination $\vect{n}$ if there are rock asperities.
Additionally, the tangential components of $\vect{f}_{\leg}$ are
constrained by the following second order friction cone:
\begin{align}
  \Vert \vect{t}_{x}^T \vect{f}_{\leg}  + \vect{t}_{y}^T
  \vect{f}_{\leg}  \Vert \leq \mu \vect{n}_c^T \vect{f}_{\leg}.
\end{align}

%
\noindent{\bf Objective Function.}
As terminal cost we minimise the
final kinetic energy $K(t_f)$ at touch-down. We project the
touchdown velocity in the direction normal to the contact location, because 
it is the one that is getting nullified by the impedance 
strategy illustrated in Section \ref{sec:landing_control}.  For the
running cost we consider 1) a smoothing term for the rope forces, and 2) a second term that penalises the hoist work (i.e. work made to wind/unwind the ropes).  Thus, the cost function is:
\begin{align}
  J &=w_s \sum_{k=0}^{N-1} (f_{{r,i}_k} - f_{{r,i}_{k-1}})^2 +  w_{hw} \sum_{k=0}^{N-1} \vert f_{r,i} \dot{l}_i  \vert T_s \\
    & \quad   + w_i      \underbrace{\frac{1}{2}m \vect{\dot{p}}(t_f)^T
  \vect{P}\vect{\dot{p}}(t_f)}_{K(t_f)}, \quad i \in \{l,r\} ,
      \label{eq:NLP_cost}
\end{align}
with $w_s$, $w_{hw}$ and $w_i$ being the weights associated to the
three cost components and $\vect{P} = \vect{n}_c \vect{n}_c^T$ a projection operator to extract the normal component from the velocity $\vect{\dot{p}}$.


\noindent\textbf{Initial guess.}
To speed-up convergence, it is worth to initialise the optimisation
with a feasible initial guess.  The rope forces are initialised with
zeros and the leg impulsive component with $\mat{f_{\leg,\max} & f_{\leg,\max} &f_{\leg,\max}}^T$  to easily
explore areas away from the singularity. The time $t_f$ is initialised
with the time constant for the system linearised around the initial
position $\vect{p}_0$.
Note that the linearised system becomes also a reasonable approximation when the jump length is small with respect to the rope elongation. 
%
%
\section{Motion Control}
\label{sec:fly_motion_control}
\subsection{Flying motion control: Linear position}

\label{sec:mpc}
During the \textit{flight} phase, which starts after the thrusting
phase, the leg has no longer influence on the linear motion of the
base 
and the ropes are the only actuators to control the robot's motion.   Several factors can make the actual robot position diverge from the
planned trajectory (e.g., tracking inaccuracies of the leg 
impulse,  approximations due to  the use of the reduced order 
and environmental disturbances like, for instance, wind).  The
deviation from the planned trajectory is  exacerbated by the
presence of strong nonlinearities in the system and can result in  unsatisfactory outcomes.  For this reason, the presence of a feedback-based motion controller is imperative.  The strong changes in the
configuration of the system during the flight impede the application
of linear strategies around a linearised equilibrium.  On the other
hand, nonlinear (i.e. Lyapunov-based) approaches are not easy to
design in the presence of unilateral constraints (that act as saturation). The presence of saturation constraints also prevents the usage of feedback linearisation approaches.
With these considerations in mind,
a constrained \gls{mpc} appeared as
the most suitable solution, since it allows to account for several types of constraints. A potential problem with \gls{mpc} is that, given a nonlinear model, the resulting optimisation problem is non-convex and potentially difficult to solve \textit{online}. Once again, the reduced-order model helps to diminish these problems.

The \gls{mpc} optimisation problem aims to  minimise the tracking error 
with respect to the optimised reference position $\vect{p}_{\text{ref}}$.  
The decision variables are: 1) the deviations
$\Delta\vect{F}_{r,l},\Delta\vect{F}_{r,r} \in \Rnum^{2N_\text{mpc}}$ 
of the rope forces with respect to the nominal
feed-forward values $\vect{F}_{r,l}^{*}, \vect{F}_{r,r}^{*}$ (the solution of \eqref{eq:NLP}) and 2) the 
trajectory of the force  $\vect{F}_p \in \Rnum^{N_\text{mpc}}$ generated by the propeller, where
$N_\text{mpc}$ is the length of the \gls{mpc} horizon.
As for motion planning, the reduced-order 
dynamics~\eqref{eq:dyn_discrete}  is integrated to obtain the states (in a single shooting fashion) but this time starting from the \textit{current} state $\hat{\vect{x}}$ at sample $k$.  The main differences w.r.t. the offline NLP \eqref{eq:NLP} are that: 1) the \gls{mpc} is active only during the \textit{flight} phase, hence,  the leg impulse is not considered, and 2) we integrate as inputs $\vect{F}_{r,i}^{*} + \Delta\vect{F}_{r,i}$ instead of $\vect{F}_{r,i}^{*}$.
%
%
%
%
The resulting optimisation problem is as follows:

{
\begin{subequations}
  \small
  \begin{align}
    \label{eq:MPC}
    & \min_{\Delta\vect{F}_{r,\{l,r\}}, \vect{F}_{p}} w_p \sum_{i=0}^{N_\text{mpc}-1} \Vert\vect{p}_{{\text{ref}}_{k + i}}  - \vect{p}_i \Vert^2 + \\ \nonumber
         & +  w_u\left(\sum_{i=0}^{N_\text{mpc}-1} (f_{{r,l}_i} - f_{{r,l}_{i-1}})^2  + (f_{{r,r}_i} - f_{{r,r}_{i-1}})^2\right) \label{eq:mpc_cost} \\ \nonumber
         & + w_{pf} \Vert\vect{p}_{{\text{ref}}_{k + N_\text{mpc}}} -
           \vect{p}_{N_\text{mpc}} \Vert^2 ,   \\ \nonumber
         &\text{s.t.}  \nonumber\\
         &\vect{x}_0 = \hat{\vect{x}}_k, \\
         &\vect{x}_{i+1} = \vect{f}\left(\vect{x}_i, \mat{\vect{F}_{{r,l}_{k+i}}^{*} + \Delta\vect{F}_{{r,l}_i} \\
                                                      \vect{F}_{{r,r}_{k+i}}^{*} + \Delta\vect{F}_{{r,r}_i} \\
                                                       \vect{F}_{p_i} } \right),  \quad i \in \mathbb{I}_0^{N_\text{mpc}-1}, \\ 
         &-f_{r,\max} \leq \vect{F}_{{r,l}_{k+i}} + \Delta
           \vect{F}_{{r,l}_i} \leq 0 , \quad  i \in    \mathbb{I}_0^{N_\text{mpc}-1} ,
           \label{eq:max_frl}  \raisetag{1\normalbaselineskip}\\ 
         &-f_{r,\max} \leq \vect{F}_{{r,r}_{k+i}} + \Delta   \vect{F}_{{r,r}_i} \leq 0 ,  \quad i \in
           \mathbb{I}_0^{N_\text{mpc}-1} ,
           \label{eq:max_frr} \raisetag{1\normalbaselineskip} \\ 
          & -f_{p,\max} \leq \vect{F}_{p_i} \leq f_{p,\max}  \label{eq:propeller_bounds}          
\end{align}
\end{subequations}
}
where $k$ is the index relative to the solution of \eqref{eq:NLP}
while $i$ is the index for the \gls{mpc} trajectory (receding horizon).  In the cost
function~\eqref{eq:mpc_cost}, we included also a regularisation
(smoothing term) for $\Delta\vect{F}_{r,l},\Delta\vect{F}_{r,r}$.
The constraints \eqref{eq:max_frl} and
\eqref{eq:max_frr} bound the rope
forces, while \eqref{eq:propeller_bounds}  bounds the propeller force.
Problem \eqref{eq:MPC} is solved every $dt_\text{mpc}$ seconds and
only the first value is retained from the solution
$\Delta\vect{F}_{r,l},\Delta\vect{F}_{r,r}$, while the rest is
discarded.  As common practice, to speed up convergence, we bootstrap
each optimisation with the solution of the previous loop.
Note that the simulation is running 
at a much higher rate  than the \gls{mpc} ($dt_\text{sim}\ll dt_\text{mpc}$) and
we apply zero-order hold.
When the last sample \gls{mpc} trajectory matches with the end of
the optimal reference trajectory, we start reducing the length of the
\gls{mpc} horizon until the end of the reference
trajectory.  
%
\subsection{Flying motion control: orientation}
\label{sec:orientation_control}
In the previous section, we showed how to control the robot Cartesian
position during the flight phase, however
controlling the orientation is also relevant.
To avoid creating undesired moments, the direction of $\vect{f}_c$
should point towards the \gls{com}.  Tracking errors on $\vect{f}_c$
could result in moments that would initiate unwanted pivoting
motions. Indeed, this eventually leads to undesirable landing postures
(i.e. with landing wheels misaligned with the rock wall), with the
risk of tipping over.
%
A solution could be to optimise also for the orientation in \eqref{eq:NLP}, but this would require the extension of our simplified model to describe also the angular dynamics.

\noindent\textbf{Leg reorientation.}
In the case moderate orientation changes are expected, a
\textit{kinematic} strategy could be adopted: rather than trying to
re-orient the base, we can consider to re-orient both the prismatic leg
and the landing mechanism, in order to align them with the wall.
Because of the presence of the ropes, most of the times the orientation errors are related to misalignment with
the vertical $Z$ axis of the base (see Fig.~\ref{fig:3dmodel2anchors_propellers}). 
\begin{figure}[t] 
	\centering
	\includegraphics[width=0.8\dimexpr\scalefactor\columnwidth]{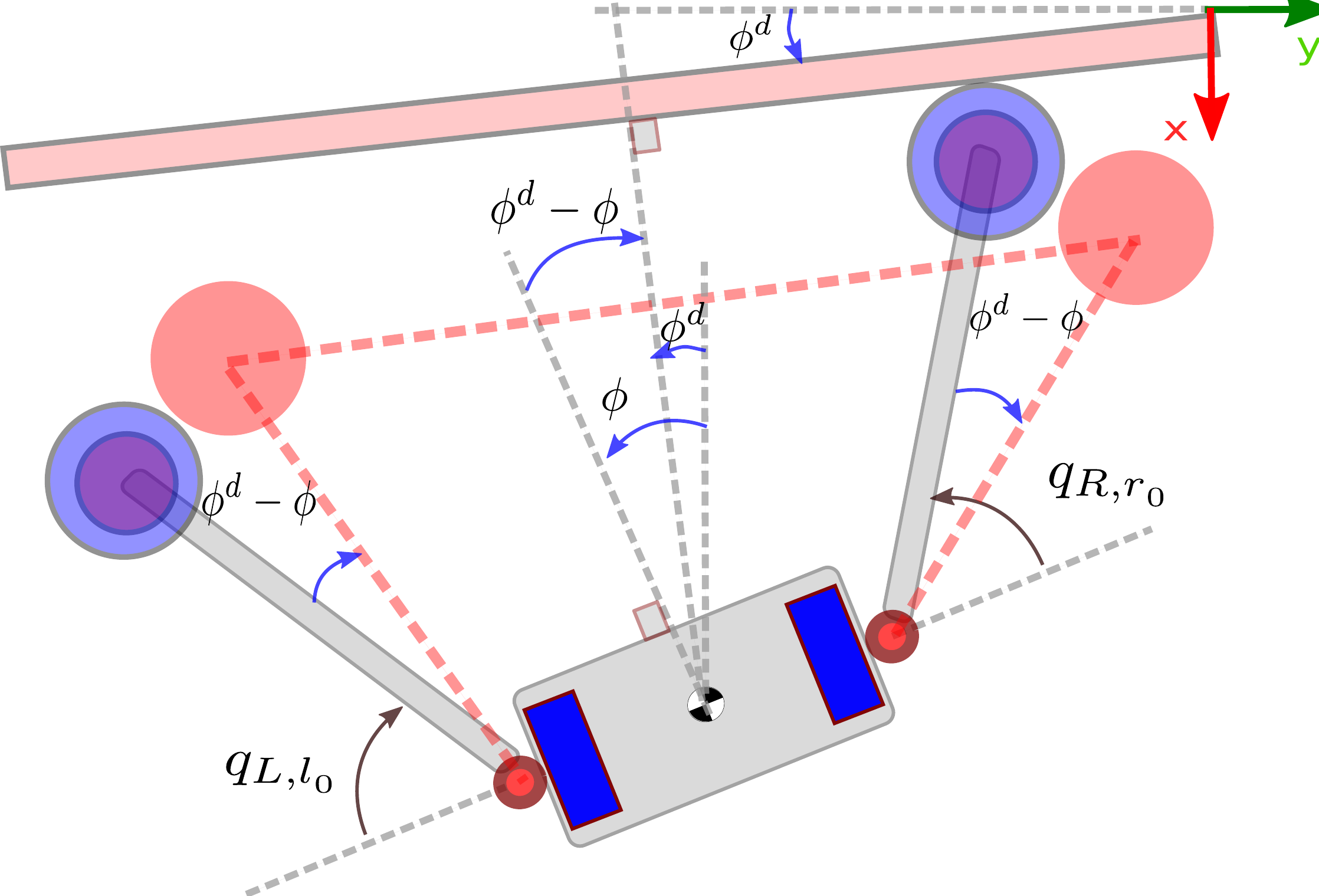}
	\caption{\small Kinematic strategy for reorientation of the landing system.}
	\label{fig:leg_reorientation}
\end{figure}
If we chose a $Z-Y-X$ sequence for the Euler angles to represent orientation, 
the orientation errors about the $Z$ axis can simply be associated
with the tracking error in the yaw direction: $\phi^d - \phi$.  To align the prismatic leg and
the landing links to the wall, it suffices to continuously adjust their
set-points as follows (see  Fig.~\ref{fig:leg_reorientation}):
\begin{align}
	q^d_{L,i} &= q_{{L,i}_0} + (\phi^d -\phi), \quad \quad\quad i\in\{l,r\}, 
\end{align}
where $\phi$ is the \textit{actual} value of the yaw angle that can be
obtained from the rotation matrix ${}_w\vect{R}_b$, which represents
the orientation of the base link:
\begin{align}
  \phi &= \arctan({}_w\vect{R}_{b_{21}}, {}_w\vect{R}_{b_{11}}) .
\end{align}
This approach is meant to have good performance in face of orientation errors up to $0.6$~rads.
This value is related to the maximum opening of the landing joints,
namely, when the landing link is aligned with the base link $Y$ axis.
%
\subsection{Flying motion control: landing}
\label{sec:landing_control}
The purpose of the landing mechanism is twofold: 1) to dissipate the
excess of kinetic energy at the landing, to avoid bounces 
and 2) to keep the robot firm on the wall for maintenance operations.
During the flight phase, the central \textit{thrusting} leg is retracted and the
landing legs are extruded in order to make contact with the wall
during the landing phase.  At impact, the kinetic energy
\begin{align}
	\tau^d_{L,i} &=  K_L(q^d_{L,i}- q_{L,i})   -D_L\dot{q}_{L,i}  \quad \quad\quad i\in\{l,r\}, 
    \label{eq:landing_strategy}
\end{align}
is dissipated through a joint impedance control law implemented for
the $q_{L,l}$ and $q_{L,r}$ joints, with stiffness $K_L$ and damping
$D_L$  selected to achieve a critically damped
behaviour. The damping value is set considering the reflected inertia
of the robot at the joint level.  
This control strategy avoids the
robot bouncing\footnote{A critically damped spring-mass-damper, is a second order system that has no overshoots to a step-response. The impact on the wall can be modelled  as a step input to the system and the impedance law act as a virtual spring-mass-damper acting in the direction normal to the wall. Being the contact unilateral  to  have bounces, it suffices that the velocity of the mass is negative (going out from the wall).  The effect of the control law that eliminates overshoots it reflects in the mass velocity being always positive (i.e approaching the wall) eventually becoming zero.  This ensure that there is no detachment from the wall (i.e. bounces).}, thus ensuring a constant contact. For the sake of
simplicity, we did not consider the influence of the ropes and gravity
force in the impedance law. This is reasonable since, at the impact,
the gravity forces and the rope forces are aligned with the landing
joint axis; this alleviates their adverse effects on the landing
joints (unless the wall is significantly slanted).  A more accurate
design of a Cartesian impedance controller could be implemented,
considering the full (constrained) robot
dynamics~\cite{mingo22parallel}, but this is left for future
developments.
%
Notice that, once the joint impedance law takes care of the kinetic
energy in the direction perpendicular to the wall, any residual
lateral effect can be dissipated with mechanical damping from the
landing wheels.
When the wall is nearly vertical, the effectiveness of the impedance-based landing strategy diminishes. In such cases, a promising approach could be to activate the propeller to push the robot against the wall, thereby eliminating any potential rebounds.
%

\subsection{Lateral manoeuvring}
\label{sec:lateral_maneuvering}
 In situations where the rock wall is free of obstacles (e.g., a clean slab), it is more efficient to navigate the wall locally by actively controlling the motion of the wheels. This approach can also help correct potential landing errors, enabling additional adjustments without requiring further jumps. For the current design, since the wheel is not steerable, only lateral motion is possible. Implementing a wheel-steering mechanism is left for future work.
 In any case, the motion of the wheels must be coordinated with the motion of the ropes to ensure kinematic consistency. To achieve this, we first compute the mapping (i.e., the Jacobian) between 
the base link linear $\dot{\vect{p}}$ and angular velocity $\boldsymbol{\omega}_b$, and the 
linear velocities of the landing wheel centres $\vect{v}_{l,l}$, $\vect{v}_{l,r} \in \Rnum^3$ ropes $\vect{v}_{r,l}$, $\vect{v}_{r,r} \in \Rnum$. 
Because the wheels $\vect{v}_{l,l}$, $\vect{v}_{l,r}$ are constrained to lie along the base axis $\vect{y}_b$, and the ropes 
velocity vectors lie along the rope axes $\vect{\hat{a}}_{r,l}$, $\vect{\hat{a}}_{r,r}$,  we can include these constraints in the  construction of the Jacobian matrix.
Then, referring to the rigid body model and its definitions in Fig. \ref{fig:static_definitions}, the Jacobian $\vect{J_{lm}}$ is a $4\times6$ and writes:

\begin{align}
\mat{v_{l,l}\\ v_{l,r}\\v_{r,l}\\ v_{r,r} } = 
\underbrace{\mat{ \vect{y}_b^T        &        \vect{y}_b^T   [-\vect{p}_{l,l}]_{\times} \\
                  \vect{y}_b^T        &        \vect{y}_b^T   [-\vect{p}_{l,r}]_{\times} \\ 
                  \vect{v}_{r,l}^T    &        \vect{v}_{r,l}^T  [-\vect{p}_{h,l}]_{\times} \\ 
                  \vect{v}_{r,r}^T    &        \vect{v}_{r,r}^T  [-\vect{p}_{h,r}]_{\times} } 
                  }_\vect{J_{lm}} 
                  \mat{\vect{\dot{p}}\\ \boldsymbol{\omega_b}}, 
\label{eq:lat_maneuv_control}                                    
\end{align}

where $v_{l,l}$,$v_{l,r} \in \Rnum$ are scalars representing the velocity of the centre wheels along $\vect{y}_b$ axis and the $v_{r,l}$,$v_{r,r} \in \Rnum$ the rope speed along the rope axes.  $[.]_{\times}$ is the skew-symmetric matrix associated to the cross product operator. 
By setting the desired values $\vect{\dot{p}}$ and $\boldsymbol{\omega_b}$ through \eqref{eq:lat_maneuv_control}, the set-points $v_{l,l}^d/R_w$, $v_{l,r}^d/R_w$, $v_{r,l}^d$,  $v_{r,r}^d$ can be computed. These set-points are then commanded to the PD controllers of the wheel and rope joints, respectively, where $R_w$ is the wheel radius.

\section{Simulation Results}
\label{sec:results}
This section presents simulation results demonstrating how the proposed solution package enables the robot to navigate along the mountain wall. The goal is to showcase the features outlined in Section~\ref{sec:motion_planning} and Section~\ref{sec:fly_motion_control}, which are essential for realizing the use case described in Section~\ref{sec:introduction}. Table \ref{tab:experiments} summarizes the experiments, highlighting the specific features demonstrated in each experiment and indicating which components are active.

\begin{table*}[!tbp]
	\centering
	\caption{Simulation Experiments}
    \resizebox{\textwidth}{!}{
       \begin{tabular}{l p{6cm} p{7cm} c c c c c} 
       \hline\hline
       \textbf{Exp. $\#$}    & \textbf{Description}    &  \textbf{Feature to validate} & \textbf{\gls{mpc}} & \textbf{Propeller} & \textbf{Land. Mech.} & \textbf{Fig./Tab.} \\
       \hline
       \multirow{2}{*}{Exp. 1} &    \multirow{2}{*}{Open Loop Model Validation (Sec. \ref{sec:model_validation})}  & \small{Validation of lower dimensional vs. full-detail model}  &   &    &      & Fig. \ref{fig:validation}  \\  
        \hline
       \multirow{2}{*}{Exp. 2}   &   \multirow{2}{*}{Disturbance tests   (Sec. \ref{sec:disturbance_tests}) }  &    \small{Evaluate the capability of the \gls{mpc} controller to reject disturbances} 
               &       \checkmark    &   \checkmark &         &               Fig.  \ref{fig:mpc_propellers}\\
       \hline   
       \multirow{2}{*}{Exp. 3}  &    Disturbance tests: Ablation Study (Sec. \ref{sec:ablation_study})    &   \small{Evaluate the capability of the \gls{mpc} controller to reject disturbances without propeller}  
              &     \checkmark    &              &         &      Fig. \ref{fig:mpc_dist_constant_no_propellers}, \ref{fig:mpc_dist_impulsive_no_propellers}\\
       \hline   
       \multirow{2}{*}{Exp. 4}  &    Disturbance tests: Robustness evaluation  (Sec. \ref{sec:disturbance_tests})    &   \small{Rejection of random disturbances}  
              &     \checkmark    &     \checkmark         &         &          Tab. \ref{tab:robustness_MPC_constant},  Fig.   \ref{fig:robustness_MPC_impulsive_ellipse_targets}\\
       \hline  
       \multirow{2}{*}{Exp. 5 }    & \multirow{2}{*}{Multiple targets  (Sec. \ref{sec:multiple_targets}) }  & \small{Test planner capability to do omni-directional jumps} 
                 & \checkmark & \checkmark &      & Tab.\ref{tab:multiple_targets}, \ref{tab:multiple_targets_energy}, Fig. \ref{fig:robustness_multiple}\\
       \hline  
       \multirow{2}{*}{Exp. 6 }    &  \multirow{2}{*}{ Obstacle avoidance  (Sec. \ref{sec:obstacle}) }   & \small{Test planner capability to do overcome obstacles of small/medium size} &  \checkmark &  \checkmark &   &    Video only         \\
       \hline    
       \multirow{2}{*}{Exp. 7}     &   \multirow{2}{*}{Landing test  (Sec. \ref{sec:landing_results})  }      &      \small{Test landing control capability to land without bounces} 
                &  \checkmark &  \checkmark &   \checkmark   & Fig. \ref{fig:landing_results}\\
       \hline\hline 					
       \multirow{2}{*}{Exp. 8}     &   \multirow{2}{*}{Slanted wall tests  (Sec. \ref{sec:slanted_wall})  }      &      \small{Test of jump from slanted walls of different inclinations} 
                &  \checkmark &  \checkmark &   \checkmark   & Only video \\
       \hline\hline 					
       \end{tabular}}
		\label{tab:experiments}
\end{table*}


\subsection{Simulation Setup }
\label{sec:gazebo_simulation}
%
We simulate the robot in a Gazebo environment, which computes  
the full robot (constrained) dynamics  
using the ODE physic engine~\cite{OdePhysicsEngine}. 
We employ the \gls{urdf} \cite{urdf} formalism to describe the 
robot model and the Pinocchio library 
\cite{carpentier2019pinocchio}  to compute the kinematic functions.
%
%
We always initialise the simulation in a way that 
the  closed loop kinematic constraints are satisfied at the startup  (i.e. at a joint configuration $\vect{q}_{\text{init}}$). 

The optimisation has been implemented in Matlab using the \texttt{fmincon} function.
To improve performance, we used a C++ implementation of the \gls{ocp}
both for motion planning and control, which 
is freely available\footnote{
  \href{https://github.com/mfocchi/climbing_robots2}{https://github.com/mfocchi/climbing\_robots2.git}}.
%

\subsection{Open Loop Model Validation (experiment 1)}
\label{sec:model_validation}
As in \cite{focchi23icra}, in a first experiment, we run the
optimisation to perform a jump from an initial position
$\vect{p}_0 = \mat{0.2& 2.5& -6}^T$~m to a target position
$\vect{p}_{tg} = \mat{0.2 & 4 &-4}^T$~m.
We validate the results in open loop for both the reduced model (in a Matlab
environment) and the full model (in Gazebo).  For the Matlab
simulation, we simply integrate~\eqref{eq:simplified_2ropes_minimal}
with an \texttt{ode45} Runge-Kutta variable-step scheme.
%
For the Gazebo simulation, we setup a state machine that orchestrates the 3 phases of the jump: leg orientation, thrusting and flying.
The offline optimisation is run before the leg orientation phase,
providing the values of the initial impulse ($\vect{f}_{\leg}^{*}$),
the pattern of the rope forces $\vect{F}_{r,l}^{*}$,
$\vect{F}_{r,r}^{*}$ and the jump duration $t_f^{*}$.  Since the
simulation runs at 1~kHz ($dt_{\text{sim}}$ = 0.001~s) while the
optimised trajectory has a different time discretisation
($d_t = t_f / N$), depending on the jump duration $t_f$, appropriate
interpolations are performed to adapt the rate difference.
During the thrusting phase, we generate the pushing impulse
$\vect{f}_{\leg}$ at the \gls{com} by applying a force $\vect{f}_c$ at
the contact point.

To avoid generating moments at the \gls{com}, we perform a preliminary
step (leg orientation phase) to align the prismatic leg to the
direction of $\vect{f}_c$.  To achieve this, we command the set-points
for hip roll and hip pitch joints to be
$q_{HR}^d = \atandue(\vect{f}_{\leg,y}, \vect{f}_{\leg,x})$ and
$q_{HP}^d = -\pi + \atandue(\vect{f}_{\leg,z},\vect{f}_{\leg,x})$.
Then, we employ the leg dynamics (last 3 rows of \eqref{eq:full_dyn}) to map the contact force
$\vect{f}_c = \vect{f}_\text{leg}$ into torques $\boldsymbol{\tau}_{\text{leg}}^d$ at the
leg joints:
\begin{equation}
	\boldsymbol{\tau}_{\text{leg}}^d= \mat{\tau_{HP} \\ \tau_{HR} \\ \tau_{K}} =  \vect{h}_{\text{leg}}(\vect{q}, \dot{\vect{q}}) -\vect{J}_{c,{\text{leg}}}^T(\vect{q}) {\vect{f}}_{c},
\label{eq:contact_force_mapping}
\end{equation}
where $\vect{J}_{c,{\text{leg}}} \in \Rnum^{3 \times 3}$ is the sub-matrix of the Jacobian $\vect{J}_c$, whose columns are relative to the leg joints, 
and $\vect{h}_{\text{leg}} \in \Rnum ^3$ represents the bias terms (Centrifugal, Coriolis, gravity). 
%

In general, a PD controller is superimposed to the feed-forward
torques $\boldsymbol{\tau}_{\text{leg}}^d$ to drive the leg joints
$q_{HR}, q_{HP},q_{K}$.  However, during the \textit{thrusting} phase,
only the feed-forward torques $\boldsymbol{\tau}_{\text{leg}}^d$ are
applied, while the PD controller is switched off. Finally, during the
\textit{flying} phase, the rope (prismatic) joints are actuated in
\textit{force} control mode.  The optimised force patterns
$\tau_{RP,l}^d = f_{r,l}^{*}$ and $\tau_{RP,r}^d = f_{r,r}^{*}$ are
set as reference forces for the whole jump duration $t_f$.  During the
flight, the landing mechanism and the prismatic leg are continuously
reoriented to be always aligned with the wall face, as explained in
Section \ref{sec:orientation_control}.

\begin{figure}[t] 
	\centering
	\includegraphics[width=1.0\dimexpr\scalefactor\columnwidth]{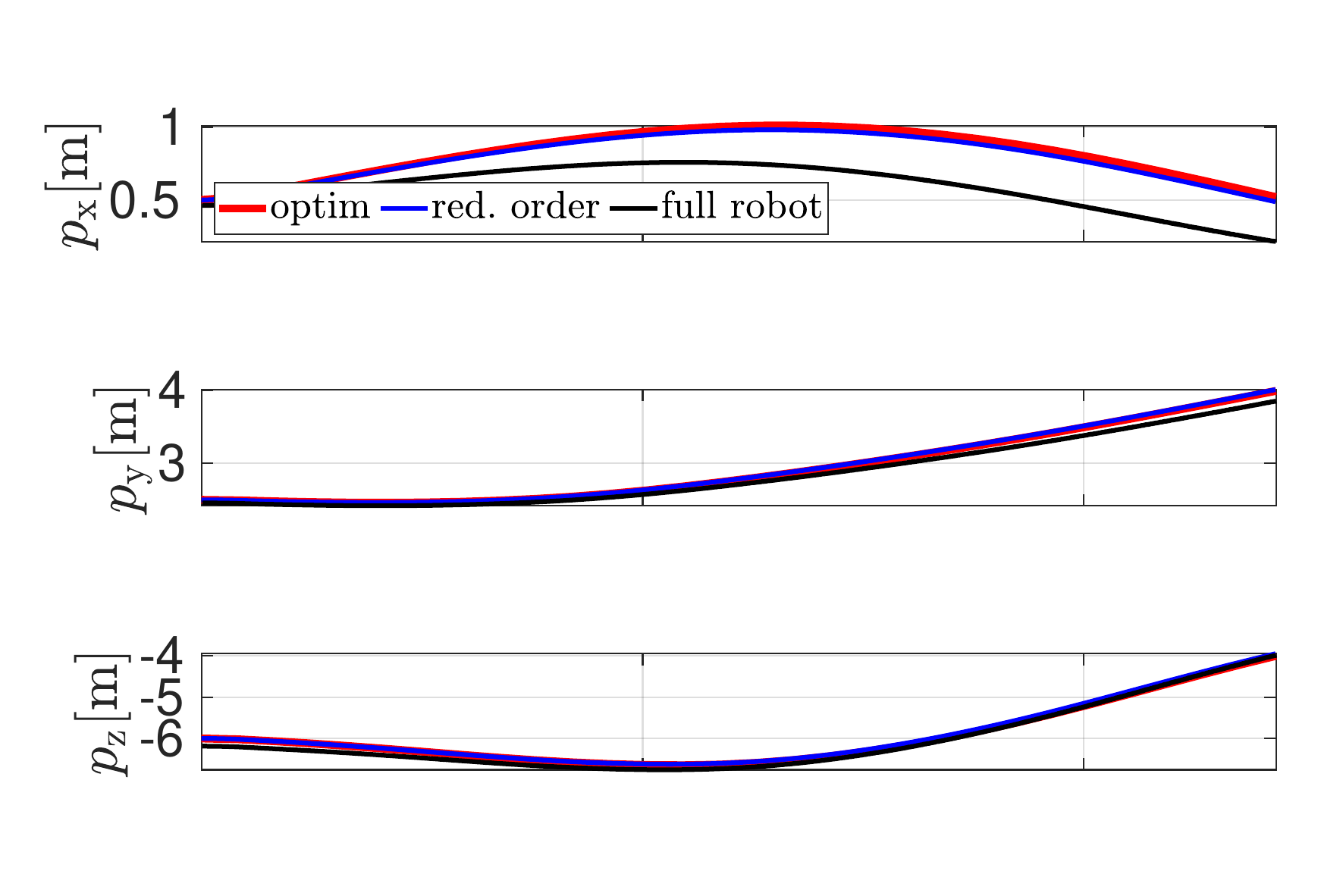}
	\caption{\small  Experiment 1. Open loop validation of the optimisation results.  
	Reference position for the \gls{com} from optimisation (red), simulated  trajectory with reduced-order model (blue) and full-robot model (black).}
	\label{fig:validation}
\end{figure}

Figure \ref{fig:validation} reports the results in \textit{open} loop.
Simulating the simplified model, the final error norm is below
$0.06$~m (due to integration errors), while in the full-model
simulation, it is around $0.34$~m.  These errors are in a range that
can be efficiently handled by a controller, as demonstrated in
Section~\ref{sec:mpc}.  However, the purpose of this validation is to
show that the proposed reduced model is a good approximation of the
real system and it is sufficient to generate feasible jumps
trajectories that do not involve large orientation changes.
%
%
Table \ref{tab:params} reports the value of the physical parameters used in the simulation, together with the  
optimisation settings, that achieve a good compromise between accuracy and convergence rate.

\begin{table}[t]
	\centering
	\caption{ Simulation and optimisation parameters}
\resizebox{\columnwidth}{!}{%
	\begin{tabular}{l c c  } \hline
		\textbf{Name} \quad          & \textbf{Symbol}                  & \textbf{Value}  \\ \hline
		Robot mass [kg]              & $m$  							& 5, 13 (with land. mech)                 \\ 
	    Anchor distance [m]          & $d_a$ 								& 5\\
		Left Anchor position [m] 		 & $\vect{p}_{a,l}$					& $ \mat{0 & 0&0}$\\
		Right Anchor position [m] 		 & $\vect{p}_{a,r}$					& $ \mat{0 &b&0}$\\
		Friction coeff.	 [] 			 & $ \mu $ 					        & 0.8   \\
		Contact normal				  & $n_c$ 								&$ \mat{1 & 0&0}$\\
        Wall normal			  & $n$ 								&$ \mat{1 & 0&0}$\\   
		Thrust impulse duration	[s]   & $t_{\text{th}}$  				        & 0.05\\		
		Discretisation steps  (NLP)   & $N$ 						        & 30\\	
		Simulation 	time interval [s] &  $dt_\text{sim}$                &0.001 \\
		Sub-integration steps         & $N_{\text{sub}}$						& 5  \\				        
		Hoist work weight            & $w_{\text{hw}}$ 						& 0.1\\	
		Smoothing weight			 & $w_s$ 							& 1 \\
		Integration method			 &          -                       &RK4 \\
		Max. (normal) leg force  [N]       	 & $f_{\leg}^{\text{max}}$	        & 300,  600 (with land. mech)    \\
		Max. rope force [N]    & $f_r^{\text{max}} $			    & 90, 300  (with land. mech)    \\
		Jump Clearance        [m]       & $c$					   	& 1 \\
        Slack       [m]                    &  $s$ & 0.02 \\
		Discretisation steps (MPC)     & $N_{\text{mpc}}$ 						& 50\\		
        Tracking term weight (MPC)            & $w_{p,\text{mpc}}$ 						& 1\\	
        Terminal cost weight (MPC)     &   $w_{pf,\text{mpc}}$ 						& 0\\	
		Smoothing weight (MPC)     &   $w_{u,\text{mpc}}$ 						& $10^{-5}$\\	
        Landing wheels radius     &   $R_w$ 						& $0.075$\\        
		\hline 					    					    					    
	\end{tabular}}
	\label{tab:params}
\end{table}

\noindent\textbf{Integration errors.}
To decrease the computational load of the optimisation process, we can
reduce the density of the trajectory's discretisation, which refers to
the number of specified points $N$, also known as knots.  This
reduction, however, results in integration errors, particularly in
single shooting settings.  First-order integration methods, such as
Explicit Euler, may not be sufficient to strike a good balance between
computational time and accuracy.  To improve accuracy, we found
beneficial using higher order integration schemes like Explicit
Runge-Kutta 4 (RK4) and to perform a number of $N_{\text{sub}}$
integration sub-steps of the dynamics within two adjacent knots.  We
considered that the rope forces remain constant on the time interval
$dt$ between two knots.

In Table \ref{tab:integration_errors} we evaluate different
combinations of: 1) number $N$ of discretisation nodes, 2) number of
sub-steps $N_{\text{sub}}$ and 3) integration method of different order 
(Euler or RK4). We benchmark on the same experimental jump employed in validation. 
We report the number of iterations needed to converge, the solution time, the absolute error
at the end of the trajectory with respect to the target
$\vect{e}_a = \vect{p}_{tg} -\vect{p}(t_f)$, and the error due to
integration $\vect{e}_i = \vect{p}_{gt} - \vect{p}(t_f)$, where
$\vect{p}_{gt}$ is the final position obtained by integrating with a
smaller time step of $0.1$~ms.
%
\begin{table}[t]
\centering
	\caption{Performances of different integration schemes  }
	\label{tab:integration_errors}
	\resizebox{\columnwidth}{!}{
	\begin{tabular}{c c c c c c c} 
		\hline
		\textbf{$N$} & \textbf{Method} & \textbf{$N_{sub}$} &  \textbf{Iters}  & \textbf{Comp, Time} [$s$] &  $\Vert \vect{e}_i \Vert $  &  $\Vert \vect{e}_a \Vert$ \\
   		\hline
   		 40    &     RK4     &  0   &    44  &  1.19 & 0.141 & 0.148 \\
   		 60    &     RK4     &  0   &    30  &  1.74 & 0.045 & 0.17\\    
   		 40    &     RK4     &  5   &    26  &  2.1  & 0.1 & 0.144\\
   		 40    &     EUL     &  0   &    28  &  0.28 & 0.83  & 0.85\\
   		 40    &     EUL     &  10  &    28  & 1.32  & 0.05  & 0.167 \\     		 
 \textbf{30}   &     RK4     &  5   &    34 &   1.71 & 0.037  & 0.134\\ 		
		 \hline			    
	\end{tabular}}	
\end{table}
From Table~\ref{tab:integration_errors} it is evident that both the
computation time and the integration error $\Vert \vect{e}_i \Vert$
are linearly proportional (directly and inversely) to the number of
discretisation points $N$.  Increasing either the number of sub-steps
$N_{\text{sub}}$ or the order of integration (i.e. using RK4 instead
of Euler) makes the single iteration slower but it makes an accuracy
improvement; instead, by reducing $N$, the accuracy decreases.  Adding
a certain number of integration sub-steps allows to reduce $N$ keeping
the accuracy unaltered. Setting $N$ too low the problem can become ill
conditioned and convergence might not be achieved.  As expected, being
RK4 a higher order method, it is superior than Forward Euler in terms
of accuracy and requires a lower number of knots $N$ to achieve the
same accuracy.
As a reasonable trade off, we selected the RK4 with $N=30$ and
$N_{\text{sub}}=5$ intermediate integration steps.

\subsection{Disturbance tests (experiment 2)}
\label{sec:disturbance_tests}
To evaluate the effectiveness of the \gls{mpc}  (Section \ref{sec:mpc}) in  
tracking and rejecting disturbances during the flight phase, 
we conducted a test similar to the one of the previous section, but in \textit{closing} 
the  loop with  the \gls{mpc} controller   presented in Section \ref{sec:mpc} in  presence of disturbances. 
We set $dt_\text{mpc}$ equal to the
optimal control time step $dt$ and  $N_{\text{mpc}} = 0.4N$.  Since the \gls{mpc} optimisation
takes on average $0.3$~s to be computed, to avoid delays in the
simulation, we paused the simulation during the optimisation. We
postpone to future work a customised C++ implementation more
performing than the C++  code generated from Matlab.

\begin{figure}[t]
  \centering
  \includegraphics[width=1.0\dimexpr\scalefactor\columnwidth]{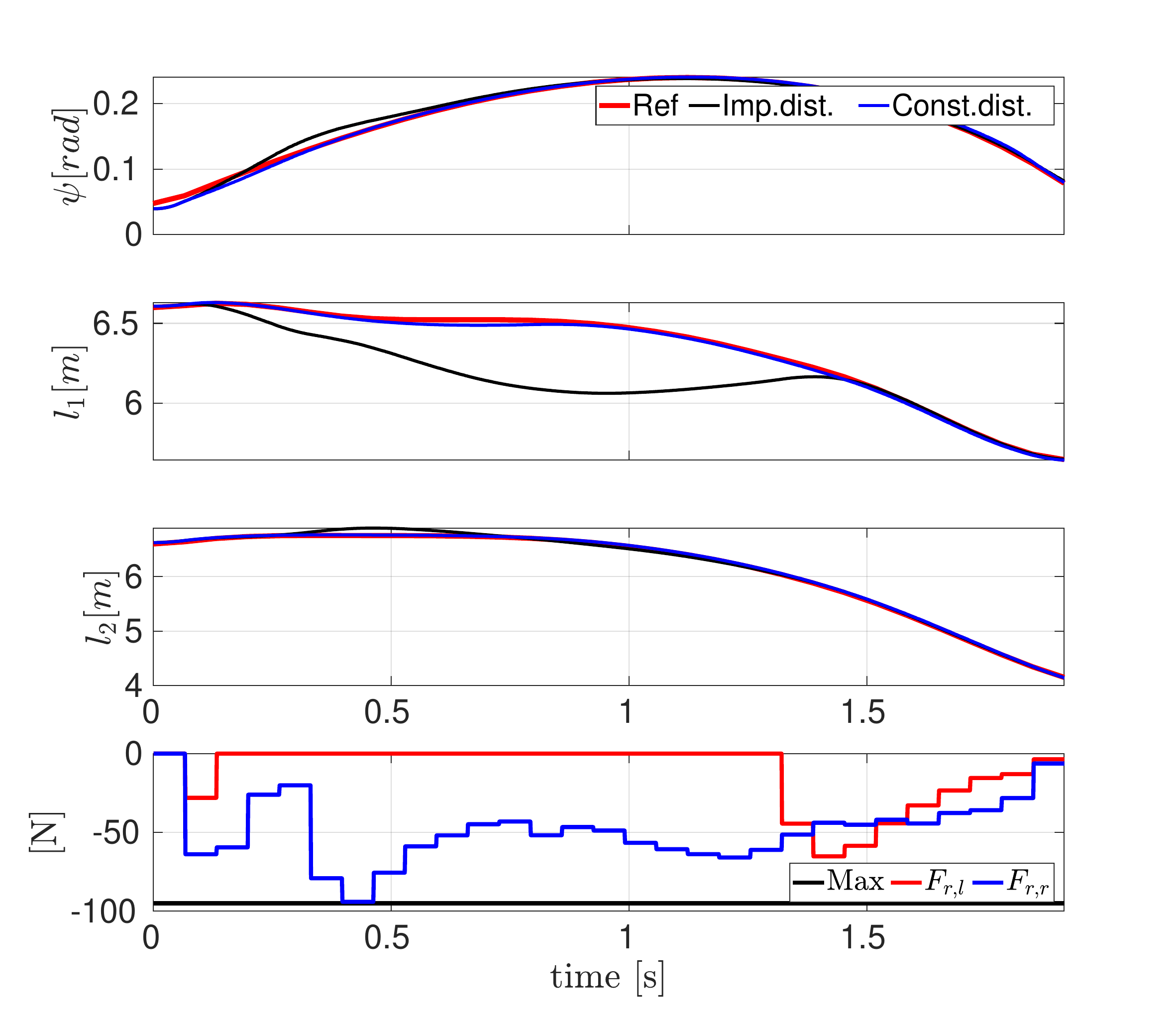}
  \caption{\small \textit{Experiment 2}. Simulation with \gls{mpc}. Tracking of the state
    variables in face of an impulsive
    $\boldsymbol{\delta}_i= [50, -50, 30]^T$~N (black) and constant
    disturbance (blue) $\boldsymbol{\delta}_c= [7, -7, 0]^T$~N.  The
    last plot shows the rope forces in the case of the impulsive disturbance.}
  \label{fig:mpc_propellers}
\end{figure}

Figure \ref{fig:mpc_propellers} presents the tracking plots for the $3$
reduced states $(\psi, l_1, l_2)$ when both an impulsive and a
constant disturbance are applied to the robot base. The impulsive
disturbance is applied after lift-off and persists for $0.2$~s, while
the constant disturbance is applied during the whole jump.  This test
has the purpose to emulate the effect of wind\footnote{Given the form factor of the robot, a 30 km/h wind translates into a
  7 $N$ disturbance force.}.
%
The \gls{mpc} is able to compensate the impulsive disturbance only
after $1.4$~s, reaching the target with a cumulative error of $0.072$~m.
The error during the transient in the impulsive case is related to the
presence of the unilateral constraint ($f_{r,i}\leq0$), which is hit
by the left rope.

Thanks to the predictive capability of \gls{mpc}, despite the
saturation of the control inputs, the tracking error is eventually reduced.
Notably, the impulsive disturbance test reveals that the disturbance
it is more promptly rejected on the $\psi$ variable compared to the
other states.  This can be attributed to the assumption of having
bilateral actuation in the propeller thrust forces (i.e. no unilateral
constraint).  In the case of constant disturbance, the controller
proficiently compensated the disturbance also during
the transient, showing a landing error of $0.073$~m.

\subsection{Disturbance tests: Ablation Study (experiment 3)}
\label{sec:ablation_study}
\begin{figure}[t]
	\centering
	\includegraphics[width=1.0\dimexpr\scalefactor\columnwidth]{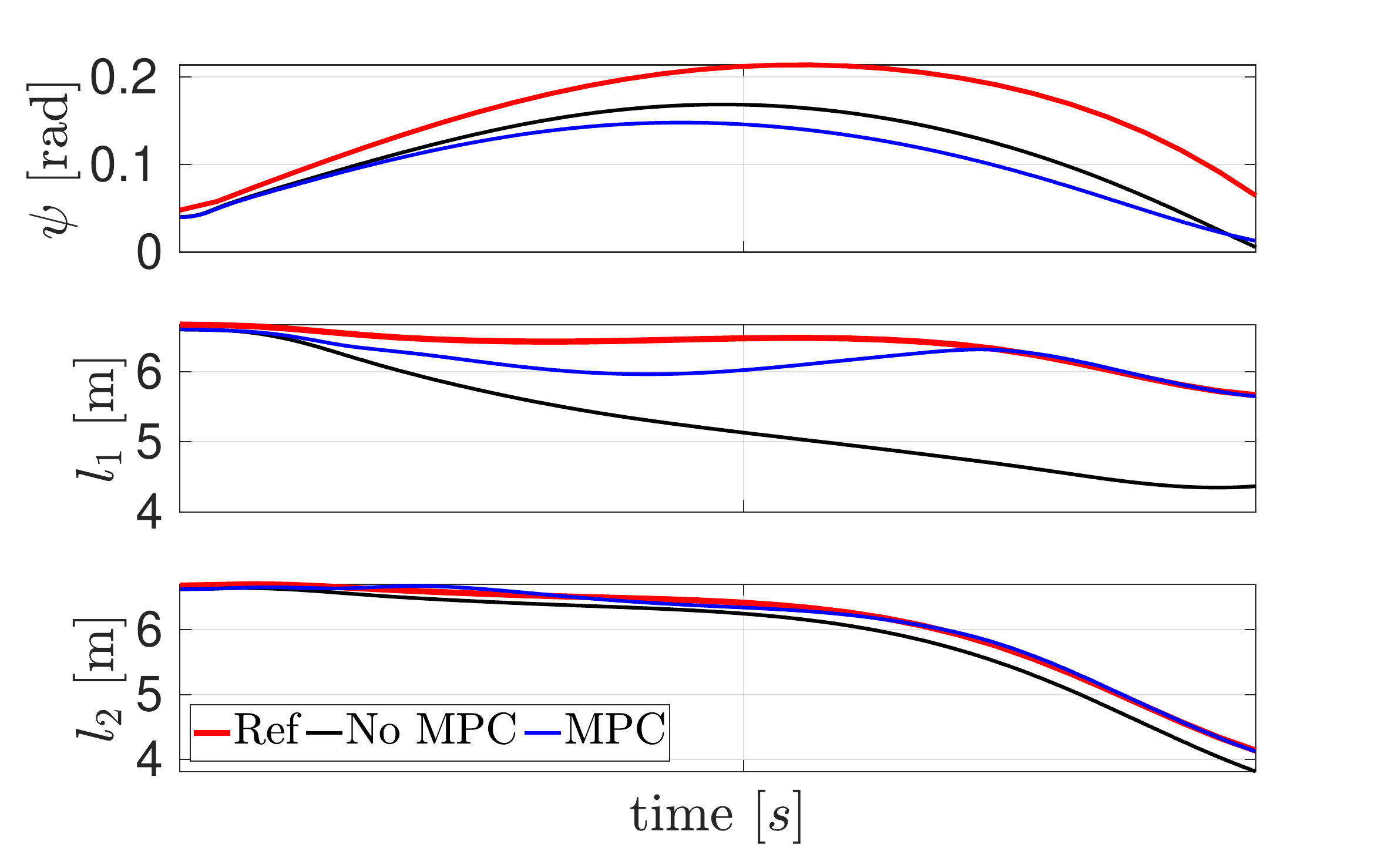}
	\caption{\textit{Experiment 3}. Ablation study. Simulation with \textbf{impulsive} 
        disturbance $\boldsymbol{\delta}_i$ without propellers. }
	\label{fig:mpc_dist_impulsive_no_propellers}
\end{figure}

\begin{figure}[t]
	\centering
	\includegraphics[width=1.0\dimexpr\scalefactor\columnwidth]{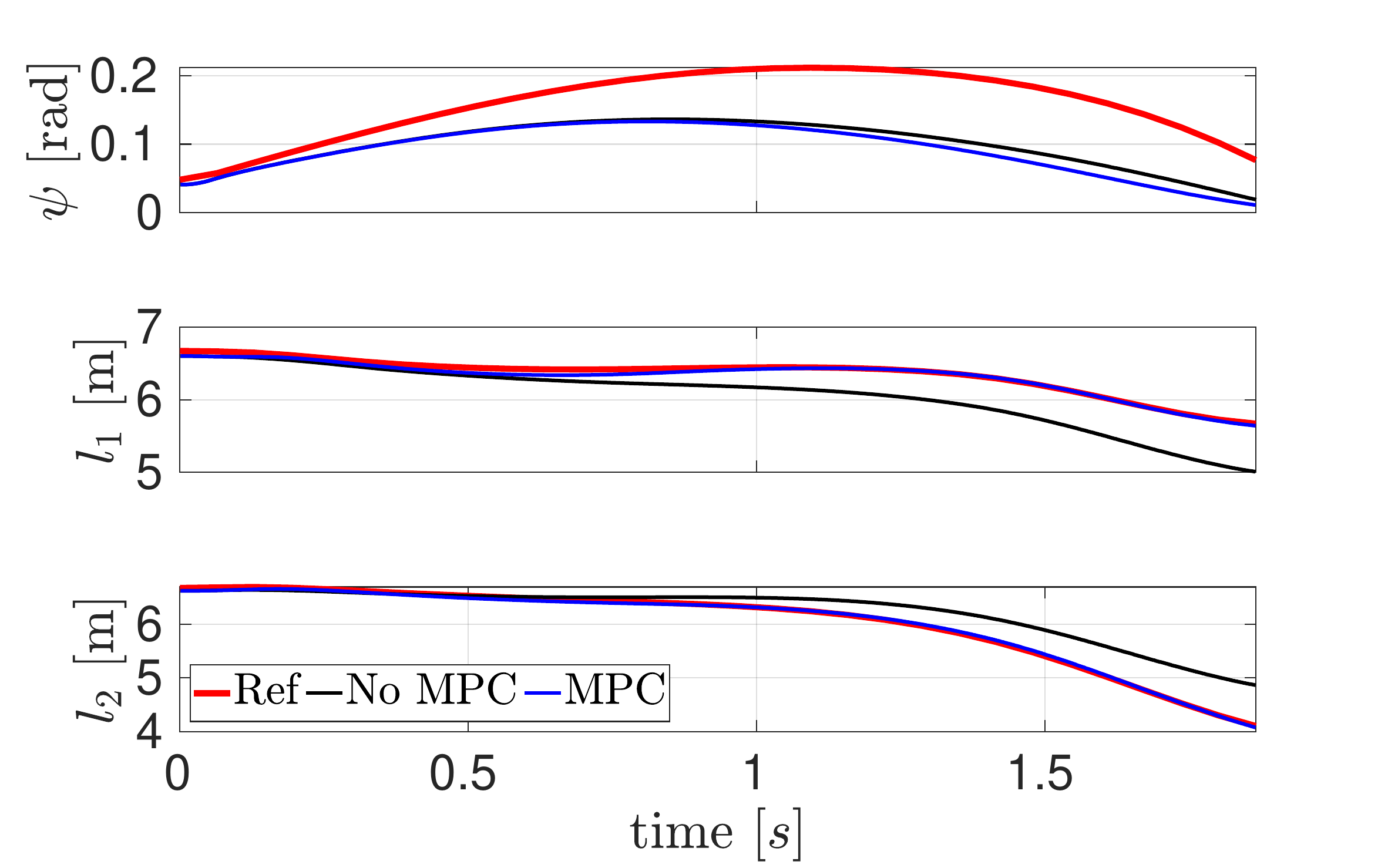}
	\caption{\textit{Experiment 3}. Ablation study. Simulation with \textbf{constant} disturbance $\boldsymbol{\delta}_c$ without propellers.}
	\label{fig:mpc_dist_constant_no_propellers}
\end{figure}

In this paragraph, we want to evaluate how performance degrades when
the propeller is not present (e.g. a propeller is not installed for
cutting costs). We repeated the last tests switching off either the
propeller or the whole \gls{mpc} controller.  The results are reported
both in Fig.~\ref{fig:mpc_dist_impulsive_no_propellers} and
Fig. \ref{fig:mpc_dist_constant_no_propellers}.  The plots reveal that
the errors are completely recovered (the \gls{mpc} controller is still
active on the ropes) only on $l_1$ and $l_2$ (with a cumulative error
of $0.06$~m for $l_1$ and $l_2$ at landing), while there is a remarkable
worsening in the tracking of $\psi$.
This is related to the fact that the system is under-actuated and the
rope forces are constrained on a plane.  Therefore, errors on $\psi$
cannot be easily recovered.

%


\subsection{Disturbance tests:~Robustness Evaluation~(experiment~4)}
\label{sec:robustness_evaluation}
\begin{figure}[t]
	\centering
	\includegraphics[width=0.49\dimexpr\scalefactor\columnwidth]{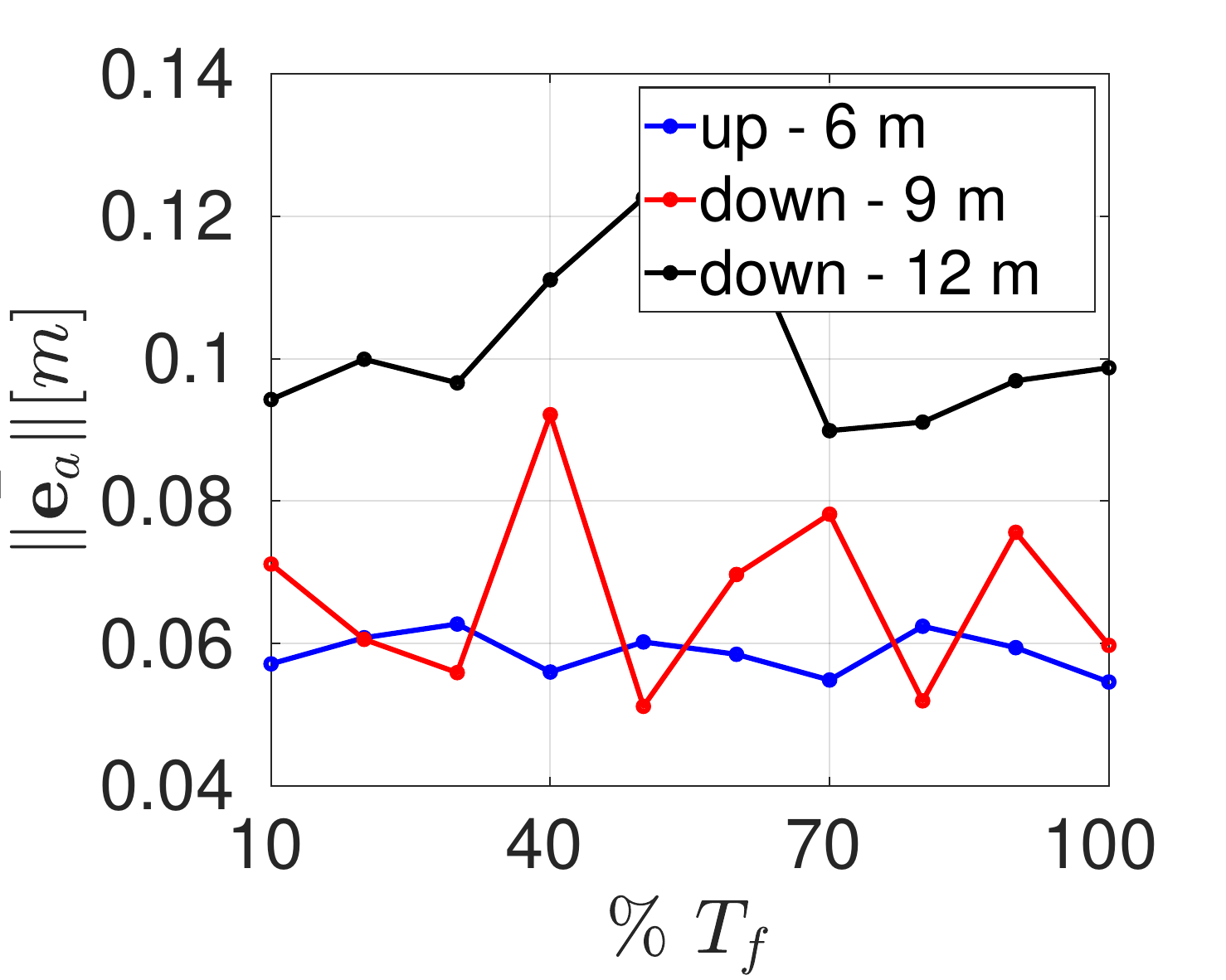} 
    \includegraphics[width=0.49\dimexpr\scalefactor\columnwidth]{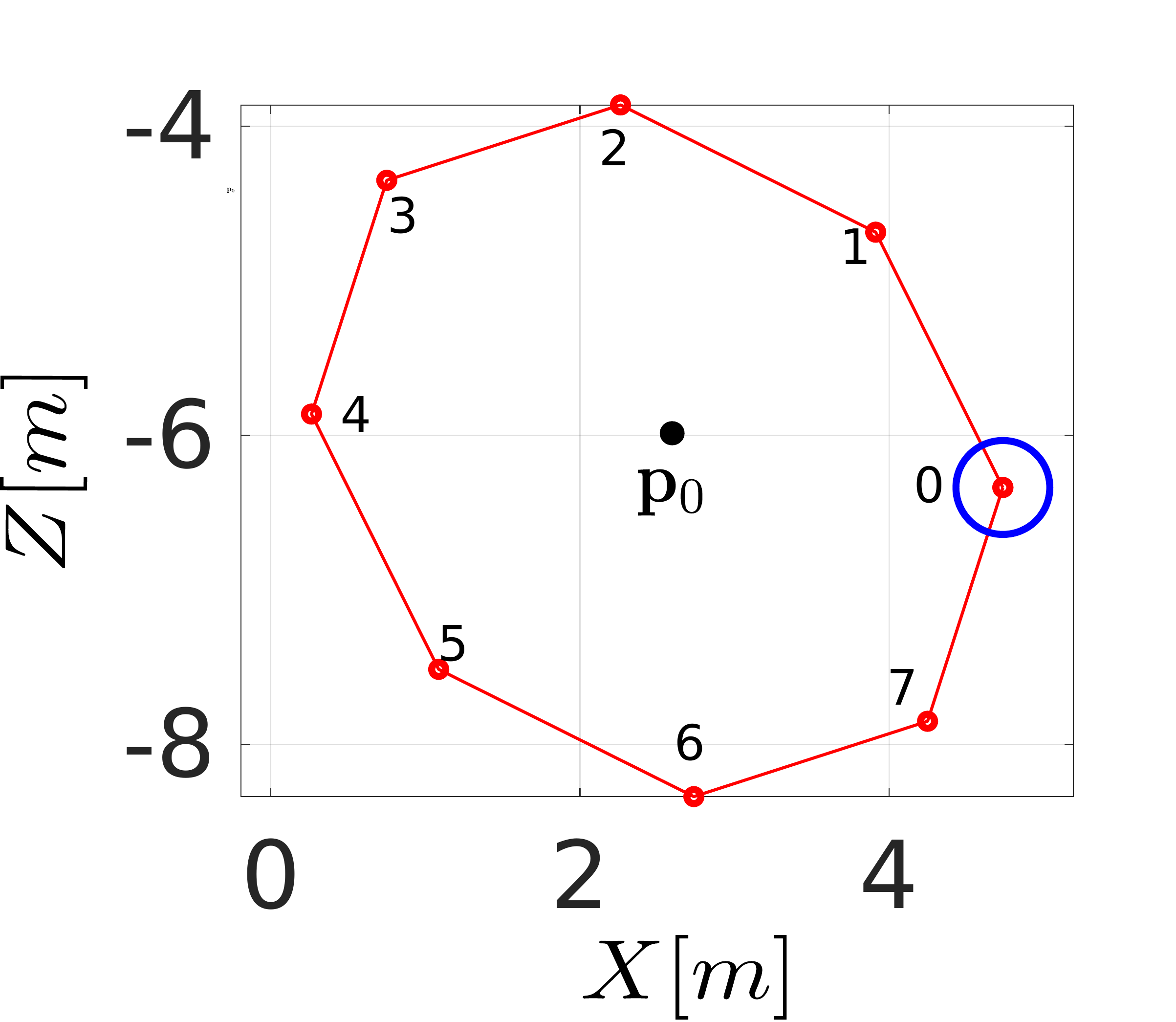}
    \caption{\small (left) \textit{Experiment 4}. Mean of the norm of the landing error
      $\Vert {\vect{e}}_a\Vert$ for a random impulsive disturbance
      applied at different moments of the flight phase for jumps involving different rope lengths:  short (blue), medium (red) and long (black).  
      The duration of the flight phase is normalised: 0 is lift-off and 100 is
      touch-down moment.  (right) \textit{Experiment 5}. Set of targets to evaluate jumps of
      variable length and directions.  The first target is the
      rightmost (blue), the others in the list are in CCW
      order. }
	\label{fig:robustness_MPC_impulsive_ellipse_targets}
\end{figure}

\noindent \textbf{Impulsive disturbance:} In this section, we test the controller under random \textit{impulsive}
disturbances applied at different moments of the flight phase.  We
repeat 100 tests with random disturbances with amplitudes between
$25$~N and $50$~N.  We limit the disturbance to a downward hemisphere
1) to avoid unloading the ropes and 2) because it is more
representative of a real situation (e.g. rock-falling from above).
To quantitatively evaluate the tracking performance during the flight
phase, we completely retract the prismatic leg to avoid early or
delayed touch-down and let the simulation stop at the end of the
horizon ($t=t_f$), where we evaluate the difference between the
position of the robot and the desired target. This discrepancy is
referred to as the \textit{landing error} $\vect{e}_a$.  We
discretised the flight phase in 10 intervals of equal duration. For
each interval we performed 10 tests applying a randomised disturbance in that
interval.  In Fig. \ref{fig:robustness_MPC_impulsive_ellipse_targets}
(left) we report $\Vert {\vect{e}}_a\Vert$ as a function of the moment
the random disturbance is applied. 
Since the result can be different for different rope lengths, we tested this for  
$\vect{p}_{tg} = \mat{0.5 & 4 & -4}^T$ (short), $\vect{p}_{tg} = \mat{0.2 & 4 &-9}^T$ 
(medium) and  $\vect{p}_{tg} = \mat{0.2 & 4 &-12}^T$ (long) jumps, 
that correspond to rope lengths at the landing of 6, 9 and 12 m, respectively.
As expected, the absolute error decreases with shorter rope lengths but exhibits more erratic behaviour during downward jumps. We hypothesize that this is because the \gls{mpc} controller has limited room for adjustment when the rope forces approach the upper bound (see the discussion in Section~\ref{sec:slanted_wall}). Overall, the norm of the absolute error, $\Vert {\vect{e}}_a\Vert$, remains constrained within a narrow range, indicating the controller's ability to reject various constant disturbances from different directions.

\noindent \textbf{Constant disturbance:} Regarding  the constant disturbance, we repeated the experiment 
for the 6 directions that represent the vertices of a cube, with a constant amplitude of 7 N.
The norm of the landing error $\Vert \vect{e}_a\Vert$ is reported in Table \ref{tab:robustness_MPC_constant}. 
Except for the $\mat{0 &1 &0}$ direction, they have the same order of magnitude, which is an indicator of the capability of the controller to reject a variety of constant disturbances of different directions. 
A disturbance in the direction $\mat{0 &1 &0}$ ($Y$) is problematic for this specific jump because 
it would require mainly the action of the left rope to be rejected, which is almost 
unloaded in the case of the designated landing point $\vect{p}_{tg}$.
\begin{table}[ht!]
	\centering
	\caption{\textit{Experiment 4}. Tracking errors with constant disturbance}
			\begin{tabular}{l c} \hline\hline
			\textbf{Dist. direction}       &   $\Vert \vect{e}_a\Vert $ [$m$] \\ \hline			
			         $\mat{1 &0 &0}$           &       0.0512                   \\
                  $\mat{-1 &0 &0}$           &       0.0539                 \\
                  $\mat{0 &1 &0}$           &        2.4885                \\
                  $\mat{0 &-1 &0}$           &       0.2109                 \\                  
                  $\mat{0 &0 &1}$           &        0.0835                \\
			       $\mat{0 &0 &-1}$           &       0.1384                  \\
            \hline\hline 					    					    					    
	\end{tabular}
	\label{tab:robustness_MPC_constant}
\end{table}
\subsection{Multiple targets: omni-directional jumps (experiment 5)}
\label{sec:multiple_targets}
We now aim to demonstrate the effectiveness of the control inputs
obtained from the \textit{offline} optimisation process presented in
Section~\ref{sec:motion_planning} in manoeuvring the robot to reach
various target locations.

It is important to highlight that while we consider the wall to be
vertical, a similar approach has also been proven to work for jumps from
\textit{non-vertical} (see Section \ref{sec:slanted_wall}).
To showcase the capabilities of our approach, we have generated a
total of 8 target points, evenly distributed on an ellipse around the
$\vect{p}_0$ location on the wall.  The main axis of the ellipse measures
$2.5$~m and is inclined at an angle of $45$~degrees with respect to
the horizontal plane.  The minor axis, on the other hand, spans a
length of $2$~m (refer to
Fig.~\ref{fig:robustness_MPC_impulsive_ellipse_targets}(right) for a
visual representation).  This has the purpose to evaluate how good the
optimisation generalises to different jump \textit{lengths} and
\textit{directions}.
%
%
The simulation results are reported in Table \ref{tab:multiple_targets} 
and in the accompanying
video\footnote{Video of experiments available at \href{https://youtu.be/FqsREaoe-28}{\texttt{https://youtu.be/FqsREaoe-28}}}.

We compute the energy consumption $E$ for each jump, considering the energy
consumed in the pushing  impulse and in winding/unwinding the ropes:
\small{
\begin{align}
  & E = \frac{1}{2}m\Vert\vect{\dot{p}}(t_{\text{th}})\Vert^2 +
    \int_0^{t_f} \left( \vert f_{r,l}(t)\dot{l}_1 \vert + \vert
    f_{r,r}(t)\dot{l}_2 \vert \right) dt ,
    \raisetag{1\normalbaselineskip} 
\end{align}}
where the former is proportional to the kinetic energy of the robot
evaluated at the lift-off ($t=t_{\text{th}}$) (assuming the robot
starting standstill).  The second term is the energy consumed by the
hoist motors as the integral of the power along the whole jump
duration $t_f$.  The hoist work, in general, dominates. In a first
batch of tests we set the weight relative to the hoist work
$w_{hw} = 0$ in \eqref{eq:NLP_cost}.  In another set of experiments
we set a non zero weight for the hoist work $w_{hw} = 100$.
To demonstrate the robustness of the approach in a real application
where states derivatives are usually obtained by numeric
differentiation of encoder readings, we added a zero-mean white
Gaussian noise to the state with standard deviation
$\boldsymbol{\sigma}= \mat{0.01 \, \mbox{rad/s} & 0.2 \, \mbox{m/s} &
  0.2 \, \mbox{m/s}}$.

Table \ref{tab:multiple_targets} reports \textit{mean} and \textit{standard deviation} of the norm of the absolute landing error $\vect{e}_a = \vect{p}_{tg} - \vect{p}(t_f) $ and the energy consumption $E$ for each jump.
To better appreciate the impact  of noise we also report between parenthesis the values without noise.
\begin{table}[t]
    \centering
    \caption{\textit{Experiment 5}. Tracking error and energy consumption without hoist work penalisation}
   	\renewcommand{\arraystretch}{1.1}
    \resizebox{\columnwidth}{!}{
        \begin{tabular}{c c c c} \hline\hline        
            \textbf{Exp.}       & \textbf{$ \vect{p}_{tg}$ [m]}  & $\Vert \vect{e}_a\Vert $ [m]  &  $E$ [J] 	               \\    
            0				& $\mat{0.28 & 4.73 & -6.34}$      &   (0.6)  0.9 $\pm$ 0.19        &  (103) 162  $\pm$ 111                        \\
            1				& $\mat{0.28 & 3.91 &-4.69}$       &  (0.078) 0.3 $\pm$ 0.008      &  (119)   129  $\pm$ 4      	 \\
            2				& $\mat{0.28 & 2.26 &-3.86}$       &  (0.079) 0.4084 $\pm$ 0.16    &  (200)   147 $\pm$ 16         	   \\
            3				& $\mat{0.28 & 0.75  &  -4.35}$    &  (0.081) 1.5 $\pm$   0.95     &  (287) 607 $\pm$ 317           \\
            4				& $\mat{0.28 & 0.26 &-5.86}$       &  (0.066) 0.17  $\pm$ 0.007     &  (228)  199 $\pm$ 5               \\
            5				& $\mat{0.28 & 1.08& -7.51}$       &  (0.067) 0.24  $\pm$ 0.06     &  (261) 214  $\pm$ 45                     \\
            6				& $\mat{0.28 &  2.73 &-8.34}$      &  (0.08)  0.2653   $\pm$ 0.044  &  (32)  202   $\pm$ 71               \\ 
            7				& $\mat{0.28 &   4.25   & -7.85}$    &  (0.32) 0.62 $\pm$ 0.27     &  (78)  152 $\pm$ 108     \\	
            \hline\hline 					    					    					    
    \end{tabular}}
    \label{tab:multiple_targets}
\end{table}
Interestingly, the highest errors are with test 0 and 7, which are
the targets that are closer to the vertical of the right anchor
(i.e. at $\vect{p}_{{a,r}_y}= 5$~m).  This is reasonable because in
these tests one of the two ropes is almost unloaded and cannot control
properly the error in the $Y$ direction.  This suggests that a
reduction in performance occurs when one of the ropes is almost
unloaded.  For the other tests, the error is on average around
$0.07$~m.  The noise results in higher errors with a similar trend,
except for test 3, which resulted in a significantly higher average
error of $2.22$~m.  Inspecting the collected data for that target, we noted that the
error had an erratic behaviour being very high only in some tests that
increase the overall mean, while others are in the same range as 1, 2,
4, 5, 6. We conjecture that this is related to the fact that test 3,
being an upward jump, was exceptionally demanding for the left rope
that was working at the boundary of its physical actuation limits.

Regarding the consumed energy, this is low for jumps to a lower target
(i.e. 6, 7), because the gravity helps and the ropes can be let to
passively unwind the ropes to attain the target. However, it was
surprisingly high for tests 3, 4, and 5.  The reason is that, because
there was no penalisation in the cost function, the optimiser found
elaborated trajectories to reach the target maximising accuracy
disregarding the energy.
Therefore, we repeated the previous tests penalising, this time, the hoist work. 
The results are reported in Table~\ref{tab:multiple_targets_energy}.

\begin{table}[t]
\centering
\caption{\textit{Experiment 5}. Tracking error and energy consumption with hoist work penalisation}
\renewcommand{\arraystretch}{1.1}
\resizebox{\columnwidth}{!}{
    \begin{tabular}{c c c c } \hline\hline        
        \textbf{Exp.}       & \textbf{$ \vect{p}_{tg}$ [m]}  &   $\Vert \vect{e}_a\Vert $ [m]  &  $E$ [J] 	\\
        0				& $\mat{0.28 & 4.73 & -6.34}$    &  (0.4) 0.4       $\pm$ 0.244      & (19.4) 152 $\pm$ 125   \\
        1				& $\mat{0.28 & 3.91 &-4.69}$     &  (0.067) 0.068   $\pm$ 0.0315	 & (54 )  120 $\pm$ 4 \\
        2				& $\mat{0.28 & 2.26 &-3.86}$     &  (0.071)	0.064   $\pm$ 0.278	     & (124)   171$\pm$ 42  \\
        3				& $\mat{0.28 & 0.75  &  -4.35}$  &  (0.077) 2.22    $\pm$ 1.21	     & ( 97)  568 $\pm$ 210    \\
        4				& $\mat{0.28 & 0.26 &-5.86}$     &  (0.07) 0.06     $\pm$ 0.023      & (45 ) 200  $\pm$ 11 \\
        5				& $\mat{0.28 & 1.08& -7.51}$     & (0.095) 0.06     $\pm$ 0.018      & (14 )  224 $\pm$ 54    \\
        6				& $\mat{0.28 &  2.73 &-8.34}$    &  (0.12) 0.053    $\pm$ 0.024      & (10 )236$\pm$ 75 \\ 
        7				& $\mat{0.28 &   4.25   & -7.85}$& (0.28) 0.23 	    $\pm$ 0.16		 & (13 )  160 $\pm$ 117 \\	
        \hline\hline 					    					    					    
\end{tabular}}
\label{tab:multiple_targets_energy}
\end{table}

\begin{figure}[t]
	\centering
	\includegraphics[width=0.7\dimexpr\scalefactor\columnwidth]{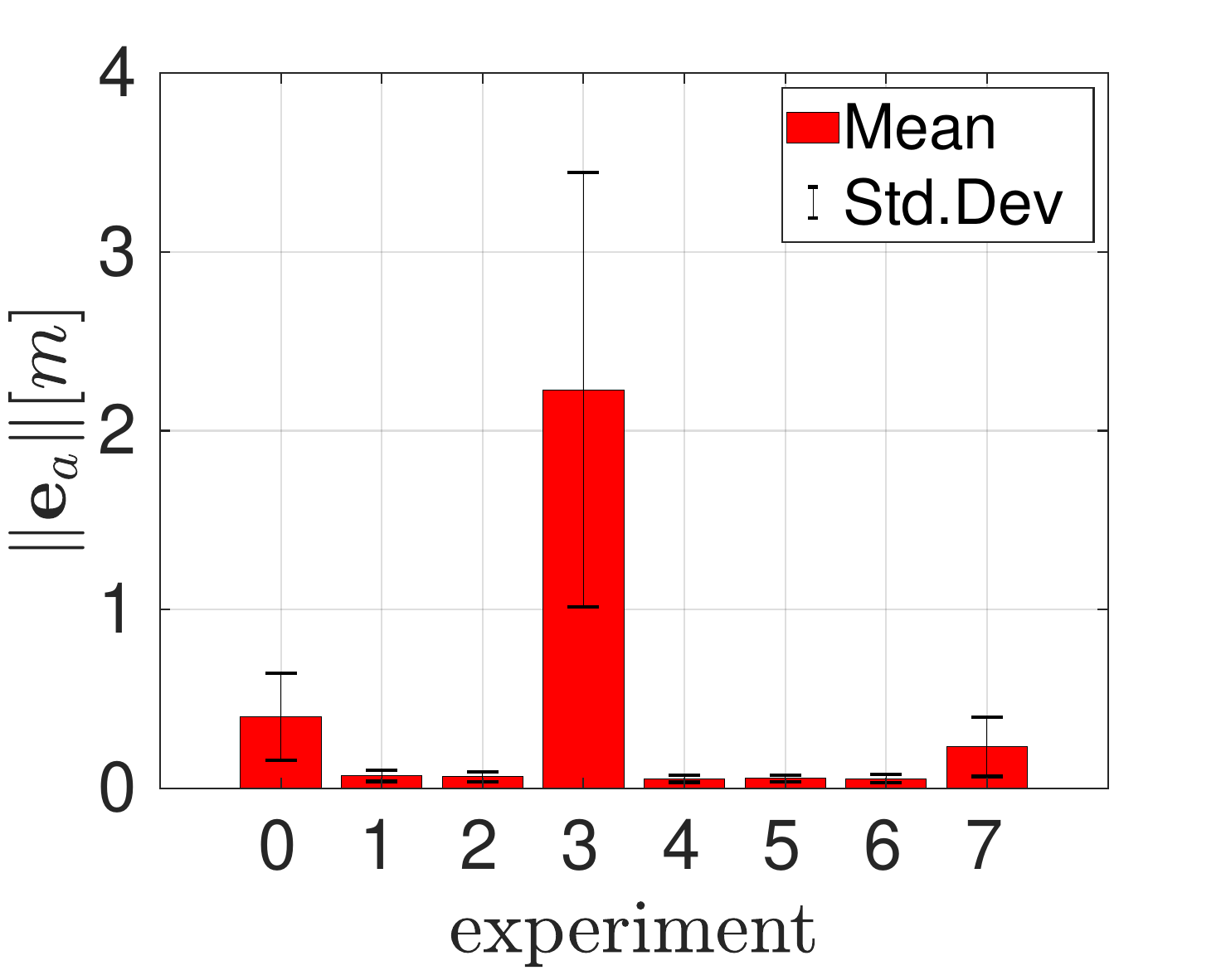}
	\caption{\small \textit{Experiment 5}. Mean and standard deviation for 50 repetitions of each test in Fig. \ref{fig:robustness_MPC_impulsive_ellipse_targets} (with hoist work penalisation)  adding white Gaussian Noise to the state derivatives.}
	\label{fig:robustness_multiple}
\end{figure}

In this case, the optimiser managed to find solutions that are more
energy efficient, with similar accuracy.  The trend is now inline with
the physical intuition that targets above $\vect{p}_0$ (i.e. 1, 2, 3)
are more energy demanding than the ones below (i.e. 5, 6, 7).  The
impact of noise (c.f. Fig. \ref{fig:robustness_multiple}) is similar to the previous case.
To highlight the versatility of the approach, in the accompanying
video, we conducted additional tests by placing the anchor points at
different locations, specifically 5, 7, 9, and 11 meters apart.
Despite these varying placements, our optimisation algorithm still
successfully produces valid results.  We observed that the proposed
method yields different solutions for the same target based on the
anchor location, yet in all cases a level of landing accuracy
comparable with the previous experiments is achieved.
\subsection{Obstacle avoidance (experiment 6)}
\label{sec:obstacle}
In this section we test capability of the optimal planner to deal with
a non-flat wall surface. We model an obstacle as a
hemi-ellipsoid 
whose semi axes are $R_x = 1.5$~m, $R_y = 1.5$~m and $R_z = 0.87$~m,
with centre $\vect{o}=\mat{-0.5& 2.5& -6}^T$~m\footnote{Note,
  that the obstacle shape is arbitrary. We have already shown in
  \cite{focchi23icra} an example with a conic pillar.  Any 3D map of
  the wall face would be suitable given that a convex representation
  is provided.}.
%
The equation of a generic ellipsoid in standard form writes:
\begin{align}
	\frac{(x - \vect{o}_x )^2}{R_x^2} + 	\frac{(y - \vect{o}_y )^2}{R_y^2} + 	\frac{(z - \vect{o}_z)^2}{R_z^2}  = 1.
	\label{eq:ellipsoid}
\end{align}
To implement obstacle avoidance we set a path constraint $f(\vect{x})>0$ on the $X$ component of the robot position: $\vect{p}_x > \hat{x} + c$.
Where $c$ is a clearance margin that should be kept from  the obstacle, for safety.
$\hat{x} \in \Rnum$ is a function $f(\vect{p}_y, \vect{p}_z)$ that can be easily obtained from the ellipsoid equation:
\begin{align}
   \begin{aligned}   
   &\hat{x} = \vect{o}_x + \sqrt{q},\\
   & q= R_x^2 - \frac{R_x}{R_y} ( \vect{p}_y - \vect{o}_y)-  \frac{R_x}{R_z} ( \vect{p}_z - \vect{o}_z)^2.\\
    \end{aligned}
\end{align}
To ensure that the constraint is enforced only for positions in the
convex hull of the ellipsoid (where the square root is defined), we
check if the argument $q$ of the square root is positive, otherwise we
consider the usual wall constraint \eqref{eq:wall_constraint} where, for vertical wall, $\vect{n}^T\vect{p} = \vect{p}_x$:
\begin{align}
	&\begin{cases}
	 \vect{p}_x >  \hat{x} + c,  \quad & q > 0, \\
	 \vect{p}_x >  \epsilon,     \quad& q < 0 .  
	\end{cases}
\end{align}
Considering this constraint, and setting a clearance $c=1$~m, we
optimise a jump starting from a point $\vect{p}_0=\mat{0.5& 0.5& -6}^T$ 
(to the \textit{left} of the obstacle) to reach a target $\vect{p}_{tg}=\mat{0.5 & 4.5 & -6}^T$
(to the \textit{right}).  The accompanying video shows that the
optimisation is able to find a trajectory that allows the robot to
successfully overpass the obstacle.  
We believe the chosen obstacle size is representative of most medium-sized obstacles commonly 
found in nature and can reasonably be cleared with a single jump. For larger obstacles (e.g., 
a rock pillar), multiple jumps may be necessary—for example, an initial jump to land on the obstacle, 
followed by a second jump to surpass it. To explore this scenario, we conducted an additional 
simulation where the robot jumps onto a taller obstacle (e.g., $R_x=2.5$ m). 
One limitation of the actual
approach, is that targets that are in the shade cone of the obstacle,
could not be reached.  This is because the ropes would collide with
the obstacle, unless the obstacle is small enough and the ropes come
from the sides.  

\subsection{Landing test and lateral manoeuvring (experiment 7)}
\label{sec:landing_results}
\begin{figure}[t]
	\centering
	\includegraphics[width=1.0\dimexpr\scalefactor\columnwidth]{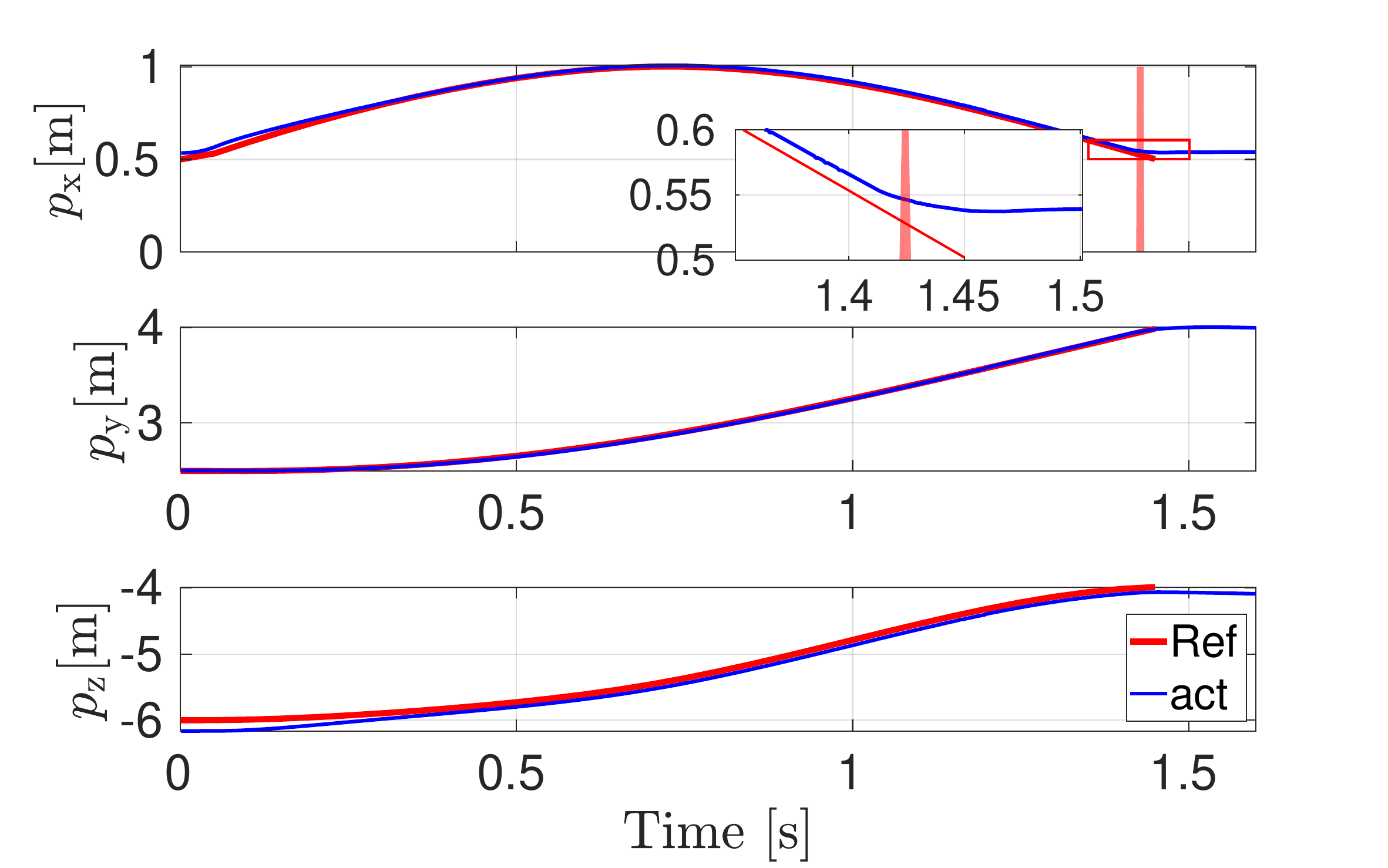}
	\caption{\small \textit{Experiment 7}. Simulation of the landing test for a wall inclination of 
            0.1 rad from the vertical. The red and blue plots are the 
            reference and actual position of the robot, respectively. 
		The vertical red shaded line highlights  the early touch-down moment. }
	\label{fig:landing_results}
\end{figure}

\noindent \textbf{Landing:} In this section, we conduct a simulation of the robot 
equipped with the landing mechanism presented in Section \ref{sec:landing}
with the aim of assessing the effectiveness of the landing
strategy in dissipating any excess kinetic energy during touchdown 
without causing any rebound. 
%
%
Depending on the accumulated errors during the flight phase, the
touch-down could be either early or delayed with respect to the
duration of the optimised trajectory.  In the case that the touch-down
is delayed, we use the feed-forward coming from the optimisation to
determine the ropes forces and we apply a gravity compensation
action. 
Due to the additional weight of the lander, in these tests, we
increased the mass to $15$~kg.  This requires that we also ought to
adjust the actuation limits: $f_{\leg,\max} = 600$~N and
$f_{r,\max} = 300$~N.  The landing joints $q_{L,f}$ are controlled
through the PD strategy \eqref{eq:landing_strategy} with stiffness and
damping set to $K_L= 60$~N/m and $D_L= 10 $~N/m respectively.
Figure \ref{fig:landing_results} reports the tracking of the Cartesian
position of the robot for the landing test.  It is evident, inspecting
the first plot ($X$ variable), that the robot is able to land without
re-bounce and that an \textit{early} touch-down occurred.

\noindent \textbf{Lateral Manoeuvring:} After the landing, with the purpose to locally explore the area, the controller \eqref{eq:lat_maneuv_control}  for lateral manoeuvring is used to drive the robot in a lateral motion, by setting $\vect{\dot{p}}^d=\mat{0.&-0.7&0.}$ m/s $\boldsymbol{\omega}_b^d=\mat{0&0&0}$ rad/s. Additionally, we commanded the propeller to apply 25 N of force to push the robot against the wall, enhancing the grip of the wheels. 
The propeller in this case is used solely to enhance contact, as gravity is already counteracted by the ropes, unlike in propeller-wheeled wall-climbing robots, where the propeller must create a bigger normal force such that frictions can  overcome gravity \cite{wall1}.
A bottom view of the resulting motion is shown in the accompanying video. 

\subsection{Jumps on slanted walls (experiment 8)}
\label{sec:slanted_wall}
\noindent The presented approach is generic and for any wall inclination within the vertical. 
Indeed, the wall normal $\mathbf{n}$ is an input for the optimal control planner and can be set according to the value of the inclination of the wall at the lift-off position.
In the accompanying video we present results for upward jumps from $\vect{p}_0 = \mat{\#  & 2.5& -6}^T$~m\footnote{\# indicates that X coordinate is adjusted to be consistent on the wall for the given ($Y$,$Z$) pair.} to a target position
$\vect{p}_{tg} = \mat{\# & 4 &-4}^T$~m and downward jumps from $\vect{p}_0$ to $\vect{p}_{tg} = \mat{\# & 4 &-12}^T$~m 
 on  slanted walls of different inclinations w.r.t the vertical (0.1,0.2,0.3,0.4 rad). 
In Table \ref{tab:leg_impulse_slanted} we report the norm of the leg impulse, the energy consumption and the absolute landing error,  for the different inclinations of the rock wall for both upwards and downwards jumps. 

\begin{table}[!tbp]
\centering
\caption{\textit{Experiment 8}. Slanted wall}
\resizebox{0.8\columnwidth}{!}{
   \begin{tabular}{l  c c c c} 
   \toprule
              \textbf{Jump}  &  \textbf{Wall incl. [rad]} & $\Vert \vect{f}_{\leg}\Vert$[N]    & $E$ [J]   & $\Vert e_a\Vert$  [m]\\ 
    \midrule
     \multirow{5}{*}{Upward} &   0.1                      &  98                                & 145       &  0.074          \\  \cmidrule(l){2-5} 
                             &  0.2                       &  124                               & 166       &  0.066          \\ \cmidrule(l){2-5}      
                             &       0.3                       & 203                                & 211       & 0.077           \\ \cmidrule(l){2-5} 
                             &       0.4                       & 230                                & 176       & 0.067           \\ 
        \midrule
    \multirow{5}{*}{Downward}&   0.1                       &  92                                & 30         &  0.66         \\ \cmidrule(l){2-5} 
                             &   0.2                       &  104                               & 40       &  0.56           \\    \cmidrule(l){2-5}   
                             &   0.3                       & 90                                 & 27       & 0.67          \\ \cmidrule(l){2-5} 
                             &   0.4                       & 114                                & 33       &  0.60         \\
         \midrule                    
 \multirow{3}{*}{Downward}   &0.1                           & 172                                & 186     &  0.1        \\ \cmidrule(l){2-5} 
                             &   0.2                       & 285                                & 186     &  0.072         \\   \cmidrule(l){2-5}    
\multirow{2}{*}{+ $\vect{f}_{r}\leq 15$N}  &   0.3         & 300                                & 178        & 0.065         \\ \cmidrule(l){2-5}  
                             &   0.4                       & 300                                & 189        & 0.064          \\ 
   \bottomrule         					    					    					    
\end{tabular}
}
\label{tab:leg_impulse_slanted}
\end{table}

The output of the optimisation shows that for upward jumps 
the more slanted is the slope the higher is the impulsive pushing force needed to detach the 
robot from the rock, because a bigger component of gravity should be overcome. 
For downward jumps this trend is less evident, on the other hand the energy consumption is significantly  lower
because the robot mostly exploits gravity to reach the target. 
In both cases, the tracking error is unaffected by inclination and remains within the same range, 
whereas it is significantly higher for downward jumps.
This occurs because the optimal solution involves rope forces close 
to the bound $\vect{f}_{r} \leq 0$ N, leaving the \gls{mpc} with minimal flexibility for corrections.
This issue can be mitigated by setting a bound of $\vect{f}_{r}\leq 15$ N. This adjustment provides the \gls{mpc} with sufficient room to correct model uncertainties, reducing the landing errors to values 
comparable to those observed for upward jumps (see Table \ref{tab:leg_impulse_slanted}), 
at the price of higher leg impulses and energy consumption.
\subsection{Energetic comparison with other robotic  solutions}
\label{sec:energy_consumption}

In this section we compare ALPINE with existing state of the art solutions in performing  maintenance operation. 
The main requirements for performing maintenance operations are related to power consumption for navigating to a designated destination and remaining stationary on the wall for 30 minutes to carry out the operation (assuming the energy consumed for the operation is the same across all solutions), as well as the travel time needed to reach the target destination.
The weight of the operating machine is modelled as a 4 kg payload that the robot must carry in addition to its own weight.
The energy consumption will be normalised by the mass of the robot to enable a fair comparison. 
As a representative navigation we consider again a dislocation from $\vect{p}_0 = \mat{0.5 & 2.5 & -6}$~m to $\vect{p}_{tg} = \mat{0.5 & 4 &-4}^T$~m.
In Table \ref{tab:power_consuption}, we report the normalised energy consumption for the ALPINE platform, 
two drones from Nanjing Hongfei Company \cite{HZH}, the lightweight model HZH C680 and the heavy-duty model 
HZH Y100, the Stickybot III \cite{stickybotIII}, and the ROCR climbing robot \cite{ROCR}.
%
\begin{table*}[!tb]
\centering
\caption{Normalised energy comparison with state of the art existing solutions}
\resizebox{\textwidth}{!}{
   \begin{tabular}{l c c c c c c } \hline\hline
       \textbf{Robot}                                        &\textbf{Weight [kg]}      &\textbf{Max Payload Capacity}[kg]&\textbf{Energy navigation}[kJ/kg]&\textbf{Energy stand. still}[kJ/kg]&\textbf{Tot. energy}[kJ/kg]&\textbf{Travel Time}[s]\\
       \hline
       ALPINE                                                & $5$   & $\approx f_r^{\text{max}}/g$ & $0.03$                  & $0$                                 & $0.03$                     & $1.52$      \\
        \hline
        Drone - HZH C680 (light-weight) \cite{HZH}  & $5$   & $1.5$                  & $0.8$                  &       $83$                          & $84$                     & $1.5$        \\
       \hline 
       Drone - HZH Y100 (heavy-duty) \cite{HZH}             & $40$   & $100$               & $0.9$                  & $75$                                & $76$                     & $1.5$      \\ 
       \hline 
       Stickybot III \cite{Kim2008}                          & $1$    & $0.6$                          & $0.03$                  & $0$                    & $0.03$                    & $50$        \\ 
       \hline   
       ROCR \cite{ROCR}                     & $0.55$   & $0$                                         & $0.1$                                & $0$                & $0.1$       & $17$ \\ 
   \hline \hline 					    					    					    
\end{tabular}
}
\label{tab:power_consuption}
\end{table*}

From table \ref{tab:power_consuption}, an important advantage of roped robots becomes evident, 
thanks to the brakes that can block the ropes, the energy consumption for standing still on the wall is 0. The consumption of navigation, because gravity is exploited, 
is very low when descending and higher when ascending.  For walking-based climbing robots, energy consumption is also limited, but payload capacity is significantly constrained, and operating speed is relatively slow. A comparison with the two drones demonstrates ALPINE's superior speed, although drones' energy consumption is largely influenced by the hovering time required to remain on-site, which is zero for the climbing robots. Overall, ALPINE combines the best aspects of both types of solutions.


%
\section{Conclusions}
\label{sec:conclusion}
This paper presents a robotic platform designed to execute rescue and
maintenance operations along mountain slopes. The robot hangs on two
ropes that are attached to anchors deployed on the top of the wall.
The motion is guaranteed by two motors that wind and un-wind the
ropes, and by a retractable leg that pushes away the robot from the
mountain. An auxiliary propeller guarantees the stability of the robot
during the flight phases. We have discussed in depth the design of the
robot, and shown a simplified dynamic model that captures the dominant
components of the system's dynamics, while being mathematically
tractable. The model has been used to design motion planning and
control strategies based on a suitable \gls{mpc} formulation. The
paper also discusses effective means to compute the locations where
the system can be in equilibrium and apply sufficient forces on the
wall (e.g., to scale boulders, or perform hammering operations).  As an essential part
of our work, we also developed a complete and realistic simulation
model that we used to collect a large number of simulation results in
different application scenarios.

This paper could open the way to a large amount of future work,
addressing the problems that this novel platform and its potential
applications have unveiled.  A non-complete list includes:
1. developing a multi-jump motion planner that considers the hybrid
dynamics of the system to plan the system actions across the mode
transitions related to the landing-take-off phase; 2. considering a
more complete simplified model that accounts for rotational dynamics
to improve the control algorithm during the flight phase;
4. investigating the usage of steering wheels for efficient navigation in  areas of the wall that are clear
of obstacles (e.g., slabs);
3. developing control strategies for specific operations (e.g.,
drilling scaling) or to deal with harsh and highly uneven terrains;
4. developing learning approaches to enable jumps on crumbly terrain
with hardly predictable responses.

\bibliographystyle{elsarticle-num-names} 
\bibliography{references}
\section{Appendix*}
Definitions of $\vect{A}_{d}$, $\vect{b}_d$ introduced in~ \eqref{eq:simplified_2ropes_minimal} of the reduced order model.
%
%
%
\renewcommand{\arraystretch}{2.5}
\begin{table*}
\centering
\begin{minipage}{0.75\textwidth}
\begin{align}
\vect{A}_{d} = \mat{ -\vect{p}_z            &          \frac{\vect{p}_x}{l_1} - \frac{\sin(\psi)^2  \vect{p}_y} {2d_a\vect{p}_xl_1}\left(2l_1^2 -d_a\right)           & \frac{ l_2 \vect{p}_y \sin(\psi)^2 }{d_a \vect{p}_x} \\
			     	0            &             \frac{l_1}{d_a}      														         &  -\frac{l_2}{d_a}   					  \\
	 	  \vect{p}_x            &         \frac{\cos(\psi)sin(\psi) \vect{p}_y }{2\vect{p}_x l_1 d_a} \left( 2l_1^2 -d_a \right) + \frac{\vect{p}_z}{l_1} &  - \frac{l_2 \vect{p}_y \cos(\psi)\sin(\psi)}{d_a \vect{p}_x} }
\label{eq:A_dyn}
\end{align}		
\begin{align}	 	
\vect{b}_{d} = \mat{ 
 q_1 -\frac{\sin(\psi)^2 q_2 }{q_4} 
      -\frac{{\vect{p}_y 2d_a}^2 \sin(\psi)  q_1^2 }{ q_3 } 
	  +\frac{\dot{\psi} \vect{p}_y2d_a \cos(\psi) q_1}{ q_6} 
	  +\frac{ \dot{l_1}  \vect{p}_y 2d_a \sin(\psi)  q_1}{ q_6 l_1}   \\ 
 \frac{-{\dot{l_2}}^2+{\dot{l}_1}^2}{d_a} \\
 q_7  + \frac{ \cos(\psi) q_2 } {q_4 }
 +\frac{{\vect{p}_y2d_a}^2 \cos(\psi) q_1^2} { q_3}
 -\frac{\dot{l}_1  \vect{p}_y2d_a \cos(\psi) q_1}{q_6 l_1} 
 + \frac{\dot{\psi} \vect{p}_y 2d_a  \sin(\psi) q_1}{q_6} 
} ^T		  
\label{eq:b_dyn}
\end{align}
with:
\begin{align}	 
\begin{cases}
		q_0 =& 2 \dot{l}_1  \frac{\vect{p}_z}{l1} \dot{\psi} -l_{1}{\dot{\psi}}^2 \frac{\vect{p}_x}{l_1}\\ 
		q_1 =& - l_2^2 \dot{l_1} + 2 l_1 l_2 \dot{l_2} - l_1^2 \dot{l_1} + d_a^2 \dot{l_1}\\
		q_2 =& 3  {\dot{l}_1}^2 {\vect{p}_y2d_a}^2 + 8 l_{1} l_{2} \dot{l}_1 \dot{l_2} \vect{p}_y2d_a -2 	  	{l_{1}}^2 {\dot{l_2}}^2 						
				\vect{p}_y2d_a+4\,{l_{1}}^2\,{l_{2}}^2\,{\dot{l_2}}^2 \\
				&- 6\,{l_{1}}^2\,{\dot{l}_1}^2\,\vect{p}_y2d_a-8\,{l_{1}}^3\,l_{2}\,\dot{l}_1\,\dot{l_2}+4\,{l_{1}}^4\,{\dot{l}_1}^2 \\
		q_3 =& \frac{ 16 d_a^4 {l_{1}}^2 {\vect{p}_x}^3 }{\sin(\psi)^3}\\
		q_4=& \frac{4 d_a^2 {l_{1}}^2 \vect{p}_x }{\sin(\psi)}\\
		%
		%
		q_6 =& \frac{2 d_a^2 l_{1}  \vect{p}_x}{\sin(\psi)}\\  
		q_7 =& - \vect{p}_z \dot{\psi}^2 + 2 \dot{l_1} \dot{\psi} \frac{\vect{p}_x} {l_1} 
\end{cases}\nonumber
\end{align}	
\hrule
\end{minipage}
\end{table*}
\renewcommand{\arraystretch}{1.0}
\end{document}